\documentclass[11pt]{article}
\usepackage[preprint]{acl}

\usepackage{times}
\usepackage{latexsym}
\usepackage[T1]{fontenc}
\usepackage[utf8]{inputenc}
\usepackage{microtype}
\usepackage{inconsolata}
\usepackage{booktabs}
\usepackage{amsmath}
\usepackage{amsfonts}
\usepackage{nicefrac}
\usepackage{multirow}
\usepackage{xcolor}
\usepackage{colortbl}
\usepackage{graphicx}
\usepackage{listings}
\usepackage{tcolorbox}
\usepackage{bm}
\usepackage{pifont}
\usepackage{hyperref}
\usepackage{xspace}
\usepackage{url}
\usepackage{nccmath}
\graphicspath{{images/}}
\usepackage{subcaption}  
\definecolor{desc}{RGB}{99,178,238}
\definecolor{acc}{RGB}{118,218,145}
\definecolor{rej}{RGB}{248,149,136}
\definecolor{uclablue}{rgb}{0.15,0.45,0.68}
\definecolor{c0}{RGB}{0,120,212}
\newcolumntype{C}{>{\Centering\arraybackslash}X}
\newcommand{\dist}{\mathrm{dist}}

\newcommand{\blue}{\cellcolor{uclablue!15}}

\usepackage{algorithm}
\usepackage{algorithmic}

\usepackage{tabularx} 
\usepackage{ragged2e} 
\newtcbox{\mybox}[1][yellow]{on line, arc = 0pt, outer arc = 0pt,
  colback = #1!10!white, colframe = #1!50!black,
  boxsep = 0pt, left = 1pt, right = 1pt, top = 1pt, bottom = 1pt,
  boxrule = 0pt, bottomrule = 1pt, toprule = 1pt, fontupper = \ttfamily}
\usepackage{xcolor}

\definecolor{myblue}{RGB}{39, 116, 174}
\hypersetup{
  breaklinks,
  colorlinks=true,
  linkcolor=uclablue,
  citecolor=uclablue
}

\title{KALE: Enhancing Knowledge Manipulation in Large Language Models via Knowledge-aware Learning}

\author{
    Qitan Lv$^{1,2}$\thanks{Equal contribution.} \quad 
    Tianyu Liu$^{1,2}$\textsuperscript{\textcolor{myblue}{*}} \quad 
    Qiaosheng Zhang$^{2}$\thanks{Corresponding authors.} \quad 
    Xingcheng Xu$^{2}$\textsuperscript{\textcolor{myblue}{$\dagger$}} \quad 
    Chaochao Lu$^{2}$ 
    \\ \vspace{0.5em}
    ${}^1$University of Science and Technology of China \\
    ${}^2$Shanghai AI Laboratory \\
    \texttt{\{qitanlv, tianyu\_liu\}@mail.ustc.edu.cn} \\
    \texttt{\{xuxingcheng, zhangqiaosheng, luchaochao\}@pjlab.org.cn}
}

\begin{document}
\maketitle

\begin{abstract}
Despite the impressive performance of large language models (LLMs) pretrained on vast knowledge corpora, advancing  their knowledge manipulation—the ability to effectively \textbf{recall, reason, and transfer relevant knowledge}—remains challenging. 
Existing methods mainly leverage Supervised Fine-Tuning (SFT) on labeled datasets to enhance LLMs' knowledge manipulation ability. However, we observe that SFT models still exhibit the \textit{known\&incorrect} phenomenon, where they explicitly possess relevant knowledge for a given question but fail to  leverage it for correct answers.
 To address this challenge, we propose KALE (\textbf{K}nowledge-\textbf{A}ware \textbf{LE}arning)—a post-training framework that leverages knowledge graphs (KGs) to generate high-quality rationales and enhance LLMs' knowledge manipulation ability.
Specifically, KALE first introduces a \textbf{K}nowledge-\textbf{I}nduced (KI) data synthesis method that efficiently extracts multi-hop reasoning paths from KGs to generate high-quality rationales for question-answer pairs.
    Then, KALE employs a \textbf{K}nowledge-\textbf{A}ware (KA) fine-tuning paradigm that enhances knowledge manipulation by internalizing rationale-guided reasoning through minimizing the KL divergence between predictions with and without rationales.
  Extensive experiments on {eight} popular benchmarks across
  {six} different LLMs demonstrate the effectiveness
  of KALE, achieving accuracy improvements of up to 11.72\% and an
  average of 4.18\%.

\end{abstract}

\section{Introduction}
\label{sec:intro}

\begin{figure*}[t]
    \centering 
    \includegraphics[width=2\columnwidth]{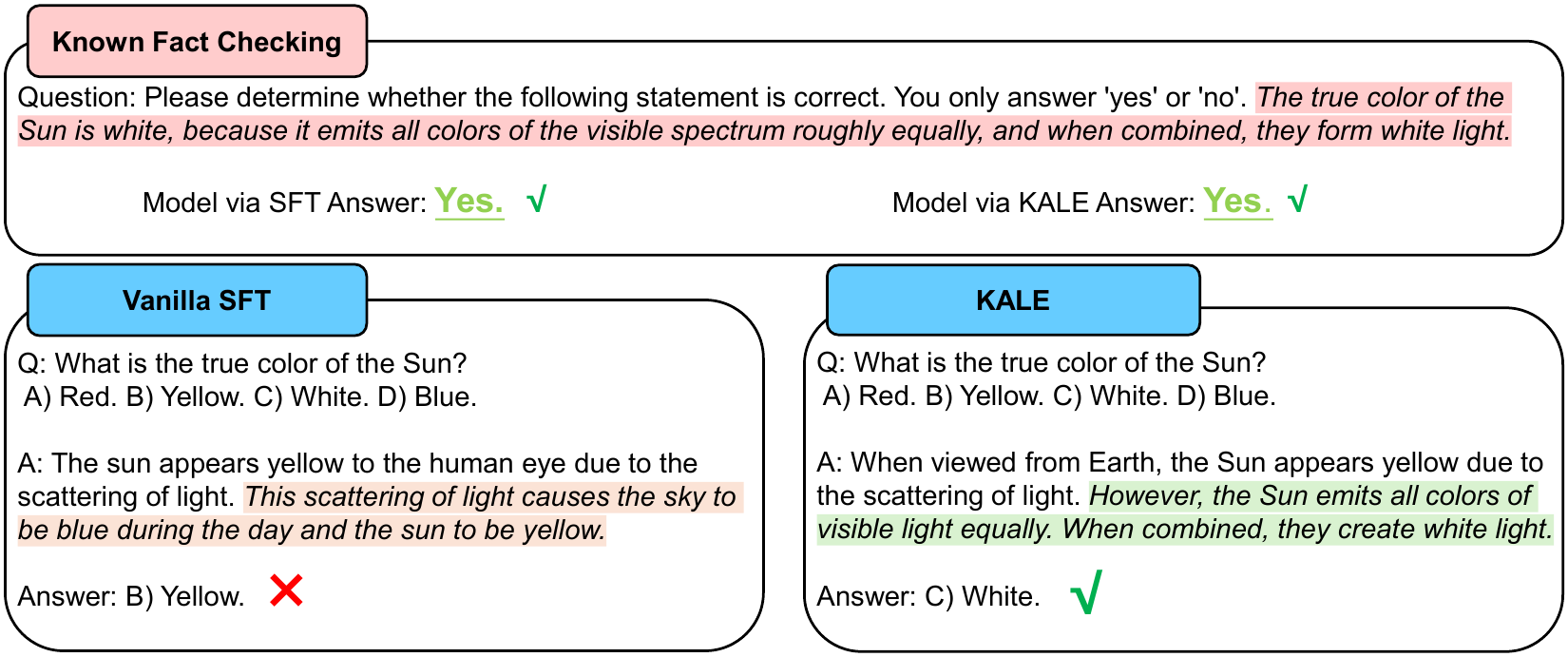}
    \caption{While both post-trained LLMs know relevant knowledge, the LLM via SFT still cannot recall the knowledge to answer. In contrast, KALE effectively recalls the knowledge and answers correctly. We use Mistral 7B \citep{mistral} as an example, and more cases are in Appendix \ref{app:more_case}.}
    \label{fig:case}

\end{figure*}

 Standing out as versatile tools with vast knowledge repositories,
  large language models (LLMs), such as ChatGPT \citep{gpt4},
  Deepseek R1 \citep{deepseek}, and Gemini 3 \citep{gemini},
  demonstrate remarkable power across a wide range of domains
  \citep{domain1, domain3}. However, the most capable LLMs produce
  errors, even when the relevant knowledge is encoded within them,
  indicating struggles to flexibly manipulate
  relevant knowledge during inference \citep{know_mani}.



Recently, extensive research efforts have focused on enhancing LLMs' knowledge manipulation abilities. Among these methods, Supervised Fine-Tuning (SFT) has been widely adopted as a standard post-training approach \citep{sft}. The key idea of SFT is to adapt pre-trained LLMs to specific tasks by training on labeled datasets, thereby optimizing their parameters for task-specific performance \citep{instruction_survey}.
  Several studies have also explored variations of SFT.
  Dual-stage Mixed Fine-Tuning (DMT) \citep{mlt} expands SFT datasets
  to achieve a balance between general and specialized abilities. 
  KG-SFT \citep{kg-sft} utilizes knowledge graphs to filter SFT
  data to enhance LLMs' ability in knowledge-intensive tasks. Extensive
  studies demonstrate the effectiveness and
  versatility of SFT methods \citep{minor_sft, mlt}.



Despite the multiple benefits of SFT methods, LLMs fine-tuned via
  SFT still exhibit the \textit{known\&incorrect} phenomenon---
  \textbf{LLMs possess relevant knowledge but cannot effectively
  manipulate it to answer questions correctly.} This phenomenon mainly
  stems from two limitations in the SFT process: (i) the lack of training data with explicit reasoning rationales. Such high-quality reasoning data is scarce in many domains, and manual creation requires substantial effort, posing significant barriers to broader LLM applications
  \citep{synthetic_data}; and {(ii) the insufficient ability to manipulate task-relevant knowledge}. SFT methods fine-tune LLMs using labeled datasets to learn specific patterns through explicit input-output pairs. However, LLMs often overly rely on these explicit mappings, which restricts their ability to effectively manipulate task-relevant knowledge \citep{rsft}.
  As shown in Figure \ref{fig:case}, although the LLM possesses the knowledge that the true color of the Sun is white, it fails to answer correctly after SFT.
Therefore, even after sufficient SFT process, LLMs still struggle to effectively manipulate task-relevant knowledge to answer correctly \citep{know_mani}.

  To address these challenges, we propose KALE (\textbf{K}nowledge-\textbf{A}ware \textbf{LE}arning)---a novel post-training framework to boost LLMs' knowledge manipulation ability. KALE consists of two components: (i) knowledge-induced data synthesis (\textbf{KI}) that leverages knowledge graphs to generate high-quality rationales, and (ii) knowledge-aware fine-tuning (\textbf{KA}) that enhances knowledge manipulation by minimizing KL divergence between outputs with and without rationales.  Specifically, for a given question-answer pair, KALE \textbf{first} identifies named entities and extracts reasoning paths from the question to answer using a multi-path A* algorithm over knowledge graphs.
   \textbf{Then}, KALE feeds the question-answer pair and reasoning paths to an LLM to generate rationales.
\textbf{Finally}, rather than learning specific patterns through explicit supervised input-output pairs, KALE minimizes the KL divergence \citep{kl} between the output distributions of LLMs with and without rationales. This encourages the two distributions to be aligned, allowing LLMs to effectively manipulate task-relevant knowledge when rationales are absent during inference.

In summary, our key contributions include:


  \begin{enumerate}
      \item [(i)] \textbf{Identification of the known\&incorrect
      phenomenon.} We identify a critical limitation
      of existing SFT methods: LLMs fine-tuned via SFT often possess
      relevant knowledge but fail to manipulate it to answer correctly,
      revealing fundamental challenges in knowledge manipulation during
      post-training.
\item [(ii)] \textbf{A novel knowledge-aware post-training method.} We propose KALE, a unified framework that addresses the known\&incorrect phenomenon through: (i) knowledge-induced data synthesis to generate high-quality reasoning rationales via KGs, and (ii) knowledge-aware fine-tuning to enable effective knowledge manipulation by aligning output distributions with and without rationales.

  \item [(iii)] \textbf{Significant improvement and versatility.}
        We conduct extensive experiments on eight
        benchmarks across six LLMs to demonstrate the
        effectiveness of KALE, yielding a maximum accuracy improvement
        of 11.72\% and an average of 4.18\% over SFT baselines.
  \end{enumerate}
\section{Related Work} \label{sec: related}


\subsection{Text Data Augmentation Methods}

With the advent of LLMs, data augmentation has undergone a significant transformation \citep{survey_da}. LLMs have shown remarkable abilities in generating high-quality text, which provides advantages in data augmentation tasks \citep{dataau1, dataau2}. 
 AugGPT \citep{auggpt} leverages the generative power of LLMs to rephrase questions in SFT data. GPT3Mix \citep{gpt3mix} extends augmentation abilities of LLMs by using few-shot prompting to generate questions semantically similar to the SFT data. StaR \citep{star} utilizes a self-taught mechanism to let LLMs provide internal thoughts. While existing augmentation methods primarily focus on expanding the data quantity of the original data, they lack the multi-hop logic rationales, { KALE can effectively generate textual rationales underlying the Q\&A pair.}

\subsection{KG Retrieval Generation Methods} \label{sec:rela_2}

  Knowledge graphs provide structured, factual knowledge that complements the unstructured, text-based knowledge encoded in LLMs \citep{roadmap}.
  Recent research has explored integrating KGs to enhance LLMs' reasoning ability through retrieval-augmented approaches~\citep{kgcot, tog, graphrag}.
  Think-on-Graph (ToG) \citep{tog} employs iterative beam search over a KG to guide LLM reasoning.
  KGR \citep{kgr} augments LLM responses with factual statements retrieved from KGs.
  KAPING \citep{kaping} enhances zero-shot Q\&A by appending retrieved facts to prompts.
  StructGPT \citep{structgpt} employs an iterative reading-then-reasoning framework over structured data.
  GraphRAG \citep{graphrag} integrates KG traversal to retrieve relationships from graph-indexed data.
  These retrieval-based methods require additional retrieval operations from knowledge bases during inference, introducing additional latency.
  {In contrast, KALE internalizes knowledge-grounded reasoning paths during training, eliminating the need for retrieval during inference (please refer to Appendix \ref{app:inference_time} for inference time comparison for each backbone and baseline).}

\subsection{SFT Variant Methods}

  Supervised Fine-Tuning (SFT) has become a standard approach to adapt pre-trained LLMs to specific downstream tasks \citep{train_sft}.
  Recent work has explored various SFT strategies to improve model performance and generalization.
  Dual-stage Mixed fine-Tuning (DMT) \citep{mlt} improves the general ability of LLMs by balancing task-specific and general knowledge during fine-tuning.
  Self-Distillation Fine-Tuning (SDFT) \citep{sdt} leverages a distilled dataset generated by the model itself to maintain the original distribution and reduce catastrophic forgetting \citep{forget}.
  KG-SFT \citep{kg-sft} uses KGs to filter high-quality SFT data for knowledge-intensive tasks.
  However, these SFT-based methods learn from explicit input-output mappings, which limits their flexibility in manipulating task-relevant knowledge when faced with varied input queries.
  {In contrast, KALE employs a knowledge-aware fine-tuning paradigm that aligns LLM distributions with and without rationales, enabling more flexible and dynamic knowledge manipulation during inference.}

\section{Preliminaries} \label{sec: prelin}

  \subsection{Notations}
  We denote $\textbf{x}^{\text{ins}}$ as instructions for downstream tasks, $\textbf{x}^{\text{que}}$ as queries, $\textbf{x}^{\text{ans}}$ as answers, and $\textbf{x}^{\text{rats}}$ as rationales.
  We define two types of input prompts for LLMs: one includes the rationale, represented as $(\textbf{x}^{\text{ins}}, \textbf{x}^{\text{que}}, \textbf{x}^{\text{rats}})$, and the other excludes the rationale, represented as $(\textbf{x}^{\text{ins}}, \textbf{x}^{\text{que}})$.
  Let $\mathcal{E}_{q}=[\textbf{e}_{q_1}, \textbf{e}_{q_2}, \textbf{e}_{q_3}, \ldots]$ denote the question entity list extracted from $\textbf{x}^{\text{que}}$, and $\mathcal{E}_{a}=[\textbf{e}_{a_1}, \textbf{e}_{a_2}, \textbf{e}_{a_3}, \ldots]$ denote the answer entity list extracted from $\textbf{x}^{\text{ans}}$.
  Let $\mathcal{P}=[\textbf{p}_{1}, \textbf{p}_{2}, \textbf{p}_{3}, \ldots]$ denote the reasoning path list connecting entities in $\mathcal{E}_{q}$ to entities in $\mathcal{E}_{a}$, where $\textbf{p}_{i}$ represents the $i$-th reasoning path.

  \subsection{A* Algorithm} \label{sec:A*}
  A* algorithm~\citep{astar, M_astar} is a heuristic search algorithm that extends traditional shortest path algorithms by incorporating a heuristic function to guide the search.
  Unlike uniform search strategies, A* prioritizes nodes with lower estimated total cost, reducing the search space:
  \begin{equation}\label{eqn:priority_func}
      f(\textbf{e}) = g(\textbf{e}) + h(\textbf{e})
  \end{equation}
  where $g(\textbf{e})$ is the cost of the current shortest path from start entity $\textbf{e}_{start}$ to entity $\textbf{e}$, and $h(\textbf{e})$ is a heuristic function estimating the cost from $\textbf{e}$ to the target entity $\textbf{e}_{end}$. $f(\textbf{e})$ represents the estimated total cost for reaching the target through $\textbf{e}$.


\begin{figure*}[t]
    \centering

    \includegraphics[width=2\columnwidth]{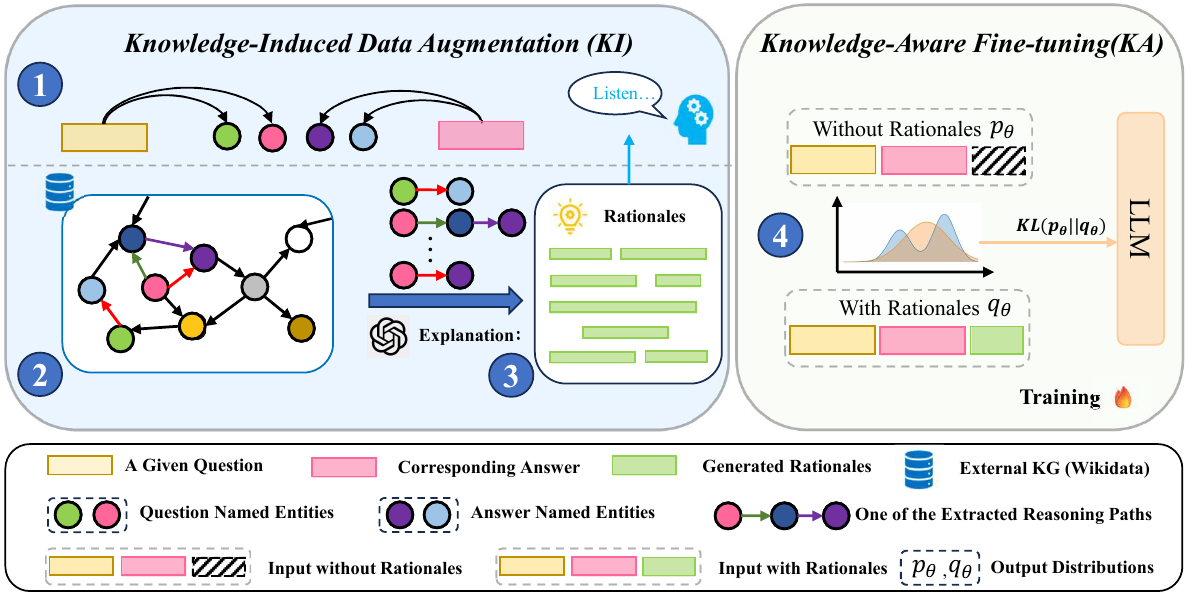}
    \caption{An overview of KALE. For a given Q\&A pair, the workflow of KALE is as follows. \textbf{(1)} Perform named entity recognition to extract relevant question and answer entities. \textbf{(2)} Search for reasoning paths via the proposed multi-path A* algorithm. \textbf{(3)} Combine reasoning paths with the Q\&A pair to generate the corresponding rationales via GPT-4o. \textbf{(4)} Align the LLM's output distributions with and without rationales via knowledge-aware fine-tuning.}
    \label{fig:overview}
\end{figure*}

\section{Method} \label{sec: method}



\subsection{Knowledge-induced Data Synthesis}\label{sec:multi-astart}

 Answering a question often requires integrating multiple knowledge fragments. For instance, for the question \textit{"What is the true color of the Sun?"} and answer \textit{"White"}, it involves multiple pieces of knowledge such as: \textbf{(i)} "The Sun emits all colors of the visible spectrum," \textbf{(ii)} "The combination of all visible light produces white light," and \textbf{(iii)} "The balanced intensity distribution of sunlight integrates into white light." The fragmented nature of such knowledge in pre-training data makes it challenging for LLMs to manipulate relevant knowledge. In contrast, KGs provide a way to organize fragmented knowledge into structured relationships. Specifically, this knowledge can be formalized into a reasoning path: "\textit{the Sun--emits-->full-spectrum light--integrates\_into-->white light}," which corresponds to a series of interconnected triples within a KG, including [the Sun, emits, full-spectrum light] and [full-spectrum light, integrates
  into, white light]. Building upon this, we propose knowledge-induced data synthesis (KI) to generate rationales.\footnote{We only generate rationales during training. This example is for illustration; KALE does not introduce additional overhead during inference (please refer to Figure \ref{fig:case}).}

Specifically, KALE \textbf{first} performs named entity recognition separately on the question and the answer, resulting in the question entity list $\mathcal{E}_q = \{\text{the Sun}\}$ and the answer entity list $\mathcal{E}_a = \{\text{white}\}$.
  \textbf{Then}, KALE leverages these entities to search for reasoning paths in a KG. Conducting a full breadth-first search (BFS) from the question entities to the answer entities in a large KG (e.g., Wikidata\footnote{We use Wikidata as the default external KG to extract all reasoning paths. To evaluate the robustness of KALE, we report results using alternative KGs in Appendix \ref{sec:diff_kgs}.}) is time-consuming. For instance, the extraction of reasoning paths from the AbsR training set \citep{meaningful} \textbf{requires over one week}. Therefore, we propose an efficient multi-path A* algorithm to extract reasoning paths. It requires \textbf{less than 4 hours} to extract all reasoning paths on the same set. Specifically, we adopt a small set of \textit{anchor entities}. For a given entity pair $\textbf{e}_q$ and $\textbf{e}_a$ in $\mathcal{E}_q$ and $\mathcal{E}_a$, we select $k$ anchor entities by randomly sampling from the $m$-hop neighbors of  $\textbf{e}_{a}$, thereby extracting a local subgraph around the answer entity. For each anchor, we conduct a limited $3$-step BFS to pre-compute partial distances, which serve as a lower bound for the remaining path cost in A*.

  Formally, let $g(\textbf{e})$ be the accumulated cost (the number of edges traversed) from the start entity to the current entity $\textbf{e}$, and $h(\textbf{e})$ be the heuristic function estimating the cost from $\textbf{e}$ to the answer entity $\textbf{e}_{a}$. We define the priority function as $f(\textbf{e}) = g(\textbf{e}) + h(\textbf{e})$, where $f(\textbf{e})$ is the priority value in A*. To ensure $h(\textbf{e})$ does not overestimate the actual distance, we use the maximum of anchor-based lower bounds derived from the BFS. Let $\{\alpha_1, \alpha_2,\dots, \alpha_k\}$ be $k$ anchor entities. We pre-compute $\mathrm{dist}(\alpha_i, \textbf{e})$ up to depth $d$; if $\textbf{e}$ is not reachable within $d$ steps, we set $\mathrm{dist}(\alpha_i, \textbf{e})= \infty$. Likewise, we compute $\mathrm{dist}(\alpha_i, \textbf{e}_{a})$ for each anchor. Then we let

  \begin{equation} \label{eq:he}
      h(\textbf{e}) \;=\; \max_{1\le i\le k} \Bigl[\mathrm{dist}(\alpha_i, \textbf{e}_{a}) \;-\; \mathrm{dist}(\alpha_i, \textbf{e})\Bigr]^+
  \end{equation}

  where $[x]^+ = \max(x, 0)$ ensures non-negative values. Intuitively, if $\textbf{e}$ is already closer to the answer entity than $\alpha_i$, this difference provides a nontrivial lower bound; otherwise, it contributes zero and does not lead to overestimation. We prove the admissibility of our multi-path A* algorithm via the proposed heuristic function in Appendix \ref{app:a_proof}. This heuristic design is simple yet efficient for reasoning path retrieval in a large KG. We can also apply KG embedding-based methods \citep{kge_survey, a*net} to incorporate semantic information, and we leave it as future work.

  To retrieve multiple reasoning paths, we extend the standard A* algorithm by incorporating a priority queue $\mathcal{Q}$, which stores multiple paths leading to the same entity. Each entry in $\mathcal{Q}$ is a tuple $(f(\textbf{e}), g(\textbf{e}), \textbf{e}, p_{\textbf{e}}^{i})$, where $p_{\textbf{e}}^{i}$ is the $i$-th path from the start entity $\textbf{e}_q$ to the current entity $\textbf{e}$.
  Algorithm \ref{algorithm1} in Appendix \ref{app:a_star} provides the pseudocode for the overall procedure. After obtaining $\mathcal{P}$, we combine the Q\&A pair and $\mathcal{P}$ as input to GPT-4o, prompting it to generate the rationale $\textbf{x}^\text{rats}$ underlying the Q\&A pair (Appendix \ref{app:prompt_temp} provides prompt details). For example, given the extracted reasoning path "\textit{the Sun--emits-->full-spectrum light--integrates\_into-->white light}," the generated rationale is "\textit{The Sun emits light that contains the entire visible spectrum. When these different colors of light are combined, they create white light.}" These rationales offer high-quality textual reasoning data from question to answer, enabling better understanding of the underlying logic and correlations. We include more examples of reasoning paths and rationales in Appendix \ref{sec:example_path_rats} to provide a comprehensive understanding of KALE.

\subsection{Knowledge-aware Fine-tuning Paradigm}


 When confronted with a question, humans typically retrieve relevant experiences and knowledge, reason based on them, and then provide a response \citep{na2, na1}. Motivated by this, we propose a simple yet effective learning paradigm called knowledge-aware fine-tuning, which encourages LLMs to recall relevant knowledge and reason with it before generating a response.



  Formally, consider an LLM denoted by $\mathcal{M}$ with parameters $\theta$ and input $\textbf{x}^\text{inp}=(\textbf{x}^{\text{ins}}, \textbf{x}^{\text{que}}, \textbf{x}^{\text{ans}})$, where $\textbf{x}^{\text{ins}}$ denotes instructions, and $\textbf{x}^{\text{que}}$ and $\textbf{x}^{\text{ans}}$ denote the Q\&A pair. The LLM models the conditional probability of output $\textbf{x}^{\text{out}}$. We consider two probabilities, which differ in whether rationales are included as input:
  \begin{subequations} \label{eq:5}
      \begin{flalign}
          &\mathcal{M}_{w/o}
          = -\sum_t \log p_\theta(\textbf{x}^{\text{out}}_t \mid \textbf{x}^\text{inp},\textbf{x}^{\text{out}}_{<t})\label{eq:5a} \\
          &\mathcal{M}_{w}
          = -\sum_t \log q_\theta(\textbf{x}^{\text{out}}_t \mid \textbf{x}^\text{inp}, \textbf{x}^\text{rats}, \textbf{x}^{\text{out}}_{<t}) \label{eq:5b}
      \end{flalign}
  \end{subequations}


  Equation \eqref{eq:5a} represents the classical process of LLM generation, where a given instruction and query are provided as input, and the LLM produces an output. We aim for the LLM to manipulate learned knowledge and reason with it. In Equation \eqref{eq:5b}, we use the generated rationales \( \textbf{x}^\text{rats} \) as input to the LLM to enable better recall of knowledge fragments relevant to the question.

  Therefore, our goal is to enable the LLM to implicitly leverage relevant knowledge based on the instruction and query before generating a response. To achieve this, we propose knowledge-aware fine-tuning to minimize the divergence between the two distributions in Equations \eqref{eq:5a} and \eqref{eq:5b} as follows:

  \begin{equation} \label{eq:loss}
      \mathcal{L}(\theta) = \mathbb{E}_{(\textbf{x}^\text{inp},\textbf{x}^{\text{out}},\textbf{x}^\text{rats})} \left[ \text{KL}\left( p_\theta \| q_\theta \right) \right]
  \end{equation}

  where \( \text{KL}(\cdot \| \cdot) \) denotes the KL divergence. We use two distributions: $p_\theta$ (i.e., $p_\theta(\textbf{x}^{\text{out}}_t \mid \textbf{x}^{\text{inp}}, \textbf{x}^{\text{out}}_{<t})$), which is updated during training, and $q_\theta$ (i.e., $q_\theta(\textbf{x}^{\text{out}}_t \mid \textbf{x}^{\text{inp}}, \textbf{x}^{\text{rats}}, \textbf{x}^{\text{out}}_{<t})$), which is fixed and serves as the alignment target. By minimizing the KL divergence in Equation \eqref{eq:loss}, KALE does not require outputs without rationales to exactly match those produced with rationales. Instead, it encourages the two distributions to align, which enables the LLM to flexibly manipulate task-relevant knowledge when rationales are absent during inference.

\section{Experiments} \label{sec: exp}

  We aim to evaluate the effectiveness of KALE in enhancing LLMs' knowledge manipulation ability and its versatility across different settings. To achieve this, we organize the experiments into the following parts:

  \begin{itemize}

      \item To demonstrate the superiority, we conduct comparative experiments on {eight} benchmarks across {six} different LLMs.

      \item To investigate the contribution of each component, we conduct ablation studies.

      \item To provide more insights, we conduct case studies on the {known\&incorrect phenomenon} and {ratios of augmented rationales}.

      \item To demonstrate the versatility, we evaluate KALE in {knowledge-intensive domains} across six different languages in Appendix \ref{app:domain}.

      \item To analyze KALE's real-world deployability, we evaluate its {inference time and hyperparameter sensitivity} in Appendix \ref{app:inference_time} and \ref{sec:hyper_sensi}.

      \item To demonstrate the effectiveness of rationales:
  \begin{enumerate}
        \item [(i)] We employ {different KGs} to generate reasoning paths in Appendix \ref{sec:diff_kgs}.

        \item [(ii)] We use {rationales of other LLMs} to show robustness in Appendix \ref{sec:diff_llm}.
        \item [(iii)] We generate {irrelevant and contrastive rationales} in Appendix \ref{sec:diff_type_rat}.
        \item [(iv)] We evaluate {the quality of the generated rationales} in Appendix \ref{sec:rat_quality}.

      \end{enumerate}
      \item To provide an in-depth understanding, we conduct comparisons including (i) combining KALE with SFT, (ii) evaluations on open-ended settings, (ii) comparisons with thinking-style models, self-taught settings, and GRPO, (iv) prompt distillation, and (v) joint training settings in Appendix \ref{sec:combine} to \ref{app:j19}.

  \end{itemize}

\begin{table*}[!t]
\centering
\caption{Results of our KALE using LlaMA3 $8$B, Mistral $7$B, and Qwen2.5 $32$B as backbone models (for more results of different backbone models, please see Appendix \ref{app:backbone}). We \textbf{bold} the best results and \underline{underline} the suboptimal results for each backbone model.}
\label{tab:main_res}
\resizebox{\textwidth}{!}{
\begin{tabular}{lllcccccccc}
\toprule
\textbf{Backbone}         & \textbf{Category} & \textbf{Method}    & \textbf{AbsR} & \textbf{ARC-c} & \textbf{ARC-e} & \textbf{Common} & \textbf{MMLU}  & \textbf{BBH}   & \textbf{RACE-h} & \textbf{RACE-m} \\ 
\midrule
\multirow{16}{*}{\textbf{LlaMA3 8B}} 
    & \multirow{2}{*}{\textbf{Prompt-based}}  
        & \textbf{Vanilla}   & 62.68 & 66.79 & 69.90 & 58.72 & 55.88 & 46.54 & 53.35 & 57.02 \\
    &                        & \textbf{CoT}       & 63.15 & 71.67 & 69.34 & 54.67 & 56.83 & 48.55 & 54.31 & 57.02 \\
    \cmidrule{2-11}
    & \multirow{3}{*}{\textbf{Retrieval-based}} 
        & \textbf{TOG}       & 65.98 & 69.93 & 72.23 & 61.87 & 56.97 & 48.81 & 58.60 & 59.80 \\
    &                        & \textbf{StructGPT} & 65.35 & 70.50 & 73.34 & 62.32 & 58.87 & 49.58 & 60.03 & 60.86 \\
    &                        & \textbf{GraphRAG}  & 75.83    & 74.83    & {75.76}    & 61.51    & 57.28    & \underline{55.83}    & 60.89    & 69.57    \\
    \cmidrule{2-11}
    & \multirow{5}{*}{\textbf{SFT-based}}    
        & \textbf{SFT}       & 67.77 & 68.23 & 71.74 & 59.79 & 58.00 & 45.39 & 56.17 & 58.91 \\
    &                        & \textbf{SDFT}      & {76.15} & \underline{74.91} & 71.44 & 62.24 & 58.78 & 52.37 & 56.88 & 61.03 \\
    &                        & \textbf{DMT}       & 74.57 & 70.82 & 72.84 & 61.43 & \underline{59.11} & 50.14 & \underline{61.46} & 60.64 \\
    &                        & \textbf{MeanLearn} & 71.09 & 72.53 & 74.53 & \underline{63.39} & 58.79 & 50.61 & 60.03 & 61.84 \\
    &                        & \textbf{KG-SFT}    & \underline{78.20}    & 73.12    & \underline{79.55}    & 63.09    & 58.79    & 53.68    & 64.98    & 62.47    \\
    \cmidrule{2-11}
    & \multirow{4}{*}{\textbf{Augmented-based}} 
        & \textbf{STaR}      & 69.95 & 71.50 & 70.99 & 58.20 & 53.41 & 50.07 & 61.21 & \underline{64.32} \\
    &                        & \textbf{AugGPT}    & 64.45 & 72.22 & {75.29} & 55.12 & 56.82 & 51.90 & 59.21 & 60.16 \\
    &                        & \textbf{GPT3Mix}   & 68.27 & 70.57 & 74.24 & 61.33 & 57.79 & {53.92} & 61.03 & 62.67 \\
    &                        & \blue{\textbf{KALE (ours)}} & \blue{\textbf{83.62}} & \blue{\textbf{81.23}} & \blue{\textbf{86.45}} & \blue{\textbf{65.69}} & \blue{\textbf{63.27}} & \blue{\textbf{57.33}} & \blue{\textbf{68.61}} & \blue{\textbf{74.12}} \\
\midrule
\multirow{16}{*}{\textbf{Mistral 7B}} 
    & \multirow{2}{*}{\textbf{Prompt-based}} 
        & \textbf{Vanilla}   & 62.35 & 52.05 & 68.31 & 39.15 & 37.43 & 28.68 & 50.14 & 55.92 \\
    &                        & \textbf{CoT}       & 67.18 & 58.45 & 66.08 & 36.94 & 43.57 & 31.60 & 55.15 & 58.98 \\
    \cmidrule{2-11}
    & \multirow{3}{*}{\textbf{Retrieval-based}} 
        & \textbf{TOG}       & 64.60 & 57.25 & 70.41 & 50.78 & 41.35 & 31.29 & 52.20 & 56.96 \\
    &                        & \textbf{StructGPT} & 65.17 & 57.94 & 69.28 & 46.11 & 44.94 & 32.98 & 55.69 & 60.10 \\
    &                        & \textbf{GraphRAG}  & 68.26    & 57.76    & 71.93    & 48.24    & 45.53    & 35.12    & 57.15    & 62.60    \\
    \cmidrule{2-11}
    & \multirow{5}{*}{\textbf{SFT-based}} 
        & \textbf{SFT}       & 68.48 & 55.89 & 71.55 & 44.14 & 48.86 & 34.90 & 57.09 & 61.00 \\
    &                        & \textbf{SDFT}      & \underline{73.82} & 61.01 & 73.61 & 51.84 & \underline{52.19} & 34.97 & \underline{64.32} & \underline{65.53} \\
    &                        & \textbf{DMT}       & 73.22 & 57.00 & 72.85 & 49.71 & 50.49 & 35.89 & 61.64 & 64.42 \\
    &                        & \textbf{MeanLearn} & 70.97 & {64.42} & 71.55 & 47.83 & 50.95 & 35.58 & 61.06 & 64.42 \\
    &                        & \textbf{KG-SFT}    & 72.39    & \underline{65.96}    & 72.94    & \underline{54.55}    & 52.10    & 34.20    & 61.15    & 63.37    \\
    \cmidrule{2-11}
    & \multirow{4}{*}{\textbf{Augmented-based}} 
        & \textbf{STaR}      & 70.02 & 57.85 & \underline{74.53} & 49.80 & 41.02 & 35.89 & 55.09 & 59.12 \\
    &                        & \textbf{AugGPT}    & 65.28 & 59.73 & 72.77 & 48.24 & 40.24 & 33.21 & 57.75 & 59.96 \\
    &                        & \textbf{GPT3Mix}   & 59.72 & 61.69 & 71.93 & {53.97} & 39.84 & \underline{36.04} & 56.75 & 60.10 \\
    &                        & \blue{\textbf{KALE (ours)}} & \blue{\textbf{76.90}} & \blue{\textbf{71.59}} & \blue{\textbf{77.95}} & \blue{\textbf{59.05}} & \blue{\textbf{54.21}} & \blue{\textbf{39.26}} & \blue{\textbf{67.98}} & \blue{\textbf{70.06}} \\
\midrule
\multirow{16}{*}{\textbf{Qwen2.5 32B}} 
    & \multirow{2}{*}{\textbf{Prompt-based}} 
        & \textbf{Vanilla}   & 66.35 & 75.09 & 80.10 & 65.52 & 80.47 & 69.01 & 71.47 & 76.95 \\
    &                        & \textbf{CoT}       & 68.72 & 76.79 & 82.07 & 66.34 & 81.65 & 69.79 & 73.58 & 77.64 \\
    \cmidrule{2-11}
    & \multirow{3}{*}{\textbf{Retrieval-based}} 
        & \textbf{TOG}       & 74.64 & 80.55 & 84.13 & 68.63 & 83.27 & 72.09 & 74.12 & 78.34 \\
    &                        & \textbf{StructGPT} & 74.17 & 82.43 & 83.29 & \underline{71.58} & 83.41 & 71.93 & 75.56 & 77.92 \\
    &                        & \textbf{GraphRAG}  & 75.24    & 80.20    & 84.18    & 69.00    & 84.85    & 73.20    & 75.84    & 77.37    \\
    \cmidrule{2-11}
    & \multirow{5}{*}{\textbf{SFT-based}} 
        & \textbf{SFT}       & 72.03 & 79.61 & 83.33 & 67.89 & 82.82 & 70.40 & 73.99 & 79.39 \\
    &                        & \textbf{SDFT}      & 73.34 & 80.80 & 84.30 & 71.25 & 84.13 & 71.01 & 74.59 & 80.71 \\
    &                        & \textbf{DMT}       & 75.24 & 81.48 & 86.07 & 70.43 & 85.17 & \underline{73.62} & 75.72 & 80.01 \\
    &                        & \textbf{MeanLearn} & 71.09 & 76.37 & 84.18 & 69.12 & 83.61 & 72.85 & 74.64 & 81.82 \\
    &                        & \textbf{KG-SFT}    & 78.91    & 78.41    & 84.13    & 69.62    & 84.26    & 72.39    & 74.80    & 80.43    \\
    \cmidrule{2-11}
    & \multirow{4}{*}{\textbf{Augmented-based}} 
        & \textbf{STaR}      & 72.99 & 83.87 & 84.60 & 69.21 & 85.24 & 73.16 & 76.30 & 80.43 \\
    &                        & \textbf{AugGPT}    & 78.91 & 84.47 & 86.27 & 68.96 & 85.04 & 71.93 & \underline{77.16} & \underline{81.89} \\
    &                        & \textbf{GPT3Mix}   & \underline{80.10} & \underline{85.23} & \underline{87.33} & 69.53 & \underline{85.69} & 73.47 & 76.21 & 80.77 \\
    &                        & \blue{\textbf{KALE (ours)}} & \blue{\textbf{91.82}} & \blue{\textbf{89.93}} & \blue{\textbf{94.90}} & \blue{\textbf{75.02}} & \blue{\textbf{88.59}} & \blue{\textbf{77.91}} & \blue{\textbf{81.76}} & \blue{\textbf{86.70}} \\
\bottomrule
\end{tabular}
}
\end{table*}

  \subsection{Experimental Setup} \label{sec:exp_set}
  \paragraph{Models and Benchmarks.} We use \textbf{six} open-source LLMs ranging from $7$B to $32$B parameters, including LlaMA3 $8$B \citep{llama3}, Mistral $7$B \citep{mistral}, Qwen2.5 $32$B \citep{qwen2.5}, Gemma2 $9$B \citep{gemma2}, OLMOE $7$B \citep{olmoe}, and Orca2 $7$B \citep{orca2}. Experiments are conducted on $8$ NVIDIA SXM A100 $80$G GPUs for models under $32$B parameters, and on $16$ NVIDIA SXM A100 $80$G GPUs for $32$B models.
  We evaluate on benchmarks for \textbf{logical reasoning}, including AbsR \citep{meaningful}, Commonsense (denoted as Common) \citep{common}, and Big Bench Hard (BBH) \citep{bbh}; \textbf{reading comprehension}, including RACE-H and RACE-M \citep{race}; and \textbf{natural language understanding}, including MMLU \citep{mmlu}, ARC-c, and ARC-e \citep{arc}. We use \textbf{accuracy} as the evaluation metric. More details and baseline descriptions are provided in Appendix \ref{app:more_datails}.


\subsection{Main Results} \label{sec:main_res}

  Table \ref{tab:main_res} presents results using three representative LLMs at different scales: LlaMA3 $8$B, Mistral $7$B, and Qwen2.5 $32$B. Additional results for three other open-source LLMs are provided in Table \ref{tab:more_backbone3}, Figures \ref{fig:leida_qiansan}, and \ref{fig:leida_housan} in Appendix \ref{app:backbone} to demonstrate the versatility of KALE. From Table \ref{tab:main_res}, we observe that KALE  outperforms other state-of-the-art baselines across all three LLMs by a substantial margin. \textbf{Notably, KALE achieves a maximum accuracy improvement of $11.72\%$ on the AbsR benchmark with Qwen2.5 $32$B as the backbone.}

  We also observe that traditional SFT-based and data augmentation methods yield marginal improvements on downstream tasks, particularly when applied to larger and more powerful LLMs (e.g., only a $1.39\%$ improvement on the BBH benchmark when using Qwen2.5 $32$B). In contrast, KALE delivers consistent and significant improvements for larger LLMs. This indicates that as LLMs scale up and become more capable, SFT-based methods that focus on learning input-output patterns or data augmentation methods that merely increase data quantity are suboptimal for further enhancing LLMs. \textbf{In contrast, KALE improves LLMs' ability to manipulate knowledge, achieving significantly better results for larger LLMs.}


  \subsection{Ablation Study}\label{sec:abla}

  To investigate the contribution of each component within KALE, we conduct ablation studies and present the results in Table \ref{tab:abla}. More results for the other three LLMs are provided in Table \ref{tab:more_backbone3} in Appendix \ref{app:abla}.

  \paragraph{Ablation on Rationale Generation} We denote KALE without \textbf{K}nowledge-\textbf{I}nduced (KI) data synthesis as KALE$_\mathrm{w/o \hspace{1mm} KI}$. That is, we do not utilize our proposed multi-path A* algorithm to provide reasoning paths. Instead, we directly input the Q\&A pair and prompt the LLM to generate rationales. As shown in Table \ref{tab:abla}, we observe that directly prompting LLMs to generate rationales without reasoning paths leads to performance degradation. Notably, when using Mistral $7$B as the backbone, the degradation on the ARC-e dataset reaches $12.50\%$. This demonstrates that the extracted reasoning paths effectively capture the reasoning process, contributing to the generation of higher-quality rationales.

  \paragraph{Ablation on KL Divergence} We denote KALE without \textbf{K}nowledge-\textbf{A}ware (KA) fine-tuning as KALE$_\mathrm{w/o \hspace{1mm} KA}$. That is, we directly train the LLM with rationale data generated from the KG using cross-entropy loss as the objective function. We observe that training LLMs with cross-entropy loss to match outputs with rationales does not achieve satisfactory results. Specifically, when using Mistral $7$B as the backbone on the ARC-e dataset, KALE$_\mathrm{w/o \hspace{1mm} KA}$ results in a $14.90\%$ degradation. This demonstrates the effectiveness of the KL divergence for better knowledge manipulation.

\begin{table*}[!t]
\centering
\caption{Results of the ablation study of KALE, using LlaMA$3$ $8$B, Mistral 7B, and Qwen2.5 $32$B as backbones (We provide more results for the other {three} backbones in  Appendix \ref{app:abla}).}
\label{tab:abla}
\resizebox{2\columnwidth}{!}{
\begin{tabular}{llllllllll}
\toprule
\textbf{Backbone}         & \textbf{Method}    & \textbf{AbsR} & \textbf{ARC-c} & \textbf{ARC-e} & \textbf{Common} & \textbf{MMLU}  & \textbf{BBH}   & \textbf{RACE-h} & \textbf{RACE-m} \\ 
\midrule
\multirow{3}{*}{\textbf{LlaMA3 8B}} 
    & \textbf{KALE$_\mathrm{w/o \hspace{1mm} KI}$}    & 78.91$_{\color{red}{\downarrow4.71}}$ & 76.79$_{\color{red}{\downarrow4.44}}$ & 81.65$_{\color{red}{\downarrow4.80}}$ & 65.52$_{\color{red}{\downarrow0.17}}$ & 60.09$_{\color{red}{\downarrow3.18}}$ & 55.21$_{\color{red}{\downarrow2.12}}$ & 64.15$_{\color{red}{\downarrow4.46}}$ & 69.50$_{\color{red}{\downarrow4.62}}$   \\
    & \textbf{KALE$_\mathrm{w/o \hspace{1mm} KA}$}   & 73.93$_{\color{red}{\downarrow9.69}}$ & 75.26$_{\color{red}{\downarrow5.97}}$ & 78.70$_{\color{red}{\downarrow7.75}}$ & 63.06$_{\color{red}{\downarrow2.63}}$ & 60.74$_{\color{red}{\downarrow2.53}}$ & 53.68$_{\color{red}{\downarrow3.65}}$ & 60.03$_{\color{red}{\downarrow8.58}}$ & 64.76$_{\color{red}{\downarrow9.36}}$   \\
    & \blue{\textbf{KALE}} & \blue{\textbf{83.62}} & \blue{\textbf{81.23}} & \blue{\textbf{86.45}} & \blue{\textbf{65.69}} & \blue{\textbf{63.27}} & \blue{\textbf{57.33}} &  \blue{\textbf{68.61}}     &     \blue{\textbf{74.12 }} \\
\midrule

\multirow{3}{*}{\textbf{Mistral 7B}} 
       & \textbf{KALE$_\mathrm{w/o \hspace{1mm} KI}$}    & 71.09$_{\color{red}{\downarrow5.81}}$ & 66.30$_{\color{red}{\downarrow5.29}}$ & 65.45$_{\color{red}{\downarrow12.50}}$ & 57.58$_{\color{red}{\downarrow1.47}}$ & 52.58$_{\color{red}{\downarrow1.63}}$ & 36.81$_{\color{red}{\downarrow2.45}}$ & 64.95$_{\color{red}{\downarrow3.03}}$ & 66.85$_{\color{red}{\downarrow3.21}}$   \\
    & \textbf{KALE$_\mathrm{w/o \hspace{1mm} KA}$}     & 65.64$_{\color{red}{\downarrow11.26}}$ & 63.91$_{\color{red}{\downarrow7.68}}$ & 63.05$_{\color{red}{\downarrow14.90}}$ & 56.84$_{\color{red}{\downarrow2.21}}$ & 49.05$_{\color{red}{\downarrow5.16}}$ & 35.74$_{\color{red}{\downarrow3.52}}$ & 62.78$_{\color{red}{\downarrow5.20}}$ & 64.00$_{\color{red}{\downarrow6.06}}$ \\
    & \blue{\textbf{KALE}} & \blue{\textbf{76.90}} & \blue{\textbf{71.59}} & \blue{\textbf{77.95}} & \blue{\textbf{59.05}} & \blue{\textbf{54.21}} & \blue{\textbf{39.26}} & \blue{\textbf{67.98}} & \blue{\textbf{70.06}}         \\
\midrule
\multirow{3}{*}{\textbf{Qwen2.5 32B}} 

    & \textbf{KALE$_\mathrm{w/o \hspace{1mm} KI}$}    & 87.32$_{\color{red}{\downarrow4.50}}$ & 87.03$_{\color{red}{\downarrow2.90}}$ & 89.98$_{\color{red}{\downarrow4.92}}$ & 71.01$_{\color{red}{\downarrow4.01}}$ & 86.87$_{\color{red}{\downarrow1.72}}$ & 75.15$_{\color{red}{\downarrow2.76}}$ & 78.04$_{\color{red}{\downarrow3.72}}$ & 83.57$_{\color{red}{\downarrow3.13}}$   \\
    & \textbf{KALE$_\mathrm{w/o \hspace{1mm} KA}$}   & 82.94$_{\color{red}{\downarrow8.88}}$ & 85.32$_{\color{red}{\downarrow4.61}}$ & 88.38$_{\color{red}{\downarrow6.52}}$ & 70.43$_{\color{red}{\downarrow4.59}}$ & 84.91$_{\color{red}{\downarrow3.68}}$ & 76.69$_{\color{red}{\downarrow1.22}}$ & 77.82$_{\color{red}{\downarrow3.94}}$ & 82.94$_{\color{red}{\downarrow3.76}}$   \\

    & \blue{\textbf{KALE}} & \blue{\textbf{91.82}} & \blue{\textbf{89.93}} & \blue{\textbf{94.90}} & \blue{\textbf{75.02}} & \blue{\textbf{88.59}} & \blue{\textbf{77.91}} & \blue{\textbf{81.76}} & \blue{\textbf{86.70}}        \\
\bottomrule
\end{tabular}
}
\end{table*}

\begin{figure}[t]
\centering
\includegraphics[width=\columnwidth]{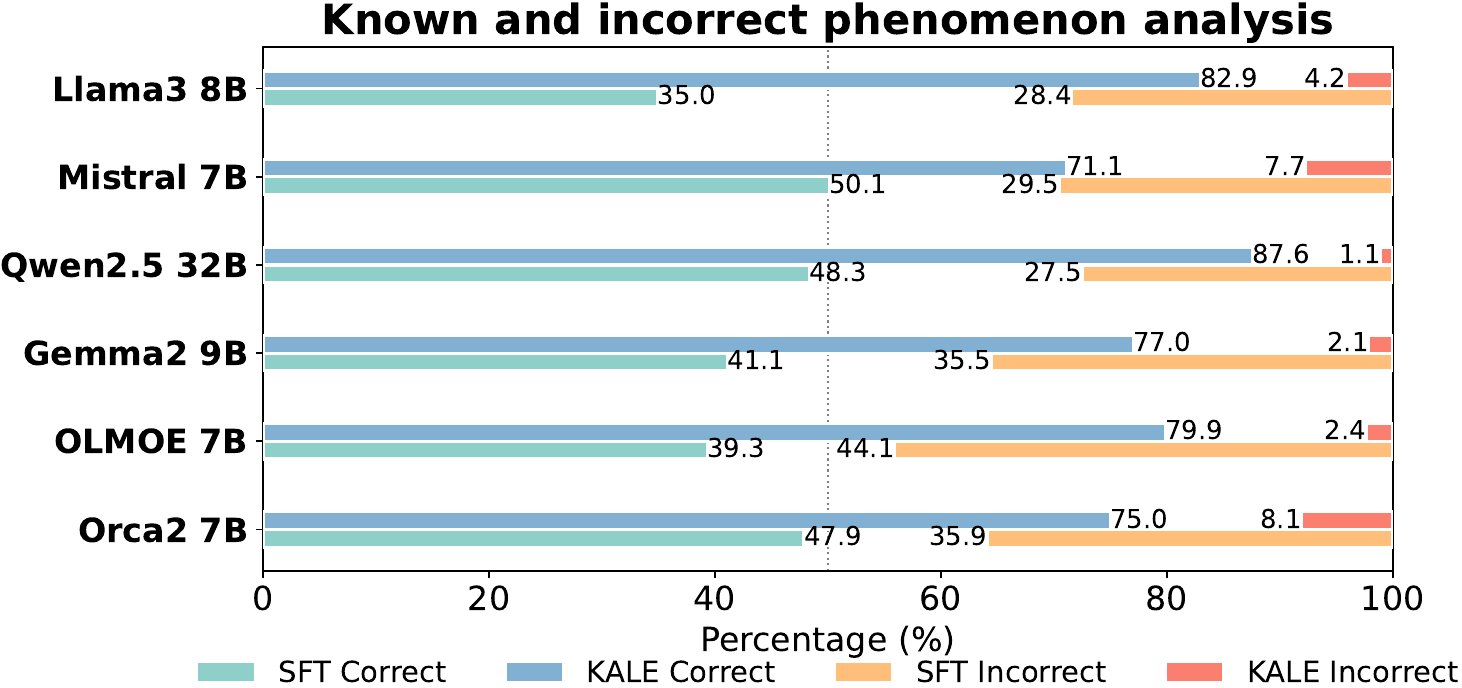}
\caption{\textbf{Known\&incorrect phenomenon analysis:} following the known fact checking in Figure \ref{fig:case}, we collect cases where LLMs possess the knowledge to answer and analyze the ratios of correct and incorrect answers, denoted as known\&correct and known\&incorrect.}
\label{fig:casevisualization1}
\end{figure}

\begin{figure}[t]
\centering
\includegraphics[width=\columnwidth]{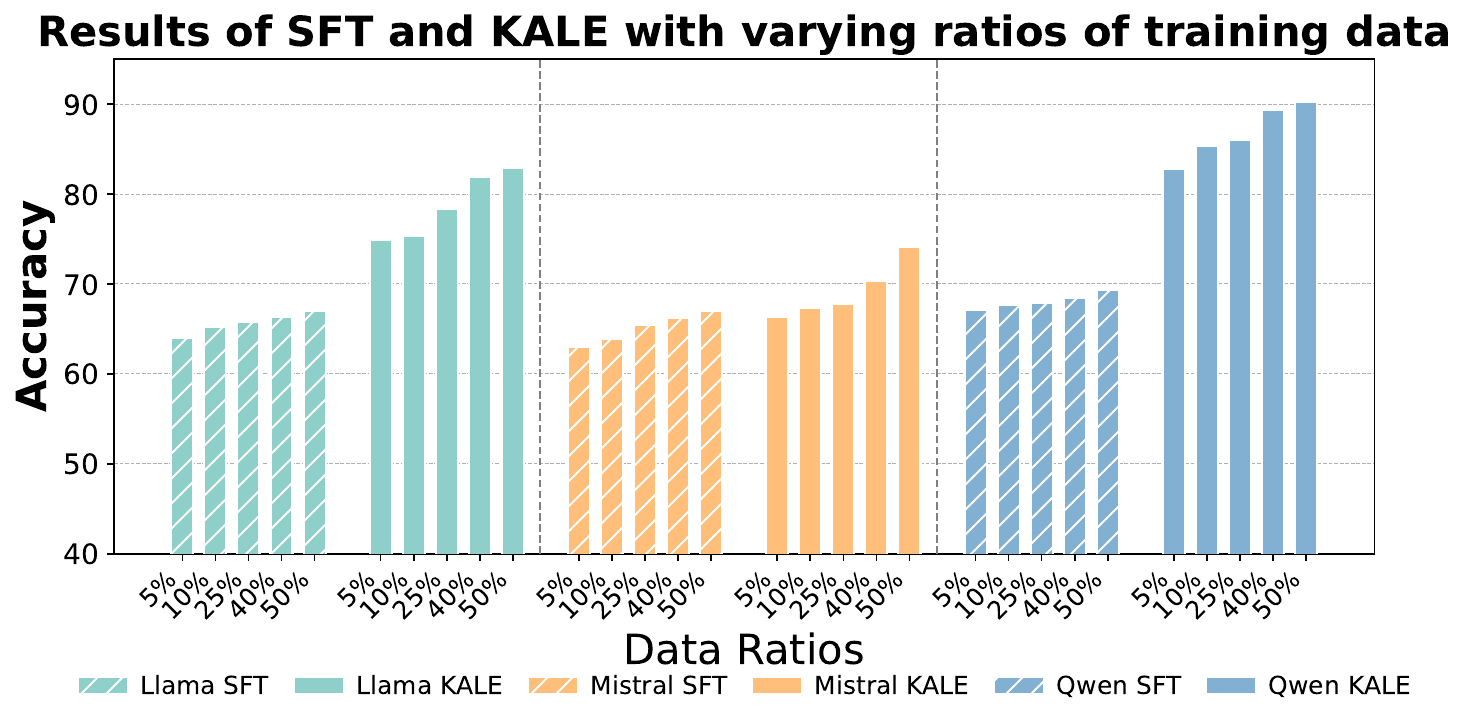}
\caption{\textbf{Ratios of augmented rationales}: by setting the data augmentation ratio from 5\% to 50\%, we explore the differences between KALE and the SFT under varying data scales. We provide results of LlaMA3 8B, Mistral 7B, and Qwen2.5 32B as the backbones as examples, with more results in Appendix \ref{app:case_ratio}.}
\label{fig:casevisualization2}
\end{figure}

  \subsection{Case Study} \label{exp:case}

  \paragraph{Known\&incorrect Phenomenon}

  As shown in Figure \ref{fig:casevisualization1}, LLMs trained via SFT still exhibit the known\&incorrect phenomenon. We provide a detailed analysis of six different LLMs trained with SFT and KALE (see Appendix \ref{app:known} for results of other baselines). We use the known fact checking process in Figure \ref{fig:case} (see Appendix \ref{app:prompt_temp} for prompt details) to categorize LLMs' responses given that LLMs already possess relevant knowledge: (i) \textbf{\textit{Known\&correct}}: LLMs possess the knowledge and correctly answer the question, indicating successful knowledge manipulation. (ii) \textbf{\textit{Known\&incorrect}}: LLMs possess the knowledge yet cannot correctly answer the question, indicating inflexible knowledge manipulation. As shown in Figure \ref{fig:casevisualization1}, we observe that SFT models often exhibit the \textit{known\&incorrect} phenomenon. More than $25\%$ of the questions are cases where the LLM possesses the knowledge but cannot provide
  correct responses. For OLMOE $7$B, it reaches $44.1\%$. In contrast, LLMs trained via KALE demonstrate excellent knowledge manipulation ability, with less than $10\%$ \textit{known\&incorrect} issues across all LLMs. Notably, for the Qwen2.5 $32$B model, this proportion drops to as low as $1.1\%$. This indicates that KALE effectively enhances knowledge manipulation ability.

  \paragraph{Ratios of Augmented Rationales}

  In real-world applications, data acquisition in certain domains can be particularly challenging due to privacy concerns and security restrictions \citep{daijia}. Therefore, we investigate KALE and SFT under limited training data scenarios. Taking the AbsR dataset as an example, by setting the training data ratio from $5\%$ to $50\%$, we present the results in Figure \ref{fig:casevisualization2}. We find that KALE consistently outperforms SFT methods across all levels of augmented rationales. Moreover, this improvement becomes more significant for Qwen2.5 $32$B, which also demonstrates that KALE is highly effective for more powerful LLMs. \textbf{This highlights the significant potential of KALE for low-data, real-world applications.}

  \section{Conclusion} \label{sec: con}

  In this paper, we propose a novel \textbf{K}nowledge-\textbf{A}ware \textbf{LE}arning (KALE) framework to improve the knowledge manipulation ability of LLMs. Specifically, KALE consists of (i) a knowledge-induced data synthesis method to generate high-quality rationales for each Q\&A pair through a structured knowledge graph, and (ii) a knowledge-aware fine-tuning paradigm to enhance the knowledge manipulation ability of LLMs. Extensive experiments on {eight} benchmarks and {six} open-source models across different scales, ranging from $7$B to $32$B, demonstrate the superiority of KALE, delivering significant, consistent, and generalizable improvements.\footnote{More discussions on KALE can be found in Appendix \ref{app:more_discuss}.}

  \section{Limitations}

  We consider a few limitations and future directions.
  {(i) Structured Q\&A dataset requirement.} Current KALE relies on a structured Q\&A dataset to facilitate knowledge-induced data synthesis. For cases where a Q\&A dataset is not available, users can consider employing GPT-4o or other LLMs to transform a raw corpus into a structured Q\&A format. We think applying KALE directly to raw data is a promising direction.
  {(ii) Hard-match reasoning path generation.} When generating reasoning paths, the multi-path A* algorithm is a hard-match approach. Obtaining vectorized embeddings for similarity-based matching is also an optimization direction.
 {(iii) Anchor node selection.} In multi-path A*, we empirically sample $k$ anchor nodes for distance estimation. Finer entity-specific selection (e.g., a neural decision module) may yield better results.
  {(iv) KG availability.} Current KALE relies on an existing KG, which may constrain its applicability in domains where specialized KGs are scarce. Meanwhile, many areas—including the medical domain—already benefit from community-maintained Wikidata, whose ongoing expansion enhances its value for diverse applications. We are further encouraged by advances in the KG community that target automatic construction of domain-specific KGs: methods like SAC-KG \citep{sac-kg} show promise in building high-quality KGs. Such approaches are pivotal for extending KALE to domains where mature KGs are not available.

  \section{Ethical Considerations}
  This work adheres to the ACL Code of Ethics. Our study does not involve human subjects or personally identifiable information, and we only use publicly available datasets under their respective licenses. We transparently report our methods and potential risks and do not recommend deployment in high-stakes settings without further safety assessments.

  \clearpage

\appendix
\clearpage

\section{More Related Works} \label{app:more_rela}

\subsection{Large Language Models}

The advent of pre-trained language models has fundamentally transformed the landscape of natural language processing (NLP), marking a significant paradigm shift in how language understanding and generation tasks are approached. The pioneering work of the GPT series \citep{gpt1,gpt3} introduced the concept of unsupervised pre-training followed by task-specific fine-tuning, demonstrating the effectiveness of leveraging large-scale unlabeled text corpora. This approach was further refined by BERT \citep{bert}, which introduced bidirectional context encoding through the masked language modeling objective, achieving state-of-the-art results across a wide range of NLP benchmarks. Subsequent advancements, such as RoBERTa \citep{roberta}, optimized the pre-training process by removing the next sentence prediction objective and training with larger batches and more data, leading to improved performance. Megatron-LM then \citep{megantron} showcased the scalability of these models, leveraging model parallelism to train significantly larger architectures.
More recently, the field has witnessed the emergence of LLMs that have pushed the boundaries of what is possible in NLP. Models such as LlaMA3 \citep{llama3, llama2}, GPT \citep{gpt4}, PaLM \citep{palm}, Gemini \citep{gemini}, Claude3 \citep{claude3}, and Deepseek V3 \citep{deepseek} have demonstrated remarkable abilities in both few-shot and zero-shot learning scenarios \citep{few-shot}. These models, often comprising hundreds of billions of parameters, have been pre-trained on diverse and extensive benchmarks, enabling them to generalize across a wide array of tasks with minimal or no task-specific fine-tuning. 
The evolution from earlier models like GPT and BERT to the current generation of LLMs underscores the importance of scale and the effectiveness of pre-training on large corpora. These advancements have not only improved performance on traditional NLP tasks but have also enabled new applications and capabilities, such as conversational agents \citep{sac-kg}, code generation \citep{code_llm, pearl}, and complex reasoning tasks \citep{coft}. The continued development and refinement of these models promise to further enhance their utility and impact across various domains.

\subsection{Classic Text Data Augmentation Methods}

Data augmentation has long been a foundational research area in natural language processing (NLP), aimed at enhancing the quality and diversity of training data to improve model generalization and performance. Traditional data augmentation techniques have predominantly focused on character-level and word-level modifications. An example is Easy Data Augmentation (EDA) \citep{eda}, which employs straightforward yet effective strategies such as random insertion, random swapping, random deletion, and synonym replacement to introduce variability into the benchmark \citep{cano1,cano2,cano3}. These methods, while computationally efficient, are often limited in their ability to generate semantically coherent and contextually rich variations, particularly at higher linguistic levels such as sentences or documents.

\subsection{Chain-of-\textbf{X} Approaches in LLMs}

The ability of LLMs to decompose complex problems into a series of intermediate steps and generate internal reasoning processes, known as Chain-of-Thought (CoT) prompting~\citep{cot1}, represents a significant advancement in enhancing their reasoning capabilities. The CoT approach emulates human problem-solving strategies by breaking down intricate problems into smaller, more manageable components. This step-by-step reasoning process allows LLMs to focus on each segment individually, reducing errors and improving logical coherence in their responses \citep{selfconsist}. Moreover, CoT explicitly encourages models to articulate their thought processes, which not only facilitates debugging and refinement of the model’s reasoning but also significantly enhances the interpretability of its outputs. As a result, responses generated through CoT are often more accurate, logically consistent, and contextually relevant compared to those produced by models that directly generate final answers without revealing intermediate cognitive steps.
The success of CoT has inspired a series of follow-up works that extend its principles to other chain-of-X methods, further broadening its applicability and effectiveness. For instance, chain-of-explanation~\citep{coe} focuses on generating detailed explanations to justify the reasoning process, while chain-of-knowledge~\citep{cok} emphasizes the integration of external knowledge to enrich the model’s responses. More recently, chain-of-verification~\citep{cov} has been proposed to enhance the reliability of LLMs by prompting them to draft initial responses, plan verification questions, answer those questions, and generate a final verified response. This iterative verification process reduces the likelihood of misunderstandings or errors in the model’s reasoning. 
Another notable extension is Chain-of-Knowledge~\citep{cokcok}, which elicits LLMs to generate explicit pieces of knowledge evidence in the form of structured triples. This approach is inspired by human cognitive behaviors, where individuals often draw mind maps or knowledge maps as reasoning evidence before addressing complex questions. By structuring knowledge in this way, LLMs can better organize and utilize information, leading to more informed and accurate responses. 


\begin{algorithm}[htb] 
\renewcommand{\algorithmicrequire}{\textbf{Input:}}
\renewcommand{\algorithmicensure}{\textbf{Output:}}
\caption{Pseudo code for Multi-path A* } \label{algorithm1}
\begin{algorithmic}[1]
\REQUIRE  Start node $\textbf{e}_q$, target node $\textbf{e}_a$, maximum number of paths $m$ and maximum search depth $d$

\STATE Initialize priority queue $\mathcal{Q}$ with $(f(\textbf{e}), g(\textbf{e}), \textbf{e}, p_{\textbf{e}}^{i})$
\STATE Initialize reasoning path and visited list $ \mathcal{P}, \mathcal{V} \gets [], []$

\WHILE{$\mathcal{Q} \neq \emptyset$ and $|\mathcal{P}| < m$}
    \STATE Dequeue the element with the smallest $f(\textbf{e})$ from $\mathcal{Q}$
    \STATE Append $\textbf{e}$ into $\mathcal{V}$
    \IF{$\textbf{e} = \textbf{e}_a$}
        \STATE Append $p_{\textbf{e}}^{i}$ into  $\mathcal{P}$ 
        \STATE \textbf{continue} \quad\textcolor{desc}{$\triangleright$ Find Reasoning Path}
    \ENDIF
    
    \IF{$g(\textbf{e}) > d$}
        \STATE \textbf{continue} \quad\textcolor{desc}{$\triangleright$ Path exceeds maximum search depth}
    \ENDIF
    
    \FOR{each neighbor $\textbf{n}$ of $\textbf{e}$}
        \IF{$\textbf{n} \in \text{path}$}
            \STATE \textbf{continue} \quad\textcolor{desc}{$\triangleright$ Avoid cycles}
        \ENDIF
        \STATE Obtain $g(\textbf{n}) \gets g(\textbf{e})+1$ 
        \STATE Compute $f(\textbf{n})$ and $h(\textbf{n})$ via Equations \eqref{eqn:priority_func} and \eqref{eq:he}
        \STATE Enqueue $(f(\textbf{n}), g(\textbf{n}), \textbf{n}, p_{\textbf{e}}^{i} + [\textbf{n}])$ into $\mathcal{Q}$
    \ENDFOR
\ENDWHILE

\ENSURE Reasoning path list  $\mathcal{P}$ 
\end{algorithmic}
\end{algorithm}

\section{More Cases of the \textit{Known\&incorrect} Phenomenon}\label{app:more_case}

\begin{figure*}[t]
    \centering 
    \includegraphics[width=2\columnwidth]{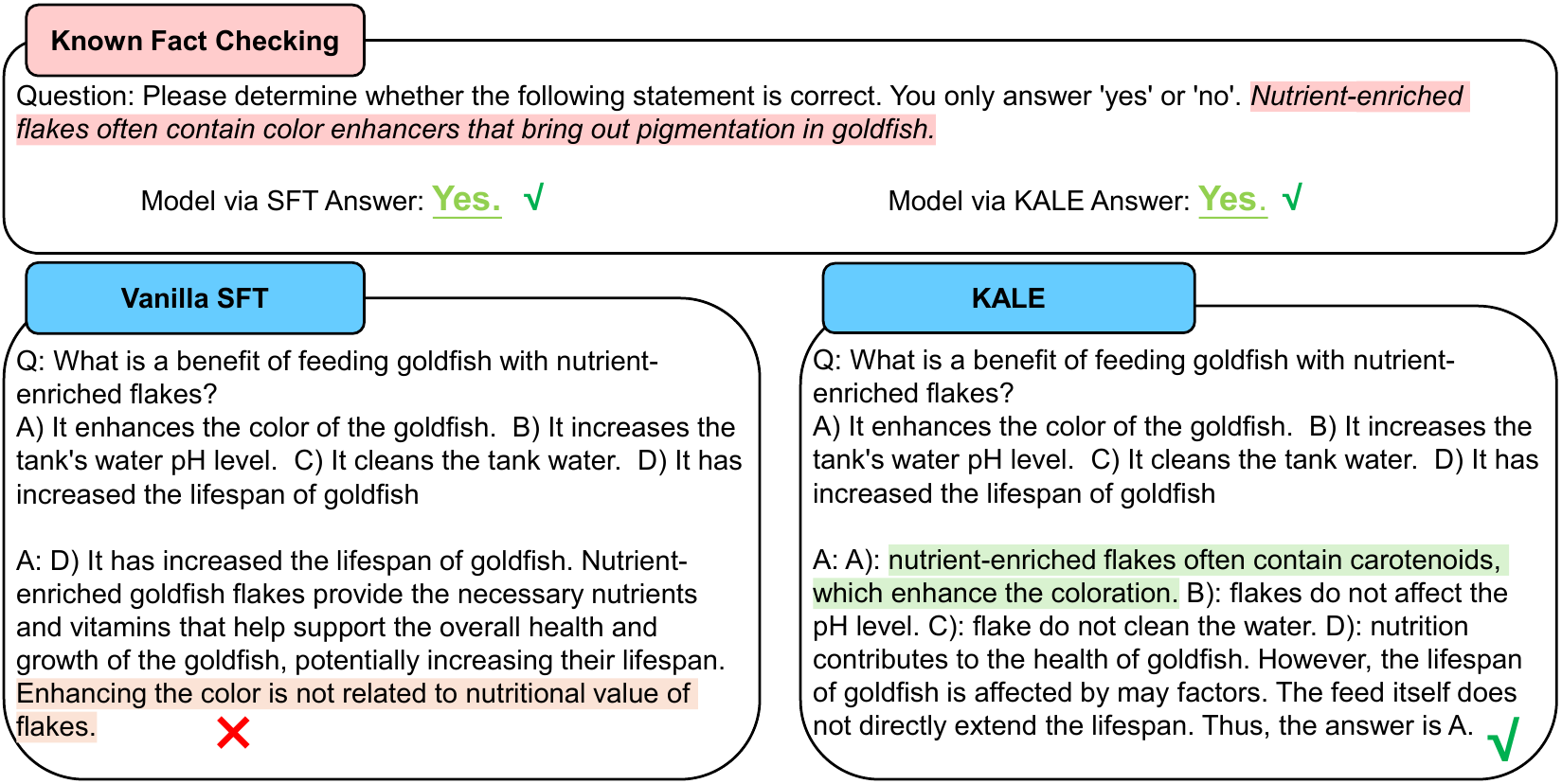}
    \caption{An example of \textit{known\&incorrect} phenomenon comparison using LlaMA3 8B as the backbone model.}
    \label{fig:case_llama}
\end{figure*}

\begin{figure*}[t]
    \centering 
    \includegraphics[width=2\columnwidth]{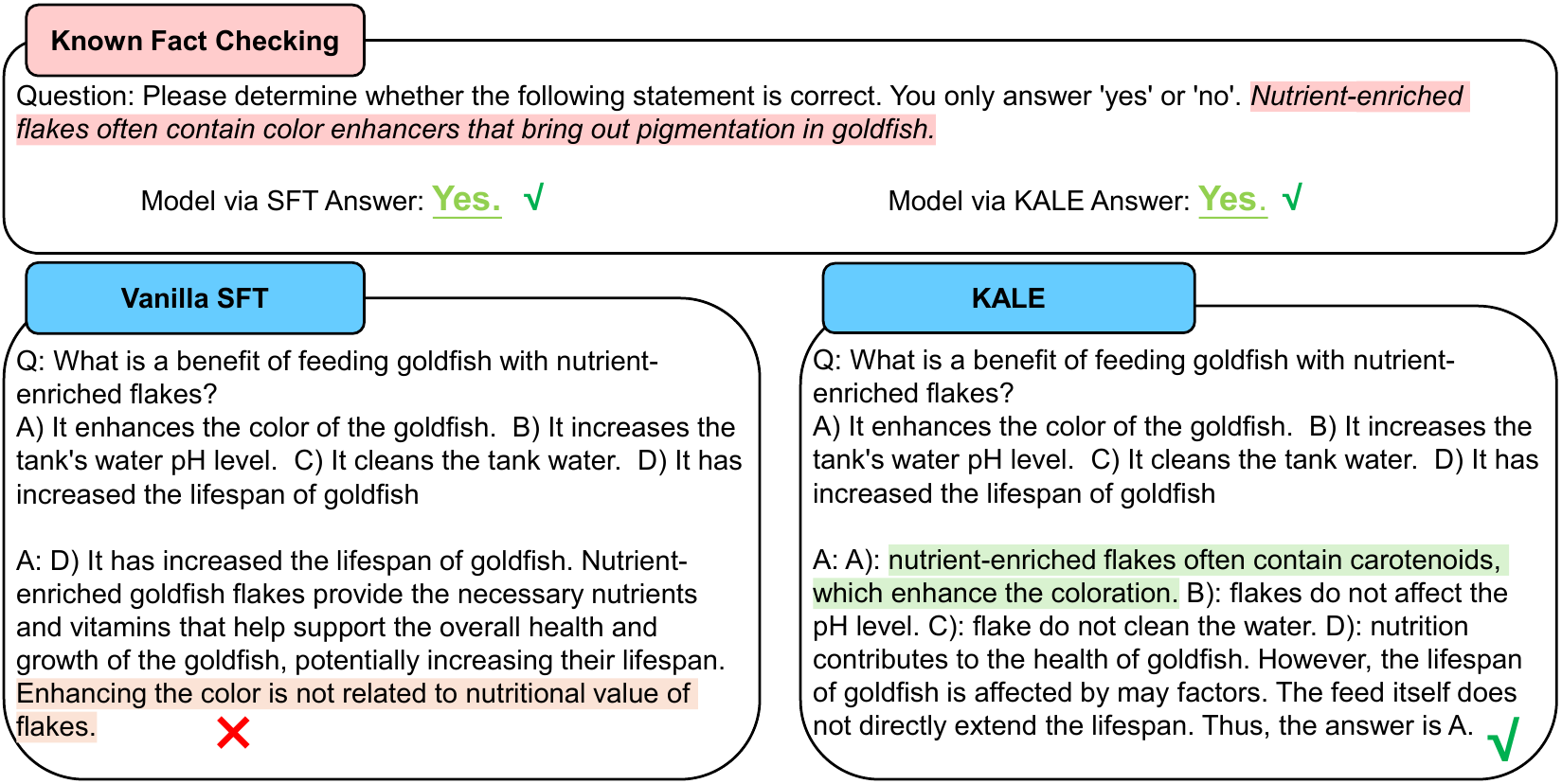}
    \caption{An example of \textit{known\&incorrect} phenomenon comparison using Qwen2.5 32B as the backbone model.}
    \label{fig:case_qwen}
\end{figure*}

\begin{figure*}[t]
    \centering 
    \includegraphics[width=2\columnwidth]{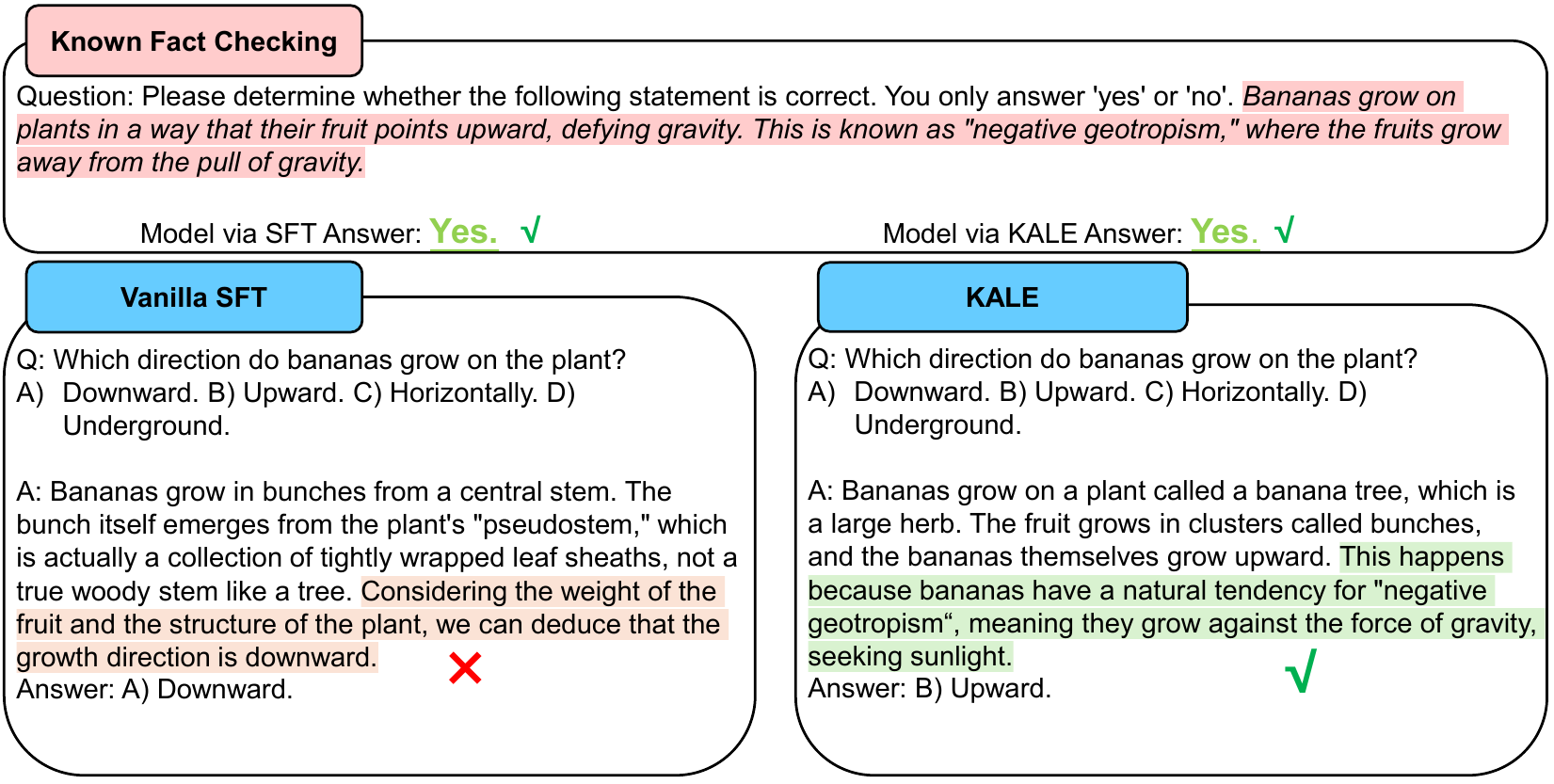}
    \caption{An example of \textit{known\&incorrect} phenomenon comparison using Gemma2 9B as the backbone model.}
    \label{fig:case_gemma}
\end{figure*}

\begin{figure*}[t]
    \centering 
    \includegraphics[width=2\columnwidth]{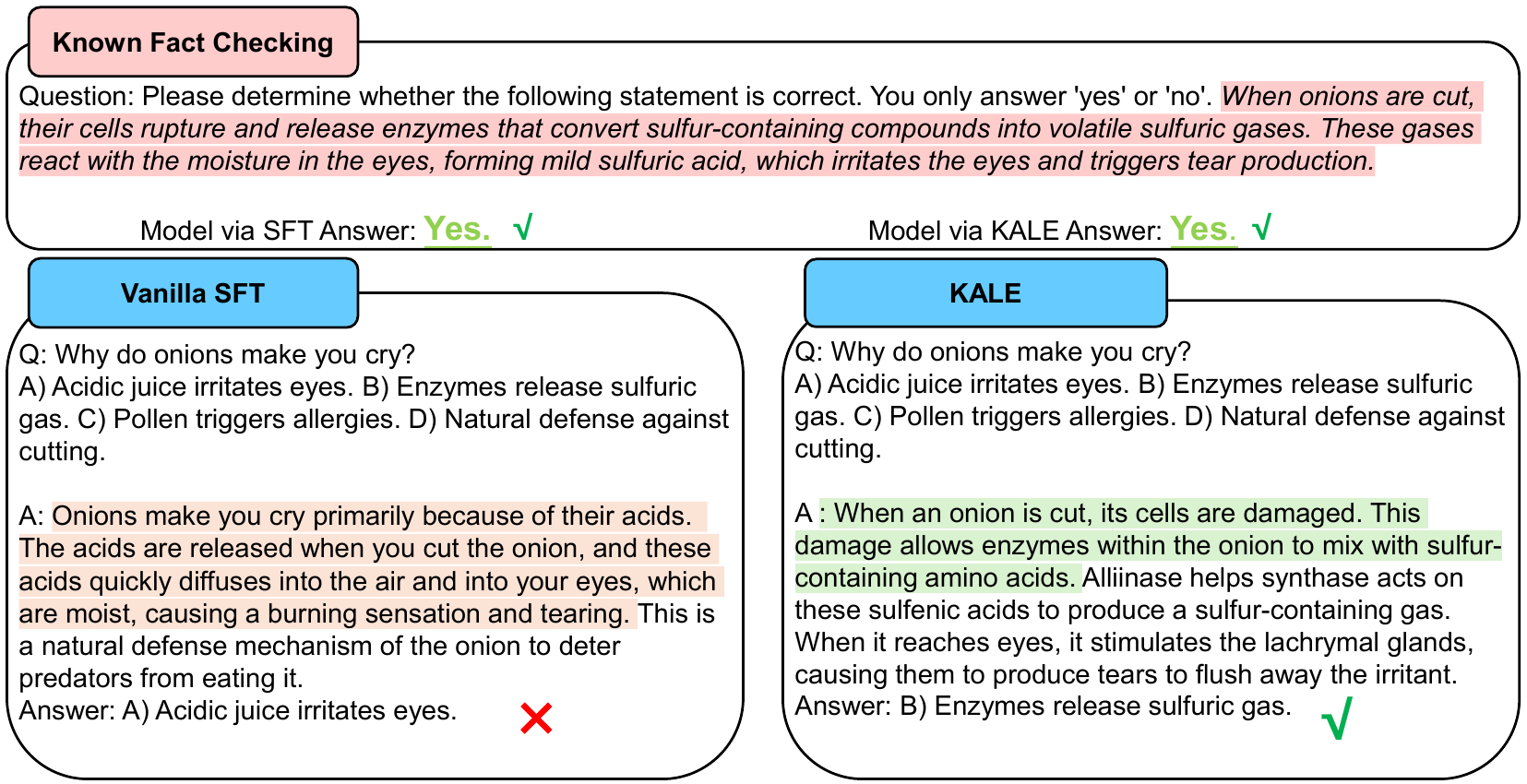}
    \caption{An example of \textit{known\&incorrect} phenomenon comparison using OLMOE 7B as the backbone model.}
    \label{fig:case_olmoe}
\end{figure*}

\begin{figure*}[t]
    \centering 
    \includegraphics[width=2\columnwidth]{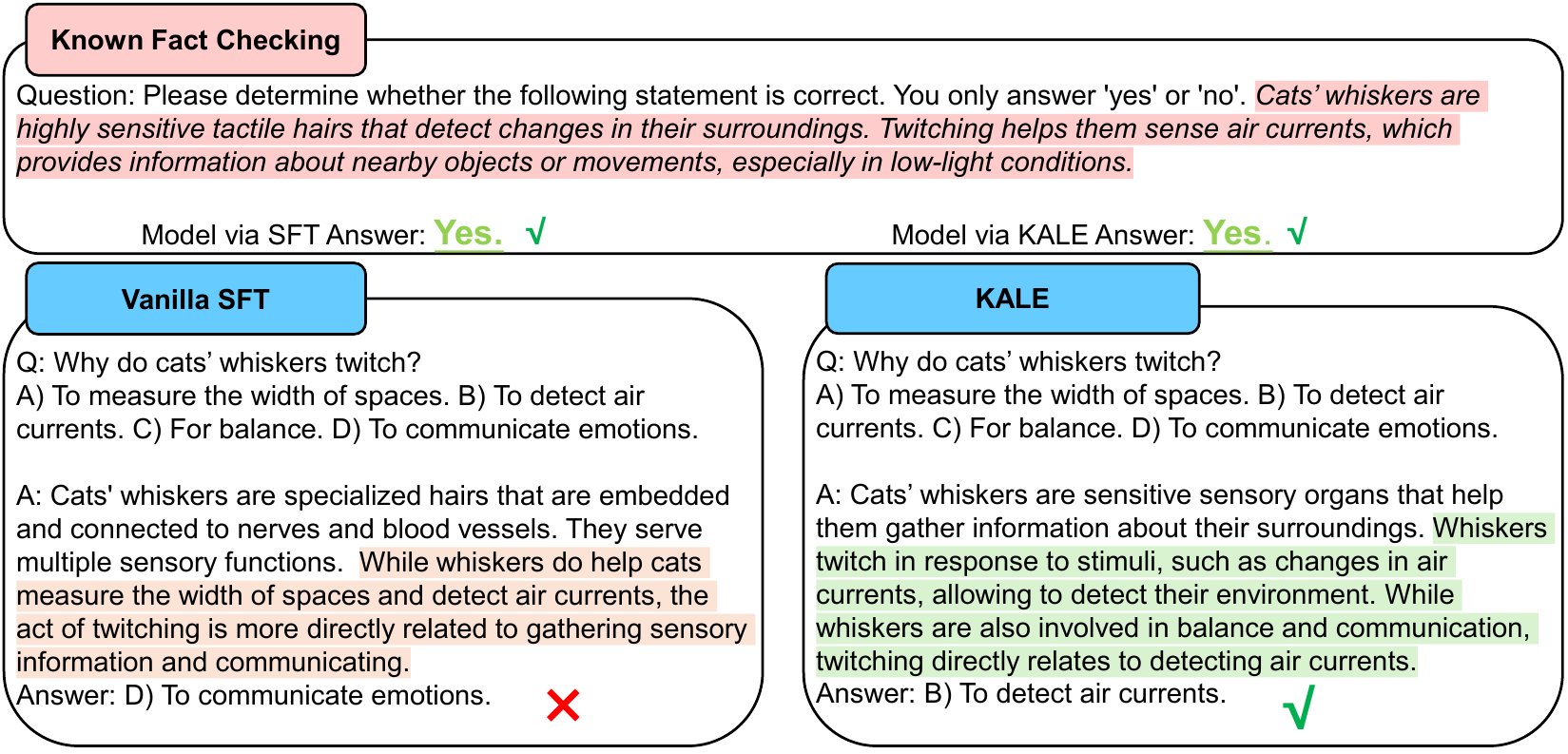}
    \caption{An example of \textit{known\&incorrect} phenomenon comparison using Orca2 7B as the backbone model.}
    \label{fig:case_orca}
\end{figure*}

In Figure \ref{fig:case}, we present a comparative analysis of the \textit{known\&incorrect} phenomenon of models fine-tuned after SFT and KALE, using Mistral-7B as the backbone model. In this section, we further extend the investigation by providing more \textit{known\&incorrect} phenomenon comparisons \textbf{across LlaMA3 8B, Qwen2.5 32B, Gemma2 9B, OLMOE 7B, and Orca2 7B} on various domains to comprehensively demonstrate the efficacy of our proposed KALE. \textbf{As illustrated in Figures \ref{fig:case_llama}, \ref{fig:case_qwen}, \ref{fig:case_gemma}, 
\ref{fig:case_olmoe}, and \ref{fig:case_orca}}, we still find that that models fine-tuned after SFT still exhibit the \textit{known\&incorrect} phenomenon, wherein the models cannot properly recall and apply acquired knowledge to answer correctly despite possessing the relevant knowledge. In contrast, LLMs fine-tuned after KALE demonstrate a better ability to effectively manipulate relevant knowledge to generate correct answers. These results also demonstrate that our KALE effectively strengthens LLMs' knowledge manipulation ability.

\section{Prompt Templates} \label{app:prompt_temp}
We list the prompt templates for different tasks to offer more visually intuitive results for each task in Figures \ref{app:pt_figs1}, \ref{app:pt_figs2}, \ref{app:pt_figs3}, and \ref{app:pt_figs4}. More detailed prompt information for the best performance of each task and dataset can be seen within the code. 

The placeholders \mybox[yellow]{Known Fact}, \mybox[yellow]{Question}, \mybox[yellow]{Answer}, \mybox[yellow]{Reasoning Path}, \mybox[yellow]{Options}, \mybox[yellow]{Generic Fact}, and \mybox[yellow]{Rationales} will be filled with the corresponding terms in each example of corresponding benchmarks.


\begin{figure*}[ht]
    \begin{tcolorbox}[fontupper = \ttfamily, title=Prompt Templates for Known Fact Checking]
    You are a cautious assistant. You carefully follow instructions. You are helpful and harmless and you follow ethical guidelines and promote positive behavior.
    Question: Please determine whether the following statement is correct. You only answer 'yes' or 'no'. \mybox[yellow]{Known Fact}.
    \end{tcolorbox}
\caption{The prompt template used for known fact checking.} 
    \label{app:pt_figs1} %
\end{figure*}


\begin{figure*}[ht]
\begin{tcolorbox}[fontupper = \ttfamily, title=Prompt Templates for Rationale Generation]

    You are a cautious assistant. You carefully follow instructions. You are helpful and harmless and you follow ethical guidelines and promote positive behavior. You are given the question: \mybox[yellow]{Question}. The corresponding answer is: \mybox[yellow]{Answer}. The reasoning paths are: \mybox[yellow]{Reasoning Path}. Please provide a detailed explanatory rationale that references  these reasoning paths. If you determine that the reasoning path is irrelevant to the current QA pair, you may generate rationales based on your own knowledge.

\end{tcolorbox}
\caption{The prompt template used for rationale generation.} 
    \label{app:pt_figs2} %
\end{figure*} 

\begin{figure*}[ht]
\begin{tcolorbox}[fontupper = \ttfamily, title=Prompt Templates for Main Results]

    You are a cautious assistant. You carefully follow instructions. You are helpful and harmless and you follow ethical guidelines and promote positive behavior. You are given a question together with a few options, you should give an explanation first and then answer the question. Your response should follow the format like Explanation: \_\_\_ Answer: \_\_\_ 
    Below is the Question and Options: \mybox[yellow]{Question} \mybox[yellow]{Options}

\end{tcolorbox}
\caption{The prompt template used for main results.} 
    \label{app:pt_figs3} 
\end{figure*} 

\begin{figure*}[ht]
\begin{tcolorbox}[fontupper = \ttfamily, title=Prompt Templates for reasoning trace quality evaluation]

You are a cautious assistant. You carefully follow instructions. You are helpful and harmless and you follow ethical guidelines and promote positive behavior. You are given a rationale for a question.
Evaluate the given rationale along five dimensions—Factual Accuracy, Logical Validity, Coherence, Completeness, and Interpretability. For each dimension, output True if the rationale is correct or meets the criterion; otherwise, output False. You should produce a five-element list in the form like [True,True,True,True,True].
 Below are the Question \mybox[yellow]{Question} and the Rationales \mybox[yellow]{Rationales}.

\end{tcolorbox}
\caption{The prompt template used for reasoning trace quality evaluation.} 
    \label{app:pt_figs4} 
\end{figure*} 




\section{Examples of generated reasoning paths and rationales} \label{sec:example_path_rats}

We present extracted reasoning paths alongside the generated rationales for some samples to provide a more intuitive and straightforward understanding of KALE. We select one sample each from the domains of \textbf{Science, Medicine, Common Knowledge, Computer Science, Economics, and Art} in Figures \ref{app:egr_figs1}, \ref{app:egr_figs2}, \ref{app:egr_figs3}, \ref{app:egr_figs4}, \ref{app:egr_figs5}, and \ref{app:egr_figs6}. For each sample, we provided the extracted reasoning paths and the generated rationales to support a more comprehensive understanding of KALE.
\begin{figure*}[ht]
\begin{tcolorbox}[fontupper = \ttfamily, title=Science Domain Example]
    \textbf{Question:} what is the true color of the Sun?
    
    \textbf{Answer Choices:}
    
    A) Red
    
    B) Yellow
    
    C) White
    
    D) Blue
    
    \textbf{Extracted Reasoning Paths:} 
    
    the Sun--emits-->full spectrum light--integrates\_into-->white light
    
    \textbf{Generated Rationales: }
    
    The Sun emits light that contains the entire visible spectrum. When these different colors of light are combined, they create white light.

\end{tcolorbox}
\caption{An example of generated rationales on science domain.} 
    \label{app:egr_figs1} %
\end{figure*}

\begin{figure*}[ht]
\begin{tcolorbox}[fontupper = \ttfamily, title=Medicine Domain Example]
\textbf{Question:} Which of the following is a typical symptom of cancer?

\textbf{Answer Choices:}

A) Weight gain

B) Persistent fever 

C) Sore muscles

D) Acne   
    
    \textbf{Extracted Reasoning Paths:} 
    
    cancer–may cause–>decreased resistance–may cause–>persistent fever

cancer–may cause–>decreased immune function–may cause–>persistent fever
    
    \textbf{Generated Rationales:}

    The common symptoms that cancer may cause include decreased resistance, which can lead to fever of unknown origin, usually manifested as persistent fever. 

\end{tcolorbox}

\caption{An example of generated rationales on medicine domain.} 
    \label{app:egr_figs2} 
\end{figure*}

\begin{figure*}[ht]
\begin{tcolorbox}[fontupper = \ttfamily, title=Common Knowledge Domain Example]
\textbf{Question:} What do people use to absorb extra ink from a fountain pen?

\textbf{Answer Choices:}

A) shirt pocket

B) calligrapher’s hand

C) desk drawer

D) blotter
    
    \textbf{Extracted Reasoning Paths: }
    
    extra ink--absorbed by-->absorbent paper--also is-->blotter
    
    fountain pen--produces-->excess ink-- absorbed by-->blotter  
    
    \textbf{Generated Rationales:}

    A blotter is a piece of special absorbent paper. People use it to press against fresh ink from a fountain pen to soak up any excess, which prevents smudging and helps the ink dry faster.

\end{tcolorbox}

\caption{An example of generated rationales on common knowledge domain.} 
    \label{app:egr_figs3} 
\end{figure*}

\begin{figure*}[ht]
\begin{tcolorbox}[fontupper = \ttfamily, title=Computer Science Domain Example]
\textbf{Question:} Which protocol secures data for websites (padlock in the address bar)?

\textbf{Answer Choices:}

A) HTTP 

B) FTP 

C) HTTPS 

D) SMTP

    \textbf{Extracted Reasoning Paths: } 

    websites--secure transport-->TLS--implemented as-->HTTPS

    websites--handle-->sensitive data--requires-->encryption--provided by-->HTTPS

    \textbf{Generated Rationales:}

Modern websites handle sensitive user data that requires encryption to prevent interception. HTTPS (Hypertext Transfer Protocol Secure) is the solution; it is essentially the HTTP protocol layered on top of a secure encryption protocol, TLS (Transport Layer Security).

\end{tcolorbox}

\caption{An example of generated rationales on computer science domain.} 
    \label{app:egr_figs4} 
\end{figure*}

\begin{figure*}[ht]

\begin{tcolorbox}[fontupper = \ttfamily, title=Economics Domain Example]
\textbf{Question:} If demand increases while supply remains constant, what happens to the equilibrium price?

\textbf{Answer Choices: }

A) Lower equilibrium price 

B) Stays the same 

C) Higher equilibrium price 

D) Becomes zero

\textbf{Extracted Reasoning Paths: }

demand--shifts right-->demand curve--causes-->higher equilibrium price

\textbf{Generated Rationales: }

A rightward shift in the demand curve, with supply held constant, leads to a higher equilibrium price. This occurs because at the original price, a shortage is created, causing buyers to compete and bid the price upward to a new equilibrium.
\end{tcolorbox}

\caption{An  example of generated rationales on economics domain.} 
    \label{app:egr_figs5} 
\end{figure*}

\begin{figure*}[ht]
\begin{tcolorbox}[fontupper = \ttfamily, title=Art Domain Example]

\textbf{Question:} The technique of dramatic light–dark contrast in painting is called:

\textbf{Answer Choices:} 

A) Impasto 

B) Fresco 

C) Chiaroscuro 

D) Sfumato

\textbf{Extracted Reasoning Paths:}

painting--contrast of-->light and dark--technique named-->chiaroscuro

painting--modeling of form-->using dramatic light--a key feature of-->chiaroscuro

\textbf{Generated Rationales:}

Chiaroscuro is the technique in painting that uses strong, dramatic contrasts between light and dark. Artists employ this method not only to create a sense of volume for modeling three-dimensional subjects, but also to produce a powerful, theatrical mood.
\end{tcolorbox}

\caption{An example of generated rationales on art domain.} 
    \label{app:egr_figs6} 

\end{figure*}

\section{Proof of Admissibility of the Proposed Multi-path A* Algorithm} \label{app:a_proof}

In this section, we show that our proposed heuristic estimated cost in \eqref{eq:he} is admissible,i.e., $h(\mathbf{e}) \le \dist(\mathbf{e}, \mathbf{e}_g)$ for any node $\mathbf{e}$, which means that our proposed multi-path A* algorithm can find the best solution.
We resort to the triangle inequality property of the distance metric $\dist(x,y)$. For any three nodes $A, B, C$, the triangle inequality states:
\begin{equation} \label{eq:triangle_inequality_general}
    \dist(A, C) \le \dist(A, B) + \dist(B, C)
\end{equation}

Let us consider an arbitrary landmark $\alpha_i$ from the set $\{\alpha_i\}_{i=1}^k$.
Applying the triangle inequality with $A = \alpha_i$, $B = \mathbf{e}$, and $C = \mathbf{e}_g$, we have:
\begin{equation} \label{eq:triangle_specific}
    \dist(\alpha_i, \mathbf{e}_g) \le \dist(\alpha_i, \mathbf{e}) + \dist(\mathbf{e}, \mathbf{e}_g)
\end{equation}
Rearranging Equation \eqref{eq:triangle_specific}, we obtain:
\begin{equation} \label{eq:rearranged_inequality}
    \dist(\alpha_i, \mathbf{e}_g) - \dist(\alpha_i, \mathbf{e}) \le \dist(\mathbf{e}, \mathbf{e}_g)
\end{equation}

Now, let $X_i = \dist(\alpha_i, \mathbf{e}_g) - \dist(\alpha_i, \mathbf{e})$. The term in the heuristic function involving $\alpha_i$ is $[X_i]^+ = \max(X_i, 0)$.
We consider two cases for the value of $X_i$:

\begin{enumerate}
    \item \textbf{Case 1: $X_i < 0$}, 
    due to the non-negative property of the distance, we have:
    $$ [\dist(\alpha_i, \mathbf{e}_g) - \dist(\alpha_i, \mathbf{e})]^+ = 0 \le \dist(\mathbf{e}, \mathbf{e}_g) $$

    \item \textbf{Case 2: $X_i \ge 0$}, 
    from Eq.\eqref{eq:rearranged_inequality}, we know that $\dist(\alpha_i, \mathbf{e}_g) - \dist(\alpha_i, \mathbf{e}) \le \dist(\mathbf{e}, \mathbf{e}_g)$.
    Therefore:
    $$ [\dist(\alpha_i, \mathbf{e}_g) - \dist(\alpha_i, \mathbf{e})]^+ \le \dist(\mathbf{e}, \mathbf{e}_g) $$
\end{enumerate}

In both cases, for any anchor $\alpha_i$ ($1 \le i \le k$), we have shown that:
\begin{equation} \label{eq:individual_term_admissible}
    [\dist(\alpha_i, \mathbf{e}_g) - \dist(\alpha_i, \mathbf{e})]^+ \le \dist(\mathbf{e}, \mathbf{e}_g)
\end{equation}
The heuristic function $h(\mathbf{e})$ is defined as the maximum of these terms over all $i$:
$$ h(\mathbf{e}) = \max_{1\le i\le k} \Bigl[ \dist(\alpha_i, \mathbf{e}_g) - \dist(\alpha_i, \mathbf{e}) \Bigr]^+ $$
Since each term $[\dist(\alpha_i, \mathbf{e}_g) - \dist(\alpha_i, \mathbf{e})]^+$ is less than or equal to $\dist(\mathbf{e}, \mathbf{e}_g)$, their maximum must also be less than or equal to $\dist(\mathbf{e}, \mathbf{e}_g)$.
Thus,
\begin{equation}
    h(\mathbf{e}) \le \dist(\mathbf{e}, \mathbf{e}_g)
\end{equation}
This inequality holds for any node $\mathbf{e}$.\textbf{ Therefore, our proposed multi-path A* algorithm is admissible, which means that for any node, our proposed multi-path A* algorithm can find the best solution.}

\section{Pseudo code of the Proposed Multi-path A*} \label{app:a_star}

In Section \ref{sec:multi-astart}, we introduce our multi-path A* algorithm, which efficiently extracts inference paths from question entities to answer entities. Here, we provide the algorithm pseudo code in Algorithm \ref{algorithm1}.



\section{More Details of Benchmarks and Experiment Setups} \label{app:more_datails}

\subsection{Implementation Details}\label{app:imp_details}
In our implementation details, we conduct fine-tuning on all evaluated benchmarks across $3$ epochs with a consistent batch size of $16$, utilizing NVIDIA A100 GPUs ($ 80$ GB) for computational processing. The computational resources are allocated based on model scale, with $8$ GPUs employed for the $7$B and $8$B parameter models, while the larger $32$B parameter models use $16$ GPUs to accommodate their increased computational demands during the fine-tuning process. For all answer entities $\textbf{e}_{a}$, we choose $10$ anchor entities randomly sampled from their $3$-hop neighbors. To guarantee stable and reproducible results, we utilize greedy decoding by setting the temperature parameter to 0 in all experiments. The optimization process employs a peak learning rate of 3e-5, implemented in conjunction with a learning rate warmup strategy that gradually increases the learning rate over the initial $1$$\%$ of training iterations to ensure stable convergence. We set the maximum truncated length as $2048$ for all the benchmarks. 

In the context of Equation~\eqref{eq:loss} for the KA process, $q_\theta$ is fixed and derived from a frozen checkpoint, while $p_\theta$ is fine-tuned starting from a separate checkpoint. The KL divergence is computed at the token level using teacher forcing. 
We apply deepspeed\footnote{https://www.deepspeed.ai/} to accelerate the training process. We implement our approach based on PyTorch 2.5.1\footnote{https://pytorch.org/} and Huggingface's Transformers\footnote{https://github.com/huggingface/transformers}. For the training code of KALE, we modified the training scripts based on LLaMAFactory \citep{llamafactory}. We are committed to providing the source code of our approach, if accepted. During testing, for all models, we follow MeanLearn \citep{meaningful} to use the same system prompt for a fair comparison: "You are a cautious assistant. You carefully follow instructions. You are helpful and harmless, and you follow ethical guidelines and promote positive behavior. You are given a question together with a few options. You should give an explanation first and then answer the question." 
More details for the best performance of each task and benchmark can be seen within our code. 
\subsection{Baseline Methods.}

 We compare \textbf{thirteen} baselines: (i) \textbf{Vanilla:} standalone LLMs without modifications. (ii) \textbf{CoT} \citep{cot1}: prompting LLMs to generate internal thoughts. (iii) \textbf{Think-on-Graph (TOG)} \citep{tog}: applying iterative beam search to enhance LLMs' reasoning ability. (iv) 
 \textbf{StructGPT}  \citep{structgpt}: proposing iterative reading-then-reasoning based on structured data. (v) \textbf{GraphRAG}  \citep{graphrag}: integrating KG traversal to retrieve structured relationships. (vi) \textbf{SFT} \citep{sft}: standalone SFT process. (vii) \textbf{Self-Distillation Fine-Tuning (SDFT)} \citep{sdt}: guiding fine-tuning with a dataset generated by model itself. (viii) \textbf{Dual-stage Mixed Fine-tuning (DMT)} \citep{mlt}: achieving a balance between general and specialized ability. (ix) \textbf{MeanLearn} \citep{meaningful}: teaching LLMs to leverage generic facts. (x) \textbf{KG-SFT} \citep{kg-sft}: utilizing KGs to filter SFT data to enhance LLMs' ability. (xi) \textbf{Self-Taught Reasoner (STaR)} \citep{star}: generating a rationale dataset from a few initials iteratively. (xii) \textbf{AugGPT} \citep{auggpt}: using an LLM to rephrase questions in original data. (xiii) \textbf{GPT3Mix} \citep{gpt3mix}: prompting an LLM to generate similar questions in the SFT data. 

\subsection{Benchmark Details}




\begin{table}[htbp]
    \centering
    \caption{The statistics of AbsR the benchmark.}
    \label{tab:absr}
    \resizebox{\columnwidth}{!}{
    \begin{tabular}{lccc}
        \toprule
              & \textbf{Examples} & \textbf{Questions} & \textbf{Generic Facts} \\ \midrule
        Train & 18,020            & 9,010              & 4,613                  \\
        Test  & 844               & 844                & 104                    \\ \bottomrule
    \end{tabular}}
\end{table}

For \textbf{more details of benchmarks}, we list below all the benchmarks used in \textbf{logical reasoning, reading comprehension, and natural language understanding,} respectively, by KALE as follows.
\textbf{Logical Reasoning Task} we employ AbsR \citep{meaningful}, Commonsense \citep{common}, and Big Bench Hard (BBH) \citep{bbh} as our evaluation benchmarks. Specifically, \textbf{the AbsR benchmark} was constructed using GPT-4 as the primary data annotator, following ~\citep{sac-kg,vicuna}. For each generic fact \( r_i \), GPT-4 was prompted to generate samples \( S_i = \{s_1^i, \dots, s_{m_i}^i \mid 1 \leq m_i \leq 3\} \) in diverse scenarios. Each sample \( s_j^i \) consists of a question \( X_j^i \) with multiple options, a response \( Y_j^i \) containing an answer and an explanation guided by \( r_i \), and forms a triple \( s_j^i = \langle X_j^i, r_i, Y_j^i \rangle \). From each sample in the training set \( s_j^i \), two types of examples were derived: (i) {K-example}, which predicts \( Y_j^i \) given \( \langle X_j^i, r_i \rangle \), and (ii) {R-example}, which predicts \( Y_j^i \) given only \( X_j^i \). These examples are designed to implicitly enhance abstract reasoning in LLMs through the {knowledge} and {reasoning} pathways. In the testing set, only the R-example is provided for each sample. The statistics of the AbsR benchmark are summarized in Table~\ref{tab:absr}. 

\begin{table}[htbp]
    \centering
    \caption{The statistics of the Commonsense.}
    \label{tab:common}
    \resizebox{1\columnwidth}{!}{
    \begin{tabular}{lcr}
        \toprule
        \textbf{Dataset}                                   & \textbf{Task Type} & \textbf{Size} \\ \midrule
        \textbf{$\alpha$NLI}~\citep{bhagavatulaabductive}  & 2 Choices          & 1,507         \\
        \textbf{CSQA}~\citep{talmor2019commonsenseqa}      & 5 Choices          & 1,221         \\
        \textbf{COPA}~\citep{roemmele2011choice}           & 2 Choices          & 500           \\
        \textbf{e-CARE}~\citep{du2022care}                 & 2 Choices          & 2,122         \\
        \textbf{Social IQa}~\citep{sap2019social}          & 3 Choices          & 1,935         \\
        \textbf{PIQA}~\citep{bisk2020piqa}                 & 2 Choices          & 1,838         \\
        \textbf{StrategyQA}~\citep{geva2021did}            & Yes or No          & 2,290         \\ \bottomrule
    \end{tabular}}
\end{table}
\textbf{The Commonsense benchmark} \citep{common} is a multiple-choice question-answering benchmark designed to evaluate the ability of LLMs to perform complex reasoning based on commonsense knowledge. Each question in the benchmark is associated with five candidate answers, only one of which is correct.
The dataset spans a diverse range of domains, including everyday scenarios, social interactions, and physical phenomena, making it a comprehensive testbed for evaluating the commonsense reasoning capabilities of LLMs. We summarize the key statistics and characteristics of Commonsense in Table \ref{tab:common}. For the BBH benchmark \citep{bbh}, it consists of a curated suite of $23$ challenging tasks derived from the broader BIG-Bench benchmark \citep{bb}. These tasks were specifically selected because prior language model evaluations failed to surpass the average human-rater performance, making them particularly suitable for assessing the limits of current models. The tasks span a wide range of domains, including logical reasoning, mathematical problem-solving, and linguistic understanding, requiring models to demonstrate robust reasoning and contextual comprehension. BBH focuses on the importance of structured reasoning pathways in tackling complex tasks. We summarize the filtering process of BBH in Table \ref{tab:bbh}.


\begin{table*}[htbp]
    \centering
    \caption{Filtering criteria to create the BIG-Bench Hard (BBH) benchmark.} 
    \label{tab:bbh}
    \resizebox{\linewidth}{!}{
        \begin{tabular}{r l}
        \toprule
        \bf{\# Tasks} & \bf{Criteria} \\
        \midrule
        \underline{209} & \underline{All BIG-Bench tasks} \\
        187 & - After filtering out tasks with more than three subtasks \\
        130 & - After filtering out tasks with fewer than 103 examples (3 for few-shot, 100 for evaluation) \\
        85 & - After filtering out tasks without human-rater baselines \\
        78 & - After filtering out tasks that do not use multiple-choice or exact match as the evaluation metric \\
        \midrule
        \underline{78} & \underline{Clean multiple-choice or exact match tasks} \\
        36 & - After filtering out tasks in which the best reported model beats average reported human-rater score \\
        23 & - After filtering out extremely difficult tasks that are outside the scope of this work \\
        \midrule
        \textbf{23} & \textbf{Remaining tasks = BIG-Bench Hard (BBH)} \\
        \bottomrule
        \end{tabular}
    }
\end{table*}

\paragraph{Reading Comprehension Task} We employ RACE-M (middle school level reading comprehension task) and RACE-H (high school level reading comprehension task) \citep{race} as our benchmarks. RACE is collected from the English exams for middle and high school Chinese students in the age range between $12$ to $18$. RACE consists of nearly $28,000$ passages and nearly $100,000$ questions generated by human experts (English instructors), and covers a variety of topics that are carefully designed to evaluate the student’s ability to understand and reason.  The reasoning types of RACE include word matching, paraphrasing, single-sentence reasoning, multi-sentence reasoning, and insufficient/ambiguous. We summarize the details  in Table \ref{dataset:race}.

\begin{table*}[htbp]
\centering
\caption{Statistics of the reading comprehension benchmarks, RACE-H and RACE-M. The values below the Training/Valid/Testing Set are the number of passages and questions in each dataset, respectively. Passage/Question/Option Len denotes the average length of the passages, questions, and options, respectively. Vocab size denotes the number of words in the vocabulary.}
\label{dataset:race}
\resizebox{2\columnwidth}{!}{%
\begin{tabular}{l c c c c c c c c}
\toprule
Dataset & Training Set & Valid Set & Testing Set & Passage Len & Question Len & Option Len & Vocab Size\\
\midrule
RACE-M & 6,409/25,421 & 368/1,436 & 362/1,436 & 231.1 & 9.0 & 3.9 & 32,811 \\
RACE-H & 18,728/62,445 & 1,021/3,451 & 1,045/3,498 & 353.1 & 10.4 & 5.8 & 125,120 \\
\bottomrule
\end{tabular}%
}
\end{table*}

\paragraph{Natural Language Understanding Task} For the natural language understanding task, we employ the \textbf{Massive Multitask Language Understanding (MMLU) benchmark} \citep{mmlu} and the ARC benchmark for evaluation. \textbf{MMLU} is a comprehensive dataset designed to assess the breadth and depth of LLMs' knowledge and problem-solving abilities. MMLU consists of 57 tasks spanning diverse domains, including STEM (Science, Technology, Engineering, and Mathematics), humanities (e.g., law, philosophy, history), social sciences (e.g., economics, sociology, psychology), and other specialized fields (e.g., medicine, finance). The dataset comprises $15,908$ questions, divided into three splits: a dev set with $5$ questions per subject for few-shot evaluation, a validation set with $1,540$ questions for hyperparameter tuning, and a test set with $14,079$ questions, ensuring at least $100$ test examples per subject.

\begin{table}[htbp]
\centering
\caption{Statistics for MMLU, ARC-C, and ARC-e datasets.} \label{tan:mmlu_arc}
\resizebox{0.8\columnwidth}{!}{
\begin{tabular}{lccc} \toprule

\textbf{Statistics} & \textbf{Train} & \textbf{Dev} & \textbf{Test} \\ \midrule
\textbf{MMLU} & 99,842 & 1,540 & 14,079 \\
\textbf{ARC-C} & 1,119 & 299 & 1,172 \\
\textbf{ARC-e} & 2,251 & 570 & 2,376 \\ \bottomrule
\end{tabular}}
\label{statistics}
\end{table}

The questions in MMLU are designed to require extensive world knowledge and expert-level reasoning, making it a rigorous benchmark for evaluating language models' generalization across multiple disciplines. We summarize the key statistics and characteristics of the MMLU dataset in Table \ref{tan:mmlu_arc}. 
The \textbf{AI2 Reasoning Challenge (ARC)} benchmark \citep{arc} is a comprehensive dataset designed to assess the ability of language models to answer complex, multi-faceted science questions on scientific reasoning and knowledge integration capabilities. The ARC dataset consists of $7,787$ multiple-choice questions derived from grade-school-level science exams, spanning grades $3$ through $9$. These questions are divided into two subsets: the Easy Set (ARC-E) and the Challenge Set (ARC-C), with the latter containing $2,590$ questions that are particularly difficult and require advanced reasoning skills.
The Easy Set (ARC-E) comprises $5,197$ questions that are relatively straightforward and can often be answered using basic retrieval or word co-occurrence methods. In contrast, the Challenge Set (ARC-C) includes questions that were specifically selected because they could not be correctly answered by retrieval-based algorithms (e.g., Information Retrieval Solver) or word co-occurrence methods (e.g., Pointwise Mutual Information Solver). These questions demand deeper comprehension, reasoning, and the integration of distributed knowledge across multiple sentences or concepts.
Each question in the ARC dataset is presented with four answer choices, with less than 1\% of questions having either three or five options. The dataset is further partitioned into training, validation, and test splits to facilitate model development and evaluation. For instance, the Challenge Set includes $1,119$ training examples, $299$ validation examples, and $1,172$ test examples.
We summarize the key statistics and characteristics of the ARC dataset in Table \ref{app:avg_hop}. 
\paragraph{Medical Domain Benchmarks} We use multiple-choice medical questions benchmarks in six languages as the representative knowledge-intensive domain, including MedQA (English and Chinese) \citep{medqa}, IgakuQA (Japanese) \citep{igaku}, RuMedDaNet \citep{rumed}, FrenchMedMCQA \citep{frenchmed}, and Head-QA \citep{headqa} to provide a comprehensive understanding of our KALE. We provide the statistics of each dataset in Table \ref{table:sta_datasets}.

\begin{table*}[htbp]
\centering
\caption{Statistical results for medical multiple-choice questions benchmarks in six languages.}
\resizebox{1.8\columnwidth}{!}{

\begin{tabular}{@{}llp{5cm}ll@{}}
\toprule
\textbf{Dataset}    & \textbf{Language} & \textbf{Source}                                            & \textbf{Train} & \textbf{Test} \\
\midrule
MedQA               & English    & United States Medical Licensing Examination               & 10178  & 1273  \\
MedQA               & Chinese    & United States Medical Licensing Examination               & 27400  & 3426  \\
IgakuQA             & Japanese   & Japan's medical licensure exams (2018-2022)         & 1590   & 199   \\
RuMedDaNet          & Russian    & Russian medical judgment question dataset           & 1052   & 256   \\
FrenchMedMCQA       & French     & Professional exams for the French Pharmacy degree   & 2171   & 622   \\
Head-QA             & Spanish    & Exams for positions in the Spanish healthcare       & 2657   & 2742  \\
\bottomrule
\end{tabular}}

\label{table:sta_datasets}
\end{table*}

\section{More In-depth Analysis of KALE}

\subsection{More Results of KALE on  Knowledge-intensive Domains} \label{app:domain}

In Tables \ref{tab:main_res} and \ref{tab:more_backbone3}, we present the performance of KALE across various downstream tasks. To further demonstrate the capabilities of KALE, this section provides its evaluation on several knowledge-intensive tasks. Following the same experimental setting of KG-SFT \citep{kg-sft}, we use MedQA as the benchmark using  LlaMA2 7B as the backbone model.  As shown in Table \ref{app:res_medqa}, we still observe that our proposed KALE significantly outperforms existing state-of-the-art baselines by a large margin, which also demonstrates that our KALE can effectively work under the knowledge-intensive scenarios.

\begin{table*}[htbp]
\centering
\caption{Experiment results for existing methods on knowledge-intensive domains. The results of the mentioned methods are taken from KG-SFT \citep{kg-sft}. We \textbf{bold} the best results for each dataset.}\label{app:res_medqa}
\resizebox{\linewidth}{!}{
\begin{tabular}{lcccccccc}
\toprule
Method & MedQA & MedQA  & IgakuQA & RuMedDaNet & MedMCQA & HeadQA & Average \\ 
     & (English)     & (Chinese) &  (Russian)     &  (Spanish)       &  (French)    &  (Japanese)    &   \\ \midrule
Vanilla       & 28.20   & 28.37   & 51.17   & 32.97   & 12.76    & 11.10    & 27.43   \\
COT  & 37.65 & 39.01 & 65.23 & 40.33 & 25.08 & 23.63 & 38.48 \\
TOG   & 34.27 & 28.13 & 48.42 & 35.59 & 12.47 & 19.61 & 29.75 \\
KGR  & 33.15 & 26.88 & 47.52 & 34.74 & 13.39 & 17.29 & 28.83 \\
KAPING  & 36.39 & 27.24 & 54.66 & 34.98 & 11.54 & 15.91 & 30.45 \\ 
\midrule
SFT       & 33.62   & 29.33   & 66.40   & 35.19   & 12.67    & 21.11    & 32.30   \\
AugGPT   & 40.29   & 36.54   & 62.14   & 40.70   & 22.99    & 27.13    & 38.30   \\
GPT3Mix  & 39.35   & 37.97   & 66.01   & 41.50   & 25.08    & 26.13    & 39.34   \\
KG-SFT   & {41.71} & {39.31} & {68.75} & {44.40} & {28.45} & {28.14} & {41.79} \\ 
\midrule
\blue{\textbf{KALE (ours)}}  & \blue{\textbf{45.89}} & \blue{\textbf{42.77}} & \blue{\textbf{69.81}} & \blue{\textbf{45.58}} & \blue{\textbf{30.39}} & \blue{\textbf{28.79}} & \blue{\textbf{43.53}} \\ 
\bottomrule
\end{tabular}}
\end{table*}

\subsection{Inference Time Comparison} \label{app:inference_time}

\begin{table*}[htbp]
\centering
\caption{Average testing time for each sample on the AbsR dataset for each method (Unit: second)}
\label{tab:absr_time}
\resizebox{1.8\columnwidth}{!}{
\begin{tabular}{lllllll}
\toprule
Backbone Models & Vanilla & CoT & TOG & StructGPT & GraphRAG & KALE (ours) \\
\midrule
LlaMa3 8B & 7.44 & 7.91 & 8.21 & 7.88 & 9.08 & 7.50 \\
Mistral 7B & 2.19 & 3.11 & 4.97 & 5.45 & 10.10 & 2.11 \\
Qwen2.5 32B & 11.20 & 11.90 & 11.8 & 12.8 & 12.30 &11.09 \\
Gemma2 9B & 3.73 & 4.19 & 4.82 & 3.98 & 8.40 & 3.93 \\
OLMOE 7B & 8.33 & 8.75 & 10.70 & 14.60 & 11.04 & 8.55 \\
Orca2 7B & 3.97 & 4.33 & 4.95 & 7.09 & 8.20 & 3.67 \\

\bottomrule
\end{tabular}}
\end{table*}

As mentioned in Section \ref{sec:rela_2}, KALE is a post-training method designed to enhance the knowledge manipulation capabilities of LLMs. \textbf{Once the model completes training, KALE maintains identical autoregressive inference characteristics to vanilla LLMs during the decoding phase, introducing zero additional temporal overhead and requiring no retrieval operations from external knowledge bases.} We conduct comparative measurements of average inference latency per sample across different methodologies (vanilla LLM, CoT, TOG, StructGPT, GraphRAG, and KALE) using an Nvidia A100 GPU (80GB). The quantitative results in Table \ref{tab:absr_time} reveal that KALE achieves nearly identical inference speed to vanilla LLMs. At the inference stage, both KALE and Vanilla models follow a similar logic: they directly take the instruction and question as input to the LLM. \textbf{Therefore, any observed speed differences between them are primarily attributable to slight variations in the length of their generated outputs.} There are instances where the Vanilla model's output length is marginally longer than KALE's, leading to KALE being slightly faster, and vice versa. This minor difference in token generation directly impacts the overall inference time. In contrast, RAG-based approaches requiring knowledge retrieval and CoT methods with extended prompt sequences incur additional computational overhead. 

\subsection{More results on the Hyperparameter sensitivity evaluation of KALE} \label{sec:hyper_sensi}

Regarding the sensitivity of KALE to the hyperparameters of each component, we conduct experiments to demonstrate its robustness. For all datasets, our default setting involved randomly sampling 10 anchor entities from their 3-hop neighbors. The consistent superior performance of KALE across diverse datasets under these unified parameter settings highlights its general effectiveness.

Moreover, to further investigate KALE's robustness, we conduct experiments by varying these key hyperparameters. As shown in Table \ref{tab:Hyperparameter}, we can observe that KALE exhibits robustness to changes in both the number of anchor entities and hops for neighbors. This further underscores the practical potential and reliability of our KALE framework in real-world applications.

\begin{table*}[htbp]
\centering
\caption{Hyperparameter sensitivity evaluation on the number of anchor entities and the hop of neighbors.}
\label{tab:Hyperparameter} 
\resizebox{\textwidth}{!}{%
\begin{tabular}{cccccccccc}
\toprule
Anchors &  Hops & Absr & ARC-c & ARC-e & Common & MMLU & BBH & RACE-h & RACE-m \\
\midrule
5 & 2 & 82.94 & 80.03 & 84.18 & 66.34 & 61.79 & \textbf{57.98} & 68.27 & 73.33 \\
5 & 4 & 84.50 & 78.50 & 84.13 & 63.64 & 61.33 & 57.82 & 66.60 & 71.73 \\
15 & 2 & 82.94 & 75.09 & 85.00 & 64.95 & 60.03 & 56.13 & 65.75 & 69.64 \\
15 & 4 & \textbf{85.31} & 76.19 & \textbf{87.40} & 65.44 & 60.68 & 54.45 & 65.41 & 71.03 \\
\blue{10 (ori)} & \blue{3(ori)} & \blue{83.62} & \blue{\textbf{81.23}} & \blue{86.45} & \blue{\textbf{65.69}} & \blue{\textbf{63.27}} & \blue{57.33} & \blue{\textbf{68.61}} & \blue{\textbf{74.12}} \\
\bottomrule
\end{tabular}%
}
\end{table*}

\subsection{Average Steps of Extracted Reasoning Paths}

\begin{table}[htbp]
\centering
 \caption{Statistics of average step in reasoning path on the AbsR dataset.} \label{app:avg_hop}
\begin{tabular}{cccc}
 \toprule
1-hop & 2-hop & 3-hop complete &3-hop partial \\
 \midrule
15.76 & 54.03 & 28.27 & 1.94 \\
\bottomrule
\end{tabular}

\end{table}

By default, we generate three-hop reasoning paths from questions to answers for each question-answer pair. If a 3-hop reasoning path cannot reach the answer entity, we still provide these paths as auxiliary information to facilitate rationale generation by the LLM. We currently provide the proportion of each hop within the generated reasoning paths, where '3hop complete' indicates that the three-hop reasoning path successfully reached the answer, and '3hop partial' indicates that the reasoning path did not reach the answer entity. As shown in Table \ref{app:avg_hop}, we find that most of the reasoning paths can directly lead to the final answer entity. Specifically, less than 2\% of the reasoning paths cannot reach the answer entity. This suggests that the extracted reasoning paths can effectively elucidate the underlying logic and correlations between the question and the answer.


\subsection{More Results of Different Backbone Models} \label{app:backbone}

\begin{table*}[!t]
\centering
\caption{More results of our KALE using Gemma$2$ $9$B, OLMOE $7$B, and Orca$2$ $7$B as backbone models. We \textbf{bold} the best results and \underline{underline} the suboptimal results for each backbone model.}
\label{tab:more_backbone3}
\resizebox{2\columnwidth}{!}{
\begin{tabular}{lllcccccccc}
\toprule
\textbf{Backbone}         & \textbf{Category} & \textbf{Method}    & \textbf{AbsR} & \textbf{ARC-c} & \textbf{ARC-e} & \textbf{Common} & \textbf{MMLU}  & \textbf{BBH}   & \textbf{RACE-h} & \textbf{RACE-m} \\ 
\midrule
\multirow{16}{*}{\textbf{Gemma2 9B}} 
    & \multirow{2}{*}{\textbf{Prompt-based}}  
        & \textbf{Vanilla}   & 52.49 & 79.95 & 88.89 & 57.66 & 53.56 & 48.93 & 73.13 & 78.62 \\
    &                        & \textbf{CoT}       & 67.54 & 81.06 & 86.91 & 61.43 & 57.35 & 53.37 & 71.07 & 79.32 \\
    \cmidrule{2-11}
    & \multirow{3}{*}{\textbf{Retrieval-based}} 
        & \textbf{TOG}       & 72.04 & 79.27 & 81.65 & 63.06 & 59.31 & 51.53 & 75.99 & 79.53 \\
    &                        & \textbf{StructGPT} & 59.24 & {83.28} & 86.87 & 59.30 & {61.40} & 57.98 & \underline{79.87} & 81.27 \\
    &                        & \textbf{GraphRAG} & 64.57 & \underline{85.07} & 84.13 & \underline{65.00} & \underline{62.53} & \underline{60.89} & 76.80 & 81.69 \\
    \cmidrule{2-11}
    & \multirow{5}{*}{\textbf{SFT-based}}    
        & \textbf{SFT}      & 61.37 & 81.06 & 89.06 & 58.97 & 55.26 & 51.38 & 74.93 & 80.78 \\
    &                        & \textbf{SDFT}     & 75.83 & 82.42 & 90.91 & 60.85 & 57.67 & 55.37 & 74.19 & 81.20 \\
    &                        & \textbf{DMT}      & \underline{77.13} & 81.83 & \underline{91.12} & 62.00 & 56.69 & 53.22 & 76.99 & 83.01 \\
    &                        & \textbf{MeanLearn} & 72.04 & 80.20 & 89.90 & 63.06 & 58.98 & 57.36 & 75.53 & 81.20 \\
    &                        & \textbf{KG-SFT} & 74.76 & 80.20 & 88.26 & 64.54 & 59.37 & 55.06 & 77.16 & 82.73 \\ 
    \cmidrule{2-11}
    & \multirow{4}{*}{\textbf{Augmented-based}} 
        & \textbf{STaR}      & 76.66 & 77.22 & 84.43 & 60.20 & 54.54 & 56.60 & 75.53 & 81.55 \\
    &                        & \textbf{AugGPT}   & 59.60 & 82.34 & 81.86 & 53.73 & 55.26 & {58.89} & 78.88 & 83.33 \\
    &                        & \textbf{GPT3Mix}   & 59.72 & 75.43 & 88.22 & {64.29} & 61.01 & 55.83 & 79.07 & \underline{83.98} \\
    \cmidrule{3-11}
    &                        & \blue{\textbf{KALE (ours)}} & \blue{\textbf{81.52}} &\blue{\textbf{88.57}} & \blue{\textbf{94.70}} & \blue{\textbf{68.63}} &\blue{\textbf{65.32}} & \blue{\textbf{65.49}} & \blue{\textbf{83.30}} & \blue{\textbf{87.74}} \\
\midrule
\multirow{16}{*}{\textbf{OLMOE 7B}} 
    & \multirow{2}{*}{\textbf{Prompt-based}}  
        & \textbf{Vanilla}   & 49.88 & 62.03 & 65.99 & 44.06 & 38.73 & 35.73 & 57.18 & 65.74 \\
    &                        & \textbf{CoT}      & 51.06 & 63.13 & 67.34 & 45.62 & 39.91 & 36.66 & 59.46 & 64.83 \\
    \cmidrule{2-11}
    & \multirow{3}{*}{\textbf{Retrieval-based}} 
        & \textbf{TOG}       & 54.50 & 64.42 & 69.82 & 47.26 & 40.89 & 38.34 & 60.35 & 67.75 \\
    &                        & \textbf{StructGPT}  & 56.87 & 65.70 & 71.12 & 51.60 & 41.61 & 40.64 & 60.03 & 69.63 \\
    &                        & \textbf{GraphRAG} & 57.82 & 60.75 & 71.25 & 50.61 & 43.50 & 41.56 & 60.66 & 68.04 \\
    \cmidrule{2-11}
    & \multirow{5}{*}{\textbf{SFT-based}}    
        & \textbf{SFT}        & 53.31 & 63.91 & 68.52 & 49.14 & 40.43 & 37.58 & 59.18 & 69.63 \\
    &                        & \textbf{SDFT}       & 59.95 & 65.52 & 70.16 & 50.61 & 42.78 & 38.65 & 59.06 & 67.84 \\
    &                        & \textbf{DMT}       & 60.43 & 66.04 & 70.83 & 51.26 & 42.36 & 39.57 & 61.18 & 68.45 \\
    &                        & \textbf{MeanLearn}  & \underline{71.09} & 66.30 & 67.80 & \underline{54.55} & \underline{44.21} & 43.10 & 60.03 & 72.42 \\
    &                        & \textbf{KG-SFT} & 61.26 & 66.41 & 70.58 & 52.66 & 43.17 & 38.04 & 61.09 & 65.25 \\ 
    \cmidrule{2-11}
    & \multirow{4}{*}{\textbf{Augmented-based}} 
        & \textbf{STaR}       & 59.24 & 66.12 & 71.04 & 50.36 & 43.76 & 41.41 & \underline{62.84} & 66.04 \\
    &                        & \textbf{AugGPT}    & 61.73 & 66.55 & 71.54 & 52.00 & 43.76 & \underline{43.40} & 60.98 & 70.19 \\
    &                        & \textbf{GPT3Mix}   & 62.20 & \underline{67.06} & \underline{72.60} & 53.23 & 43.50 & 42.02 & 60.26 & \underline{75.48} \\
    \cmidrule{3-11}
    &                        & \blue{\textbf{KALE (ours)}} & \blue{\textbf{81.99}} & \blue{\textbf{72.78}} & \blue{\textbf{74.60}} & \blue{\textbf{58.25}} & \blue{\textbf{46.96}} & \blue{\textbf{45.88}} & \blue{\textbf{64.35}} & \blue{\textbf{75.84}} \\
\midrule
\multirow{16}{*}{\textbf{Orca2 7B}} 
    & \multirow{2}{*}{\textbf{Prompt-based}}  
        & \textbf{Vanilla}   & 61.37 & 68.34 & 70.75 & 47.67 & 44.09 & 37.27 & 72.36 & 75.49 \\
    &                        & \textbf{CoT}        & 67.77 & 70.90 & 77.40 & 50.86 & 43.77 & 39.20 & 72.58 & 75.84 \\
    \cmidrule{2-11}
    & \multirow{3}{*}{\textbf{Retrieval-based}} 
        & \textbf{TOG}        & 59.60 & 73.89 & 75.72 & 62.24 & 51.14 & 42.94 & 73.41 & 74.09 \\
    &                        & \textbf{StructGPT}  & 65.17 & 67.66 & 77.95 & 53.40 & 45.40 & 46.01 & {76.01} & 78.41 \\
    &                        & \textbf{GraphRAG} & 67.06 & 69.97 & 78.87 & 54.71 & 50.75 & 47.70 & \underline{76.02} &75.77 \\  
    \cmidrule{2-11}
    & \multirow{5}{*}{\textbf{SFT-based}}    
        & \textbf{SFT}        & 63.98 & 71.33 & 76.56 & 48.24 & 52.90 & 47.70 & 73.33 & 76.88 \\
    &                        & \textbf{SDFT}       & 76.66 & 72.53 & 75.72 & 52.33 & 52.25 & 46.63 & 73.99 & 75.14 \\
    &                        & \textbf{DMT}        & 75.24 & 73.55 & 77.15 & 51.27 & 52.63 & {48.31} & 73.41 & 77.30 \\
    &                        & \textbf{MeanLearn}  & {77.01} & \underline{77.22} & \underline{86.57} & \underline{66.50} & {53.04} & 35.58 & 73.36 & \underline{78.76} \\
    &                        & \textbf{KG-SFT} & \underline{78.91} & 72.44 & 78.87 & 52.42 & \underline{54.00} & \underline{48.93} & 74.01 & 76.90 \\ 
    \cmidrule{2-11}
    & \multirow{4}{*}{\textbf{Augmented-based}} 
        & \textbf{STaR}       & 71.68 & 75.00 & 81.57 & 64.53 & 45.85 & 44.33 & 75.33 & 77.30 \\
    &                        & \textbf{AugGPT}     & 61.73 & 73.89 & 80.05 & 53.89 & 47.81 & 44.32 & 75.24 & 78.55 \\
    &                        & \textbf{GPT3Mix}    & 69.79 & 74.58 & 79.67 & 54.46 & 50.75 & 45.25 & 75.53 & 77.51 \\
    \cmidrule{3-11}
    &                        & \blue{\textbf{KALE (ours)}}  & \blue{\textbf{83.41}} & \blue{\textbf{78.16}} & \blue{\textbf{88.51}} & \blue{\textbf{69.62}} & \blue{\textbf{61.20}} & \blue{\textbf{50.77}} & \blue{\textbf{78.62}} & \blue{\textbf{80.02}} \\
\bottomrule
\end{tabular}
}
\end{table*}

\begin{table*}[t]
  \centering
  \caption{Results comparison of KALE using different KGs to extract reasoning path using Llama$3$ $8$B as the backbone model.}
  \label{tab:kale_kg_comparison}
  \resizebox{\textwidth}{!}{
  \begin{tabular}{lcccccccc}
    \toprule
     & \textbf{AbsR} &\textbf{ ARC-c} & \textbf{ARC-e} & \textbf{Common} & \textbf{MMLU}  & \textbf{BBH} & \textbf{RACE-h} & \textbf{RACE-m} \\
    \midrule
    \textbf{KALE$_{DBpedia}$} & 80.81 & 77.89 & 83.77 & 60.07 & 61.28 & \textbf{58.00} & 65.58 & 68.73 \\
    \textbf{KALE$_{ConceptNet}$}  & 79.93 & \textbf{81.54} & 84.19 & 62.03 & 61.06 & 55.94 & 66.93 & 71.17 \\
    \blue{\textbf{KALE$_{Wikidata}$}}  & \blue{\textbf{83.62}} & \blue{81.23} & \blue{\textbf{86.45}} & \blue{\textbf{65.69}} & \blue{\textbf{63.27}} & \blue{57.33} & \blue{\textbf{68.61}} & \blue{\textbf{74.21}} \\
    \bottomrule
  \end{tabular}
  }
\end{table*}

\begin{figure*}[t]
    \centering
    \begin{minipage}{\textwidth}
        \includegraphics[width=0.33\textwidth, height=0.29\textwidth]{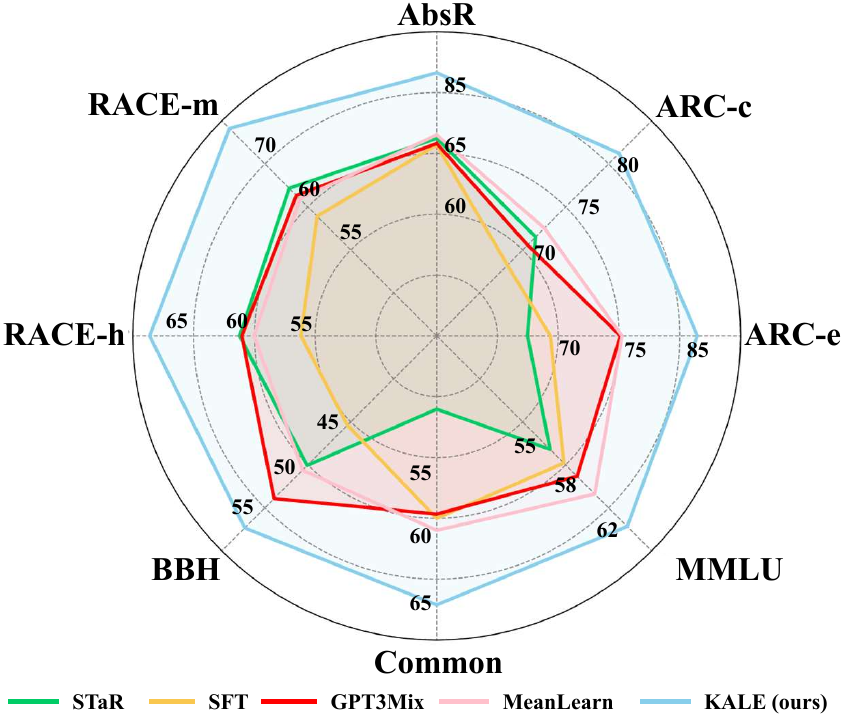}
        \includegraphics[width=0.33\textwidth, height=0.29\textwidth]{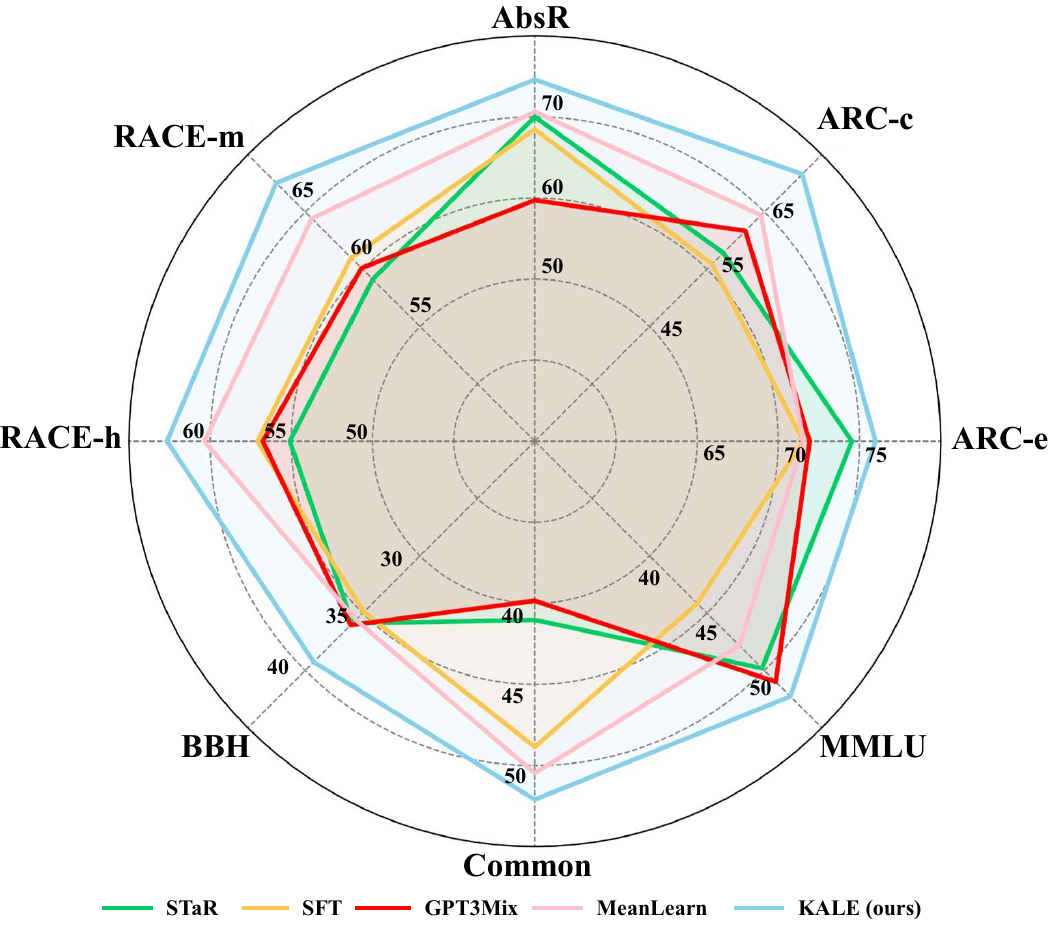}
        \includegraphics[width=0.33\textwidth, height=0.29\textwidth]{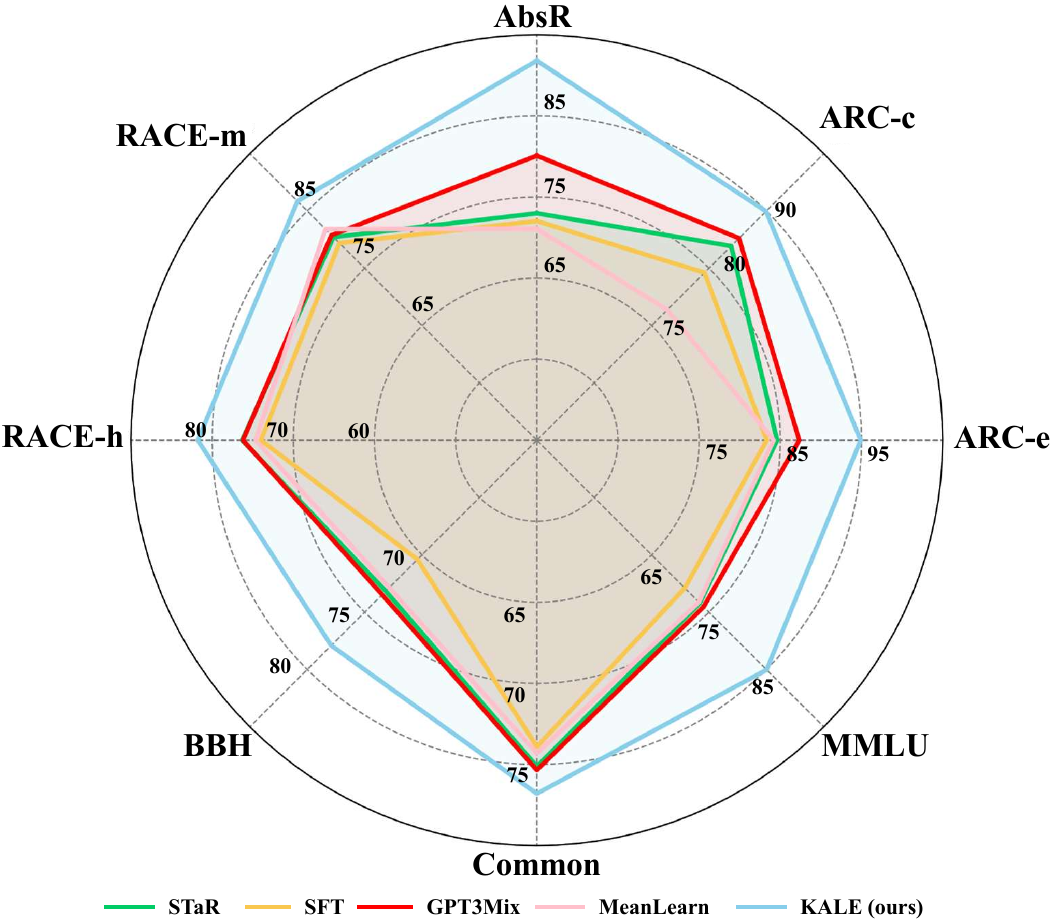}
        \caption{KALE achieves state-of-the-art performance on a broad range of scientific optimization tasks compared with existing methods, using LlaMA$3$ $8$B, Mistral $7$B, and Qwen2.5 $32$B as backbone models, respectively.}
        \label{fig:leida_qiansan}
    \end{minipage}
\end{figure*}

\begin{figure*}[t]
    \centering
    \begin{minipage}{\textwidth}
        \includegraphics[width=0.33\textwidth, height=0.29\textwidth]{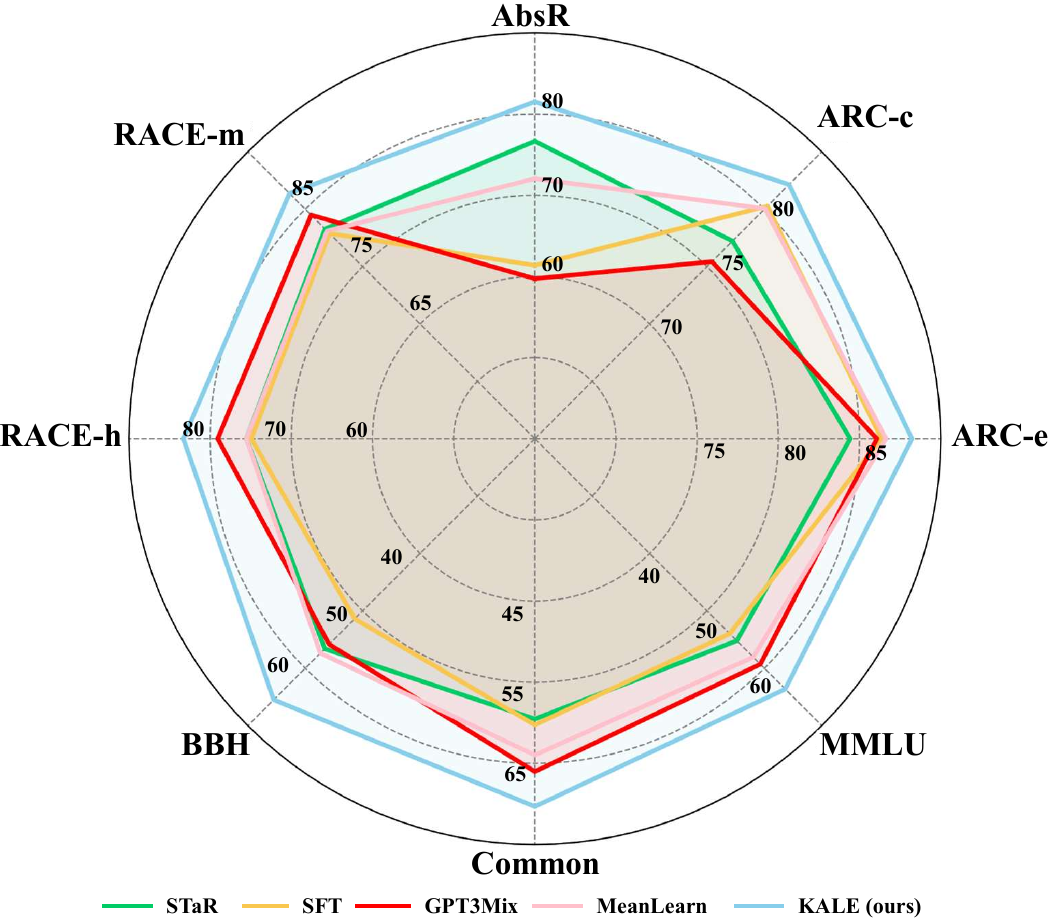}
        \includegraphics[width=0.33\textwidth, height=0.29\textwidth]{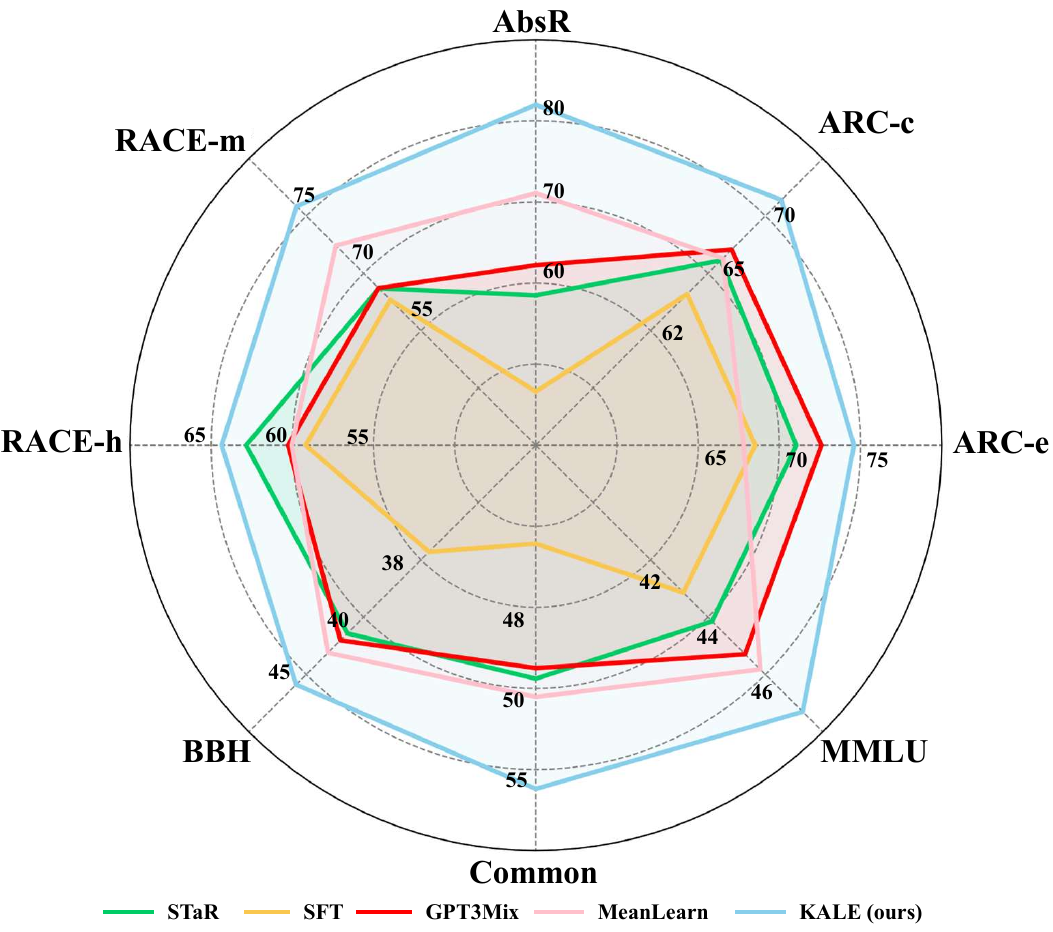}
        \includegraphics[width=0.33\textwidth, height=0.29\textwidth]{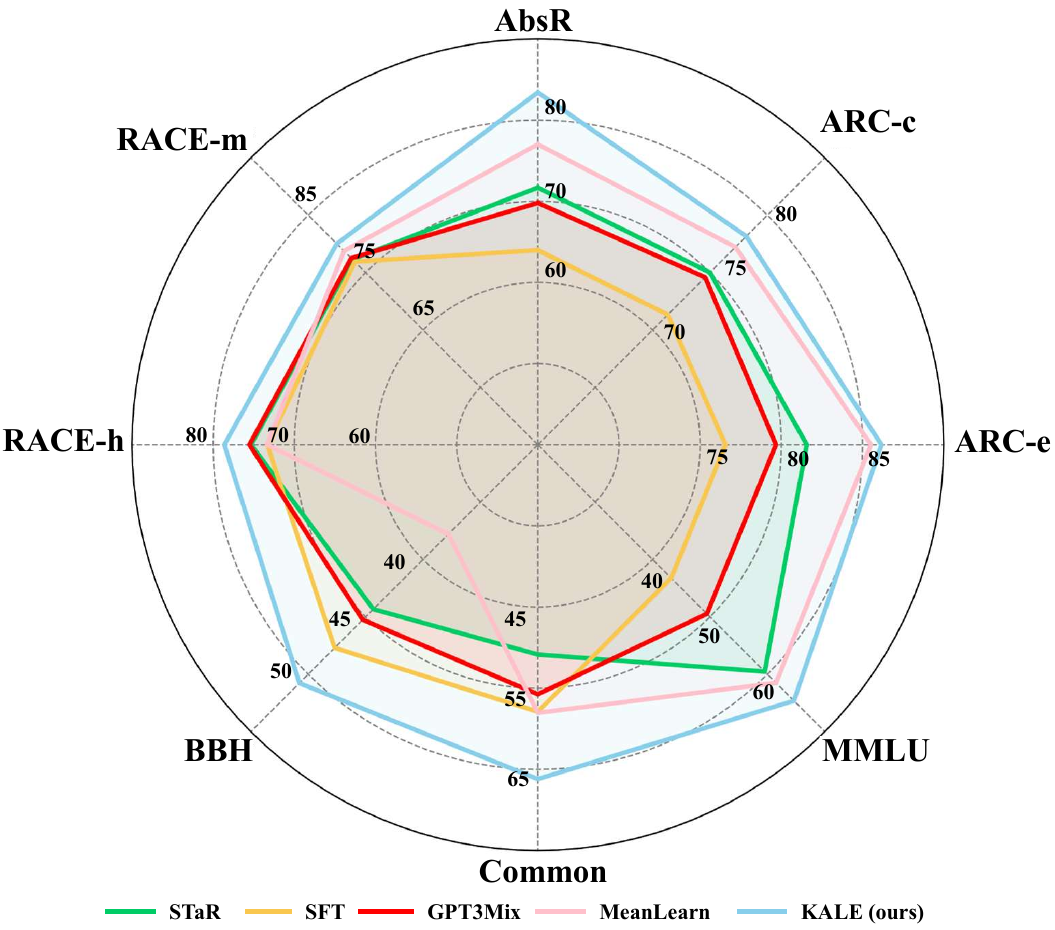}
        \caption{KALE achieves state-of-the-art performance on a broad range of scientific optimization tasks compared with existing methods, using Gemma$2$ $9$B, OLMOE $7$B, and Orca$2$ $7$B as backbone models, respectively.}
        \label{fig:leida_housan}
    \end{minipage}
\end{figure*}

As mentioned in Section \ref{sec:main_res}, we select LlaMA$3$ $8$B, Mistral $7$B, and Qwen2.5 $32$B as representative models in Table \ref{tab:main_res}. In this section, to further demonstrate the generalization and versatility of KALE, we also conducted experiments on several popular open-source LLMs, including Gemma$2$ $9$B, OLMOE-$1$B-$7$B, and Orca$2$ $7$B. As shown in Table \ref{tab:more_backbone3}, we can still observe that our KALE method significantly outperforms existing baselines on these backbone models as well. This further demonstrates the effectiveness of our KALE approach.
We also present radar charts for each backbone model to provide a more intuitive performance comparison in Figures \ref{fig:leida_qiansan} and \ref{fig:leida_housan}. The effectiveness of our KALE across various popular open-source models further demonstrates its strong versatility and generalization capabilities.

\subsection{More Results of Applying Different KGs to Extract Rationales} \label{sec:diff_kgs}

In the main experiments, we used Wikidata as the default KG for extracting reasoning paths. To further evaluate the robustness of KALE under different, smaller-scale KGs, we additionally extracted reasoning paths from alternative KGs and generated corresponding rationales. Specifically, we employed DBpedia \citep{dbpedia} and ConceptNet \citep{conceptnet} to extract reasoning paths, based on which we generated rationales for training. We used LLaMA3-8B as the backbone model. 

As shown in Table \ref{tab:kale_kg_comparison}, We observe that our KALE model exhibits relatively robust performance across different KGs. This implies a strong potential for KALE to generalize to various KGs in complex real-world datasets, thereby demonstrating its significant applicability in practical scenarios.

\subsection{More Results of Ablation Study} \label{app:abla}

\begin{table*}[!t]
\centering
\caption{More results of the ablation study of our KALE, using Gemma$2$ $9$B, OLMOE $7$B, and Orca$2$ $7$B as the backbone models.}
\label{tab_app:abla}
\resizebox{2\columnwidth}{!}{
\begin{tabular}{llllllllll}
\toprule
\textbf{Backbone}         & \textbf{Method}    & \textbf{AbsR} & \textbf{ARC-c} & \textbf{ARC-e} & \textbf{Common} & \textbf{MMLU}  & \textbf{BBH}   & \textbf{RACE-h} & \textbf{RACE-m} \\ 
\midrule
\multirow{3}{*}{\textbf{Gemma2 9B}} 
    & \textbf{KALE$_\mathrm{w/o \hspace{1mm} KI}$}   & 76.54$_{\color{red}{\downarrow4.98}}$ & 84.47$_{\color{red}{\downarrow4.10}}$ & 92.17$_{\color{red}{\downarrow2.53}}$ & 65.52$_{\color{red}{\downarrow3.11}}$ & 61.14$_{\color{red}{\downarrow4.18}}$ & 61.35$_{\color{red}{\downarrow4.14}}$ & 80.02$_{\color{red}{\downarrow3.28}}$ & 84.26$_{\color{red}{\downarrow3.48}}$   \\
    & \textbf{KALE$_\mathrm{w/o \hspace{1mm} KA}$}  & 73.22$_{\color{red}{\downarrow8.30}}$ & 78.41$_{\color{red}{\downarrow10.16}}$ & 90.32$_{\color{red}{\downarrow4.38}}$ & 66.99$_{\color{red}{\downarrow1.64}}$ & 63.42$_{\color{red}{\downarrow1.90}}$ & 60.12$_{\color{red}{\downarrow5.37}}$ & 78.70$_{\color{red}{\downarrow4.60}}$ & 82.66$_{\color{red}{\downarrow5.08}}$   \\
    & \blue{\textbf{KALE}} & \blue{\textbf{81.52}}&\blue{\textbf{88.57}} & \blue{\textbf{94.70}}& \blue{\textbf{68.63}} &\blue{\textbf{65.32}}& \blue{\textbf{65.49}} & \blue{\textbf{83.30}}& \blue{\textbf{87.74}}\\
\midrule
\multirow{3}{*}{\textbf{OLMOE 7B}} 
       & \textbf{KALE$_\mathrm{w/o \hspace{1mm} KI}$}    & 78.91$_{\color{red}{\downarrow3.08}}$ & 69.80$_{\color{red}{\downarrow2.98}}$ & 73.23$_{\color{red}{\downarrow1.37}}$ & 56.51$_{\color{red}{\downarrow1.74}}$ & 40.89$_{\color{red}{\downarrow6.07}}$ & 43.25$_{\color{red}{\downarrow2.63}}$ & 62.92$_{\color{red}{\downarrow1.43}}$ & 70.26$_{\color{red}{\downarrow5.58}}$   \\
    & \textbf{KALE$_\mathrm{w/o \hspace{1mm} KA}$}      & 74.17$_{\color{red}{\downarrow7.82}}$ & 68.26$_{\color{red}{\downarrow4.52}}$ & 70.92$_{\color{red}{\downarrow3.68}}$ & 55.28$_{\color{red}{\downarrow2.97}}$ & 44.35$_{\color{red}{\downarrow2.61}}$ & 42.48$_{\color{red}{\downarrow3.40}}$ & 60.26$_{\color{red}{\downarrow4.09}}$ & 69.22$_{\color{red}{\downarrow6.62}}$   \\
   & \blue{\textbf{KALE}} & \blue{\textbf{81.99}} & \blue{\textbf{72.78}} & \blue{\textbf{74.60}} & \blue{\textbf{58.25}} & \blue{\textbf{46.96}} & \blue{\textbf{45.88}} & \blue{\textbf{64.35}} &  \blue{\textbf{75.84}}         \\
\midrule
\multirow{3}{*}{\textbf{Orca2 7B}} 

    & \textbf{KALE$_\mathrm{w/o \hspace{1mm} KI}$}    & 79.68$_{\color{red}{\downarrow3.73}}$ & 76.37$_{\color{red}{\downarrow1.79}}$ & 84.18$_{\color{red}{\downarrow4.33}}$ & 67.81$_{\color{red}{\downarrow1.81}}$ & 58.59$_{\color{red}{\downarrow2.61}}$ & 48.31$_{\color{red}{\downarrow2.46}}$ & 74.96$_{\color{red}{\downarrow3.66}}$ & 77.99$_{\color{red}{\downarrow2.03}}$   \\
    & \textbf{KALE$_\mathrm{w/o \hspace{1mm} KA}$}   & 77.61$_{\color{red}{\downarrow5.80}}$ & 75.43$_{\color{red}{\downarrow2.73}}$ & 82.49$_{\color{red}{\downarrow6.02}}$ & 65.52$_{\color{red}{\downarrow4.10}}$ & 54.41$_{\color{red}{\downarrow6.79}}$ & 45.86$_{\color{red}{\downarrow4.91}}$ & 73.16$_{\color{red}{\downarrow5.46}}$ & 75.91$_{\color{red}{\downarrow4.11}}$   \\

    & \blue{\textbf{KALE}}  & \blue{\textbf{83.41}} & \blue{\textbf{78.16}} & \blue{\textbf{88.51}} & \blue{\textbf{69.62}} & \blue{\textbf{61.20}} & \blue{\textbf{50.77}} & \blue{\textbf{78.62}} & \blue{\textbf{80.02}}       \\
\bottomrule
\end{tabular}
}
\end{table*}
In Section \ref{sec:abla}, we report the results of the ablation study using LlaMA$3$ $8$B, Mistral $7$B, and Qwen2.5 $32$B as the backbone model. In this section, we will further present the results using Gemma$2$ $9$B, OLMOE $7$B, and Orca$2$ $7$B as backbone models to obtain more insights into the individual components constituting KALE across various backbone models. As illustrated in Tables \ref{tab_app:abla}, we still observe that the absence of each component within KALE leads to a decline in performance across diverse domains for almost all applied backbone models in all tested benchmarks, which further demonstrates that KALE organically integrates the knowledge-induced data synthesis method and knowledge-aware fine-tuning into a unified framework as well. We still observe that the absence of knowledge-aware fine-tuning (\textbf{KALE$_\mathrm{w/o \hspace{1mm} KA}$}) leads to a more significant decline in accuracy, which further demonstrates the importance of effectively implicit knowledge learning.

\subsection{More Results of \textsc{Known\&Incorrect} Phenomenon on Different Baselines}\label{app:known}

As mentioned in Section \ref{exp:case}, models fine-tuned using vanilla SFT still exhibit a serious known-incorrect phenomenon. In this section, we provide more analysis of the known-incorrect phenomenon to include additional baselines involving the training of LLMs. 
As shown in Table \ref{tab:app_knowledge_manipulation}, we observe that our KALE consistently achieves the best results in knowledge manipulation analysis. If the model possesses relevant knowledge, KALE exhibits the lowest \textit{known\&incorrect} rate. Specifically, on Qwen-2.5 32B, KALE demonstrates only a 1.07\% \textit{known\&incorrect} rate. This  further indicates that KALE effectively enhances LLMs' knowledge manipulation ability in downstream tasks.


\begin{table*}[htbp]
\caption{Experiment results on the AbsR benchmark in {six} LLM backbones range for the knowledge manipulation analysis. We \textbf{bold} the best results for each method.}
\label{tab:app_knowledge_manipulation}
\centering
\resizebox{\linewidth}{!}{
\begin{tabular}{llcccccc}
\toprule
Category & Method & LlaMA3 8B & OLMOE 7B & Qwen2.5 32B & Gemma2 9B & Mistral 7B & Orca2 7B \\
\midrule
\multirow{9}{*}{\textit{Known\&Correct}}
& SFT  & 34.95 & 39.33 & 48.34 & 41.11 & 50.12 & 47.88 \\
& SDFT  & 56.28 & 47.87 & 53.00 & 58.89 & 54.03 & 55.57 \\
& DMT  & 56.64 & 44.91 & 65.17 & 61.85 & 51.78 & 57.35 \\
& Meanlearn  & 48.43 & 59.12 & 60.19 & 48.93 &56.52 & 59.12 \\
& KG-SFT  & 59.83 & 50.36 & 67.06 & 55.33 & 59.36 & 62.90 \\
& STaR  & 48.93 & 45.97 & 58.41 & 51.09 & 56.75 & 59.60 \\
& AugGPT  & 47.27 & 45.97 & 65.76 & 43.48 & 43.01 & 48.93 \\
& GPT3Mix & 54.15 & 48.34 & 62.90 & 42.30 & 43.84 & 56.52 \\
& \blue{\textbf{KALE}} & \blue{\textbf{82.94}} & \blue{\textbf{79.86}} & \blue{\textbf{87.56}} & \blue{\textbf{77.01}} & \blue{\textbf{71.09}} & \blue{\textbf{75.00}} \\
\midrule
\multirow{9}{*}{\textit{Known\&Incorrect}}
& SFT  & 28.43 & 44.08 & 27.49 & 35.55 & 29.50 & 35.90 \\
& SDFT  & 19.87 & 12.08 & 20.73 & 16.79 & 19.79 & 21.09 \\
& DMT  & 17.93 & 15.52 & 10.07 & 15.28 & 21.44 & 17.89 \\
& Meanlearn  & 22.75 & 11.97 & 10.90 & 23.11 & 14.45 & 17.89 \\
& KG-SFT  & 18.37 & 10.90 & 11.85 & 19.43 & 13.03 & 16.00 \\
& STaR  & 21.02 & 13.27 & 14.58 & 24.76 & 13.27 & 12.08 \\
& AugGPT  & 17.18 & 15.76 & 13.15 & 16.12 & 22.27 & 12.80 \\
& GPT3Mix & 14.12 & 13.86 & 17.20 & 17.42 & 15.88 & 13.27 \\
& \blue{\textbf{KALE}} & \blue{\textbf{4.15}} & \blue{\textbf{2.37}} & \blue{\textbf{1.07}} & \blue{\textbf{2.13}} & \blue{\textbf{7.7}} & \blue{\textbf{8.06}} \\
\bottomrule
\end{tabular}
}
\end{table*}


\subsection{More Results of SFT and KALE with Varying Ratios of Training Data} \label{app:case_ratio}

\begin{figure}[htbp]
\centering
\includegraphics[width=1.0\linewidth]{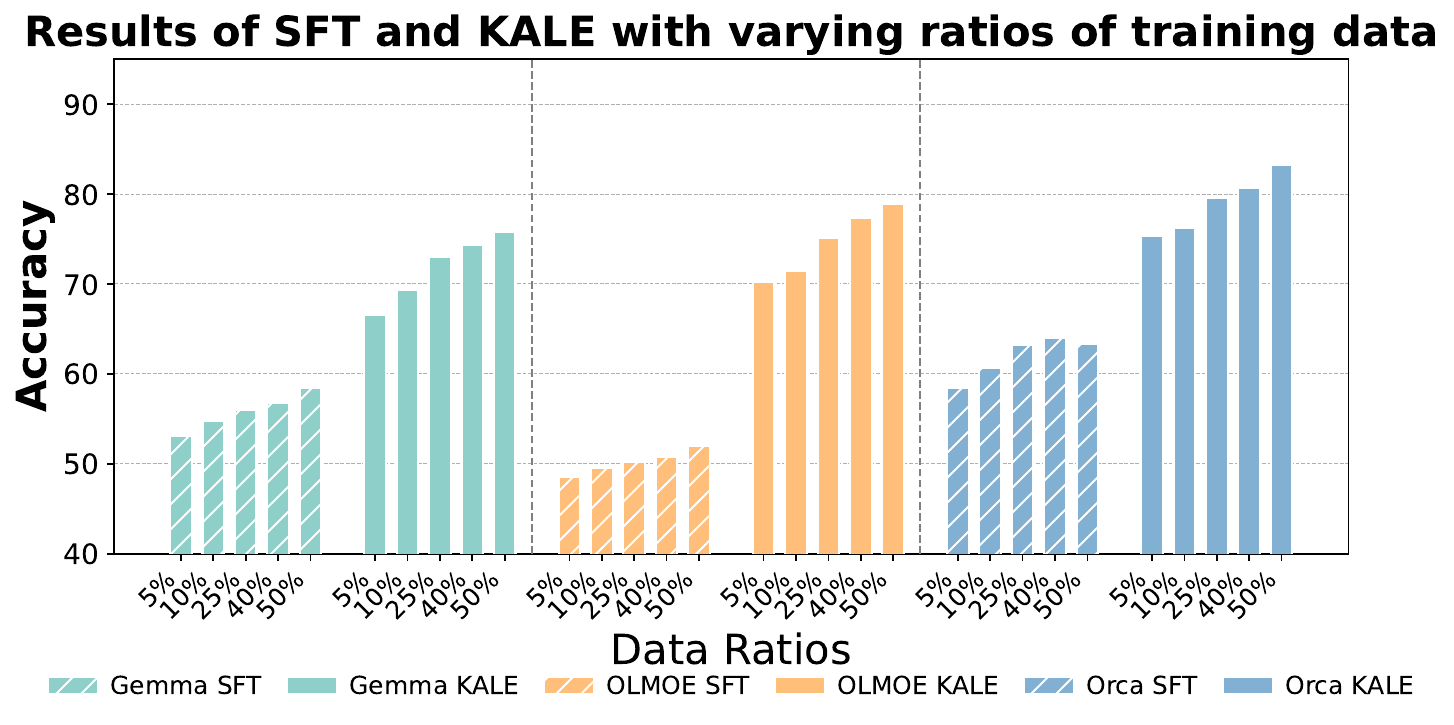} 
\caption{Results of different ratios of augmented rationales on SFT and KALE on Gemma2 9B, OLMOE 7B, and Orca2 7B, respectively.}
\label{app:ratio_app}
\end{figure}

In Section \ref{exp:case}, we utilized LlaMA3 8B, Mistral 7B, and Qwen2.5 32B as backbone models to investigate the performance of models trained with the SFT and KALE methods on downstream tasks under varying ratios of training rationales. 

In this section, we provide additional results using other LLMs as backbone models, including Gemma2 9B, OLMOE 7B, and Orca2 7B.
As shown in Figure \ref{app:ratio_app}, we observed that KALE demonstrated superior performance on downstream tasks even when only a small proportion of rationales was used for training. Specifically, the improvement of the OLMOE model can reach over 20\% on low-data scenarios. These findings highlight the effectiveness of KALE in low-resource scenarios, which also impies a great potentials of our KALE for scenarios with limited high-quality data.

\begin{table*}[htb]
\caption{Experiment results on the AbsR benchmark in {six} LLM backbones range from different data ratios. We  \textbf{bold} the better results for each backbone model.}
\label{main_res_transposed}
\centering
\resizebox{\linewidth}{!}{
\begin{tabular}{@{}c*{12}{c}@{}}
\toprule
  & \multicolumn{2}{c}{LlaMA3 8B} & \multicolumn{2}{c}{Mistral 7B} & \multicolumn{2}{c}{Qwen2.5 32B} & \multicolumn{2}{c}{Gemma2 9B} & \multicolumn{2}{c}{OLMOE-1B-7B} & \multicolumn{2}{c}{Orca2 7B} \\ 
\cmidrule(lr){2-3} \cmidrule(lr){4-5} \cmidrule(lr){6-7} \cmidrule(lr){8-9} \cmidrule(lr){10-11} \cmidrule(lr){12-13}
 \% Data & SFT & KALE & SFT & KALE & SFT & KALE & SFT & KALE & SFT & KALE & SFT & KALE \\ 
\midrule
5\%     & 63.98 & \textbf{74.88} & 63.03 & \textbf{66.35} & 67.06 & \textbf{82.78} & 53.12 & \textbf{66.60} & 48.58 & \textbf{70.22} & 58.39 & \textbf{75.31} \\
10\%    & 65.17 & \textbf{75.32} & 63.86 & \textbf{67.28} & 67.65 & \textbf{85.53} & 54.73 & \textbf{69.31} & 49.53 & \textbf{71.49} & 60.63 & \textbf{76.20} \\
25\%    & 65.76 & \textbf{78.31} & 65.40 & \textbf{67.79} & 67.89 & \textbf{86.01} & 55.98 & \textbf{73.00} & 50.19 & \textbf{75.07} & 63.17 & \textbf{79.55} \\
40\%    & 66.35 & \textbf{81.89} & 66.23 & \textbf{70.38} & 68.48 & \textbf{89.36} & 56.75 & \textbf{74.39} & 50.71 & \textbf{77.38} & 63.99 & \textbf{80.69} \\
50\%    & 66.94 & \textbf{82.93} & 66.94 & \textbf{74.11} & 69.31 & \textbf{90.22} & 58.44 & \textbf{75.75} & 52.01 & \textbf{78.91} & 63.35 & \textbf{83.21} \\
\bottomrule
\end{tabular}
}
\end{table*}


\subsection{More Results of Rationales Generated by Different LLMs}\label{sec:diff_llm}

In our main experiments, we utilize GPT-4o to generate rationales for each sample. We choose GPT-4o due to its exceptional performance in generating high-quality rationales, as it has demonstrated impressive results on numerous understanding and reasoning tasks. To demonstrate the generalizability of our KALE, we also incorporate two popular open-source LLMs—i.e., DeepSeek V3 and LLaMA3.1-70B-Instruct—for rationale generation and apply LLaMA3 8B as the backbone model. The results in Table \ref{tab:diff_llm_backbone} indicate that training on rationales generated by LLaMA3 70B and DeepSeekV3 still yields performance that significantly surpasses vanilla methods and achieves results comparable to those derived from GPT-4.0-generated rationales. \textbf{This demonstrates that KALE is relatively robust to rationales generated by different LLMs, highlighting its effectiveness for practical applications.}

\begin{table*}[ht] 
  \centering
  \caption{Results of KALE for rationales generated by different LLMs.}
  \label{tab:diff_llm_backbone} 
  \resizebox{\textwidth}{!}{ 
    \begin{tabular}{lcccccccc}
      \toprule
      \textbf{Method} & \textbf{AbsR} & \textbf{ARC-c} & \textbf{ARC-e} & \textbf{Common} & \textbf{MMLU} & \textbf{BBH} & \textbf{RACE-h} & \textbf{RACE-m} \\
      \midrule
      Vanilla & 62.68 & 66.79 & 69.90 & 58.72 & 55.88 & 46.54 & 53.35 & 57.02 \\
      KALE\_DeepSeek V3 & 82.70 & \textbf{81.48} & \textbf{86.70} & 64.70 & 62.25 & \textbf{58.13} & 64.69 & 71.03 \\
      KALE\_Llama3 70B & 78.91 & 77.56 & 84.05 & 63.72 & 60.03 & 54.45 & 65.52 & 69.78 \\
      \blue{KALE\_GPT-4o (Original)} & \blue{\textbf{83.62}} & \blue{81.23} & \blue{86.45} & \blue{\textbf{65.69}} & \blue{\textbf{63.27}} & \blue{57.33} & \blue{\textbf{68.61}} & \blue{\textbf{74.12}} \\
      \bottomrule
    \end{tabular}
  }
\end{table*}

\subsection{More Results of Different Types of Generated Rationales}\label{sec:diff_type_rat}

To further investigate whether the model genuinely benefits from meaningful knowledge or merely from the presence of any rationale, we generate two sets of modified rationales based on the original reasoning paths:

\begin{itemize}
    \item  KALE$_{irrelated}$: We instruct GPT-4o to generate factually irrelevant rationales to the reasoning paths.

    \item  KALE$_{contrast}$: We instruct GPT-4o to generate rationales that are factually contrasting to the reasoning paths.
\end{itemize}

We denote our original method as KALE$_{ori}$
 and present the comparative results in Table \ref{tab:generated_types}. The performance obtained using irrelevant or contrasting rationales is significantly lower than that of KALE$_{ori}$. This demonstrates the effectiveness of our knowledge-induced data synthesis, confirming that the model truly benefits from high-quality, factually accurate, and logically coherent rationales.

\begin{table*}[htbp]
\centering
\caption{Comparion of KALE on different types of generated rationales.}
\label{tab:generated_types} 
\resizebox{\textwidth}{!}{%
\begin{tabular}{lcccccccc}
\toprule
& \textbf{AbsR} & \textbf{ARC-c} & \textbf{ARC-e} & \textbf{Common} & \textbf{MMLU} & \textbf{BBH} & \textbf{RACE-h} & \textbf{RACE-m} \\
\midrule
KALE$_{irrelated}$ & 65.05 & 64.76 & 66.20 & 56.18 & 54.21 & 47.55 & 50.03 & 54.11 \\
KALE$_{contrast}$ & 59.60 & 59.64 & 63.56 & 51.84 & 52.25 & 42.48 & 51.11 & 48.96 \\
\blue{\textbf{KALE$_{ori}$}} & \blue{\textbf{83.62}} & \blue{\textbf{81.23}} & \blue{\textbf{86.45}} & \blue{\textbf{65.69}} & \blue{\textbf{63.27}} & \blue{\textbf{57.33}} & \blue{\textbf{68.61}} & \blue{\textbf{74.12}} \\
\bottomrule
\end{tabular}%
}
\end{table*}

\subsection{More Results on Reasoning Trace Quality Evaluation of Generated Rationales} \label{sec:rat_quality}

We incorporate a reasoning trace quality metric to evaluate the quality of the generated rationales to provide more insight into our KALE. We assess rationale quality across five critical dimensions: \textbf{Factual Accuracy, Logical Validity, Coherence, Completeness, and Interpretability} \citep{trace_quality}. Each dimension is evaluated as a binary classification task. Following the "LLM-as-a-judge" paradigm \citep{vicuna}, we utilize GPT-5 for this assessment. As shown in Table \ref{tab:quality_metric}, we use the AbsR dataset as an example and find that rationales generated via our KALE exhibit strong performance across all five dimensions, which further validates the effectiveness of KALE. 

\begin{table*}[htbp]
\centering
\caption{Reasoning trace quality evaluation for rationales on the AbsR dataset via GPT-5.} 
\label{tab:quality_metric} 
\begin{tabular}{ccccc}
\toprule
\textbf{Factual Accuracy} & \textbf{Logical Validity} & \textbf{Coherence} & \textbf{Completeness} & \textbf{Interpretability} \\
\midrule
98.82 & 97.63 & 99.53 & 100.00 & 99.64 \\
\bottomrule
\end{tabular}
\end{table*}


\subsection{More Results of Combining SFT with KALE}\label{sec:combine}

We also conduct an additional experiment using Llama3 8B as the backbone model. We compare two approaches: our original KALE method (denoted as KALE$_{ori}$) and a sequential approach where the model is first fine-tuned with SFT and then further trained with KALE (denoted as KALE$_{joint}$) 

As shown in Table \ref{tab:joint}, we find that while combining SFT first with KALE (KALE$_{joint}$) yields improvements only on some datasets. This presents a promising avenue for future work to thoroughly explore the optimal integration of KALE with existing post-training methods to achieve more consistent and significant performance enhancements for specific downstream domains.

\begin{table*}[htbp]
\centering
\caption{Comparison results of combining SFT with KALE pipeline using Llama3 8B as the backbone model.}
\label{tab:joint} 
\resizebox{\textwidth}{!}{%
\begin{tabular}{lcccccccc}
\toprule
&\textbf{AbsR} & \textbf{ARC-c} & \textbf{ARC-e} & \textbf{Common} & \textbf{MMLU} & \textbf{BBH} & \textbf{RACE-h} & \textbf{RACE-m} \\
\midrule
KALE$_{joint}$ & 80.21 & \textbf{82.25} & 84.18 & 61.34 & 60.42 & 53.22 & 67.98 & 72.91 \\
\blue{KALE$_{ori}$} & \blue{\textbf{83.62}} & \blue{81.23} & \blue{\textbf{86.45}} & \blue{\textbf{65.69}} & \blue{\textbf{63.27}} & \blue{\textbf{57.33}} & \blue{\textbf{68.61}} & \blue{\textbf{74.12}} \\
\bottomrule
\end{tabular}%
}
\end{table*}

\subsection{More Results of KALE on Open-ended Generations}

Our primary objective in KALE is to improve LLM performance on knowledge manipulation tasks, which are essential for enabling models to reason and respond accurately based on existing factual and procedural knowledge, such as in mathematics and multi-hop QA \citep{know_mani}. Consequently, our core evaluations primarily focus on QA-style benchmarks that reflect these capabilities. 

In this section, \textbf{we investigate KALE's general abilities, particularly in open-ended generation, which requires fluency, coherence, and creativity.} To this end, we conduct an additional evaluation on open-ended generation using MT-Bench~\citep{vicuna}, a benchmark suite that covers a broad range of tasks, including creative and free-form responses. We compare KALE, SFT, and vanilla models instantiated on the Llama 3 8B backbone, and we employ a GPT-5 LLM-Judge for automatic evaluation.

When evaluating with MT-Bench, we perform pairwise comparisons between the KALE model and each of the vanilla and SFT models separately. To mitigate position bias~\citep{vicuna}, the evaluation protocol for each \emph{(question, Model A, Model B)} instance is defined as follows:
\begin{enumerate}
    \item We first evaluate using the response order $(A, B)$.
    \item We then swap the responses and evaluate again using the order $(B, A)$.
\end{enumerate}
We only count a ``win'' for a model if both judgments agree on the same winner. If the two judgments yield conflicting outcomes, we record the result as a ``tie.'' We report the corresponding proportions of win rate, loss rate, and tie rate.

The results in Table~\ref{tab:openended_kale} show that the \textbf{KALE-trained model consistently outperforms both the vanilla model and the SFT baseline in terms of open-ended generation quality, achieving a higher win rate under the LLM-based evaluation.} \textbf{This indicates that KALE not only preserves but can also enhance the model's general generation abilities. Overall, these findings suggest that KALE does not harm, and may in fact improve, general fluency and creativity.}

\begin{table*}[t]
\centering
\caption{Comparison of vanilla, SFT, and KALE models on open-ended generation settings.}
\resizebox{0.8\textwidth}{!}{%
\begin{tabular}{lcccccc}
\toprule
& \multicolumn{3}{c}{Vanilla} & \multicolumn{3}{c}{SFT} \\
\cmidrule(lr){2-4} \cmidrule(lr){5-7}
Model & Win Rate & Loss Rate & Tie Rate & Win Rate & Loss Rate & Tie Rate \\
\midrule
KALE & 82.5 & 6.25 & 11.25 & 77.5 & 8.75 & 13.75 \\
\bottomrule
\end{tabular}}
\label{tab:openended_kale}
\end{table*}

\subsection{More Results of KALE on Thinking Style Models}

We also conduct additional experiments evaluating our KALE pipeline on Qwen3-32B (thinking)~\citep{Qwen3}: a state-of-the-art model with enhanced multi-step reasoning capabilities. As shown in Table \ref{tab:kale-qwen3-32b}, the KALE pipeline continues to provide significant improvements over vanilla and SFT methods when applied to these newer LLMs. This supports our claims regarding the robustness and generalizability of our KALE across different model backbones and reasoning capabilities.

\begin{table*}[t]
\centering
\caption{Results of KALE on Qwen3-32B-thinking model.}
\label{tab:kale-qwen3-32b}
\resizebox{\textwidth}{!}{%
\begin{tabular}{lcccccccc}
\toprule
Method   & AbsR   & ARC-c  & ARC-e  & Common & MMLU   & BBH    & RACE-h & RACE-m \\
\midrule
Vanilla  & 82.33  & 81.06  & 84.60  & 71.99  & 83.61  & 87.38  & 77.47  & 81.48  \\
SFT      & 85.82  & 84.22  & 87.54  & 74.53  & 85.56  & 88.73  & 80.05  & 84.57  \\
\textbf{KALE} & \textbf{93.60} & \textbf{91.09} & \textbf{93.43} & \textbf{79.08} & \textbf{89.44} & \textbf{92.02} & \textbf{84.61} & \textbf{88.54} \\
\bottomrule
\end{tabular}}
\end{table*}
\subsection{More Results of KALE on Self-taught Settings}
To validate the quality of our generated rationales in a self-taught setting, we conduct an additional experiment using LLaMA3-8B as both the teacher and student, as shown in Table \ref{tab:self_taught_kale_llama3_8b}. Even under this constrained setting, KALE consistently improves performance, highlighting that the reasoning paths synthesized from KI are of sufficient quality to facilitate learning even without access to a powerful teacher.

\begin{table*}[t]
  \centering
  \caption{Results of self-taught KALE using Llama3 8B as the backbone model.}
  \label{tab:self_taught_kale_llama3_8b}
  \resizebox{\textwidth}{!}{%
  \begin{tabular}{lcccccccc}
    \toprule
    Method               & AbsR      & ARC-c     & ARC-e     & Common    & MMLU      & BBH       & RACE-h    & RACE-m    \\
    \midrule
    Vanilla              & 62.68     & 66.79     & 69.90     & 58.72     & 55.88     & 46.54     & 53.35     & 57.02     \\
    \textbf{KALE$_{self \hspace{1mm} taught}$} & \textbf{75.31} & \textbf{73.19} & \textbf{80.08} & \textbf{60.84} & \textbf{58.59} & \textbf{52.11} & \textbf{61.19} & \textbf{67.57} \\
    \bottomrule
  \end{tabular}}
\end{table*}

\begin{table*}[t]
\centering
\caption{Comparison of GRPO and KALE using Llama3 8B as the backbone model.}
\label{tab:grpo_kale_llama3_8b}
  \resizebox{\textwidth}{!}{%
\begin{tabular}{lcccccccc}
\toprule
Method & AbsR & ARC-c & ARC-e & Common & MMLU & BBH & RACE-h & RACE-m \\
\midrule
GRPO & 75.83 & 75.77 & 73.53 & 63.64 & 59.37 & 51.07 & 63.32 & 68.87 \\
\textbf{KALE} & \textbf{83.62} & \textbf{81.23} & \textbf{86.45} & \textbf{65.69} & \textbf{63.27} & \textbf{57.33} & \textbf{68.61} & \textbf{74.12} \\
\bottomrule
\end{tabular}}
\end{table*}

\subsection{More Results of KALE on Comparison of GRPO}

Recently, methods such as GRPO~\citep{grpo} represent a compelling alternative by relying solely on questions and ground truth answers, thereby bypassing the need for high-quality intermediate textual reasoning annotations—the central challenge KALE is designed to address.

To investigate this, we have conducted additional experiments using GRPO with LLaMA3-8B as the backbone, implemented via LLaMAFactory~\citep{llamafactory}. The results in Table \ref{tab:grpo_kale_llama3_8b} demonstrate that GRPO, when directly trained on QA samples, still underperforms compared to our proposed KALE pipeline, especially on the ARC-e dataset, which leads to 12.92\%. This empirical evidence further supports the effectiveness of KALE in enhancing the reasoning capabilities of LLMs.

Moreover, we note that outcome-based RL approaches often suffer from lower training efficiency. Specifically, RL methods require extensive sample rollouts and typically operate under sparse reward settings. This results in a high computational burden during training. In contrast, the KI component of KALE utilizes an efficient multi-path A* algorithm, while the KA component optimizes a sample-level distribution loss that provides dense supervision signals, leading to significantly more efficient training.  

We also note that KALE and RL-based methods are not mutually exclusive. KALE is designed to mitigate the "known and incorrect" phenomenon commonly observed in SFT models, which tend to overfit to explicit input-output mappings and fail to dynamically retrieve and manipulate task-relevant knowledge~\citep{know_mani}. KALE instead enhances the model's knowledge manipulation capabilities—specifically, its ability to recall, reason, and transfer relevant knowledge effectively.

On the other hand, RL methods often aim to elicit emergent reasoning capabilities by leveraging trial-and-error dynamics, sometimes leading to "aha moments" in reasoning. From an engineering standpoint, one could further fine-tune a KALE-pretrained checkpoint using GRPO (or similar RL techniques) to improve performance on specific downstream tasks such as long-chain-of-thought reasoning or self-correction. 

We believe that integrating KALE into a broader LLM post-training framework that includes RL techniques is a promising direction for future research. Such a unified approach may yield deeper insights into the post-training landscape and further unlock the full potential of knowledge-aware language modeling.

\subsection{More Results of KALE on Comparison of Prompt Distillation}

Our proposed KA module utilizes rationales only during training by distilling the behavior induced by rationale-augmented inputs into models that do not require rationales at inference time. \textbf{This process is akin to knowledge distillation, where the model learns to emulate the internal reasoning patterns induced by rationales, without depending on them at test time.} Specifically, we use the rationale-enhanced model distribution as a soft target and train the model with an additional KL divergence loss to align the rationale-free predictions during test time. This approach allows the model to internalize the reasoning path during training and retrieve task-relevant knowledge during inference, even in the absence of explicit rationales. A variety of knowledge distillation approaches have investigated this paradigm~\citep{knowtuning, kar, knife, scott}. Therefore, we select and compare some recent popular knowledge distillation  method, including Distilling Step-by-Step (abbreviated as Distill-SBS)~\citep{dissbs}, PD~\citep{prompt_d}, and CoT distillation (abbreviated as COT-Distill)~\citep{cotdis}  to provide a more comprehensive understanding of KALE. To ensure a fair comparison, we adopt the standard settings of the PD method. For both CoT-Distill and Distill-SBS, we consistently use GPT-4o as the teacher model. Regarding CoT-Distill, we report the results obtained at the optimal granularity level. As shown in Table \ref{tab:pd_kale_llama3_8b}, our KALE consistently outperforms this prior method (referred to as prompt distillation (PD)), which also demonstrates the effectiveness of our KALE. We also emphasize that our primary objective lies not in distilling a compact model from a superior teacher, but rather in enhancing the model's knowledge manipulation capabilities via our KALE framework. From this perspective, our approach is orthogonal to distillation-based methods, and investigating their synergistic integration represents a promising avenue for future research.

\begin{table*}[t]
    \centering
    \caption{Comparison of PD and KALE using Llama3 8B as the backbone model.}
    \label{tab:pd_kale_llama3_8b}
      \resizebox{\textwidth}{!}{%
    \begin{tabular}{lcccccccc}
        \toprule
        Method & AbsR & ARC-c & ARC-e & Common & MMLU & BBH & RACE-h & RACE-m \\
        \midrule
        Distill-SBS       & 77.61 & 74.32 & 75.25 & 60.85 & 58.72 & 56.75 & 60.29 & 63.16 \\
        PD        & 75.31 & 72.17 & 76.98 & 57.73 & 60.17 & 55.41 & 58.89 & 68.91 \\
        COT-Distill        & 80.64 & 77.75 & 82.92 & 61.64 & 60.94 & 54.14 & 65.50 & 69.57 \\

        \textbf{KALE} & \textbf{83.62} & \textbf{81.23} & \textbf{86.45} & \textbf{65.69} & \textbf{63.27} & \textbf{57.33} & \textbf{68.61} & \textbf{74.12} \\
        \bottomrule
    \end{tabular}}
\end{table*}

\textbf{We also want to clarify that KALE is not in conflict with any distillation methods.} KALE is a flexible framework that provides a high-quality rationale generation approach and an effective training method to enhance the model's knowledge manipulation capabilities. \textbf{Any more specific methods for improving the model's performance on downstream tasks, including RL, distillation, contrastive learning, and others, can be seamlessly integrated with KALE to further enhance the model's performance in specific domains.}

\subsection{More Results of KALE on Joint Training Settings } \label{app:j19}

In the main text, we adopt the setting of fine-tuning on each benchmark’s training set separately, primarily to ensure consistency with prior work, such as MeanLearn~\citep{meaningful}, which follows the same protocol. This alignment allows for a more direct and fair comparison with existing methods.

To validate the effectiveness of KALE on cross-task generalization settings, we have conducted an additional joint training experiment. Specifically, we fine-tune our model using the merged training sets of the eight benchmarks listed in Table \ref{tab:kale_joint} in the original manuscript, using Llama3 8B as the backbone model.

\begin{table*}[t]
\centering
\caption{Results of KALE on joint training experiments using Llama3 8B as the backbone model.}
\label{tab:kale_joint}
\resizebox{\textwidth}{!}{%
\begin{tabular}{lrrrrrrrr}
\toprule
Method & AbsR & ARC-c & ARC-e & Common & MMLU & BBH & RACE-h & RACE-m \\
\midrule
\textbf{KALE\_ori}   & 83.62 & \textbf{81.23} & \textbf{86.45} & 65.69 & \textbf{63.27} & 57.33 & \textbf{68.61} & \textbf{74.12} \\
\textbf{KALE\_joint} & \textbf{84.48} & 78.16 & 84.13 & \textbf{67.81} & 60.68 & \textbf{58.13} & 64.41 & 71.57 \\
\bottomrule
\end{tabular}}
\end{table*}

 Notably, we observe that joint training does not significantly degrade performance across the benchmarks. On the contrary, the model even achieves improved performance for the logical reasoning task on datasets including AbsR, Commonsense, and BBH. 

These findings highlight that KALE is not limited to a task-specific fine-tuning paradigm. Instead, it is compatible with multi-task or unified training setups, further demonstrating its flexibility and potential for real-world deployment scenarios where a single model is required to handle diverse tasks.


\section{More Discussions On KALE} \label{app:more_discuss}

\paragraph{What named entity recognition method is employed in KALE, and does it have any tailored designs?}
Given the relative maturity of named entity recognition (NER), we do not elaborate on it in the main text. Considering the need for rapid deployment and ease of implementation, we utilized the SpaCy open-source library for the NER component. Moreover, we employ noun phrase extraction from the NLTK library to retain some non-named yet significant nouns in given Q\&A pairs. We  reference the entity list from Wikidata for entity recognition. 

\paragraph{The entity linking is a very important component in the proposed approach. Which entity linker is used for KALE, and does it have any tailored designs?} In our entity linking process, we employ a simple keyword-matching algorithm as our entity linker. Given that a KG may contain multiple distinct entities with identical names, our entity linking might link incorrect entities, i.e., entity linking errors. In the entity linking process of KALE, potential errors may not be consistently measured and filtered out. Instead, for each set of candidate entities, we link their neighbor entities to enrich candidate entities. Then, we utilize the context to retain the relevant candidates. Generally, incorrect candidate entities are unrelated to the context, which are less likely to be retained.
For instance, neighbor entities related to "Apple" (fruit) include "Cider Making", "Wild Apples", and "Apples from Maine" whereas those related to "Apple" (company) include "FileMaker", "Apple Germany", and various products like "iPod", "iMac", and "iPhone". If the correct entity in the reference context is "Apple" (fruit), it is less likely that neighbors related to Apple (company) appear in the reference context and vice versa.
Moreover, in certain cases, we might still retain some incorrect entities (unrelated to the query). However, in most situations where the context and query are related, these incorrect entities are less likely to be key lexical units of the context. 
 We acknowledge that integrating statistical or deep learning-based entity linkers—such as TAGME~\citep{TAGME}, DBpedia Spotlight~\citep{DB_spotlight}, BLINK~\citep{BLINK}, and GENRE~\citep{GENRE}—represents a promising direction for further optimizing KALE. Crucially, it is worth noting that the modules within KALE are designed to be highly decoupled. This architecture ensures flexibility in practical applications, allowing users to independently select and substitute linkers for downstream tasks according to specific resource constraints or precision requirements.  

\paragraph{Why is the A* algorithm employed for knowledge-induced data synthesis instead of the na\"{\i}ve BFS algorithm?} 
 In knowledge-induced data synthesis, we select the A* algorithm over the na\"{\i}ve BFS based on algorithmic efficiency. The A* algorithm guides the search direction by incorporating a heuristic function $h(\textbf{n})$, which significantly reduces the exploration scope. Particularly in large-scale KGs such as Wikidata, employing BFS to identify reasoning paths often requires days of computation. As mentioned in Section \ref{sec:multi-astart}, the extraction of reasoning paths from the AbsR's training set 
\textbf{requires over one week}. Therefore, we propose an efficient multi-path A* algorithm to extract reasoning paths. It requires \textbf{less than 4 hours} to extract all reasoning paths on the same  set. Consequently, we adopt the A* algorithm as a scalable and efficient solution for reasoning path search.

\paragraph{Is it possible for some reasoning paths to not reach the answer entity?} During the process of extracting reasoning paths, instances may arise where the hop between the question entity and the answer entity exceeds the predefined threshold $m$. Nevertheless, the statistical data on the ABS dataset, as in Table \ref{app:avg_hop}, indicates that less than $2$\% of the $3$-hop inference paths are unable to reach the answer entity. This suggests that employing $3$-hop inference paths is a highly effective approach for extracting relevant information from the question to the answer. In such cases, we utilize the partial reasoning path that can be extracted-the path from the question entity to its neighboring entities within three hops---as enriched information for input. The ablation study results in Tables \ref{tab:abla} and \ref{tab_app:abla} further demonstrate the simplicity and effectiveness these types of reasoning paths.

\paragraph{If the KG contains errors that lead to incorrect reasoning paths, would GPT-4o generate wrong rationales?} 
(i) Owing to Wikidata's factually accurate, high-quality, community-driven, and dynamically growing character, extracted reasoning paths contain \textbf{negligible} factual or logical errors. This motivates us to generate high-quality rationales via large-scale Wikidata. (ii) To address potential errors in the rationale generation, we leverage GPT's in-context learning (ICL) by incorporating specific instructions in the prompt. This allows for a filtering and correction mechanism to be implicitly applied during reasoning. As shown in Appendix \ref{app:prompt_temp}, we instruct LLM to generate rationales by referring to the provided reasoning path: \textit{however, if the given reasoning path is irrelevant to the QA, generate a rationale based on your own knowledge.} This instruction helps that incorrect information is reduced.  
We empirically observe that utilizing the in-context learning ability is simple yet effective to reduce the error propagation with great domain robustness. 
Moreover, the design logic behind KALE is simplicity and practicality. We acknowledge that using a dedicated filtering mechanism could be an option.

\paragraph{What is the reason for choosing the KL divergence as the loss function?}

The selection of KL divergence is due to its ability to quantify the difference between two probability distributions. It encourages the model to compress the information contained in the rationale into its parameters \(\theta\). By forcing the two distributions to align, the model must ``internalize'' the information from the rationale \(\textbf{x}^\textbf{{rats}}\) into its parameters \(\theta\), such that it can perform well even when \(\textbf{x}^\textbf{{rats}}\) is unavailable (e.g., at test time). This minimization process implicitly guides the model to capture the underlying structure of the data, thereby facilitating the learning of meaningful representations without explicit supervision. Furthermore, KL divergence is essentially composed of entropy and cross-entropy. The knowledge-aware learning module in KALE can be viewed as a distillation process, designed to enhance the knowledge manipulation capabilities of LLMs during the testing phase, where rationale input is unavailable. \textbf{The addition of KL divergence is intended to enable the model to dynamically retrieve the task-relevant knowledge it has already mastered, which improves its knowledge manipulation capability.} We also believe that a theoretical analysis of KALE, especially the KL divergence part, could lead to a deeper understanding of our KALE. We agree that this is a promising direction for future work.

\section{LLM Usage}
We used a large language model (LLM)–based writing assistant for grammar and wording improvements on draft text. The LLM did not generate research ideas, claims, proofs, algorithms, code, figures, or analyses, and it did not have access to any non-public data. During rationale generation, we use LLMs to transfer reasoning path into rationales.  All edits suggested by the LLM were manually reviewed and either accepted or rewritten by the authors, who take full responsibility for the final content. The LLM is not an author of this paper.


\begin{thebibliography}{103}
\providecommand{\natexlab}[1]{#1}

\bibitem[{Allen{-}Zhu and Li(2025)}]{know_mani}
Zeyuan Allen{-}Zhu and Yuanzhi Li. 2025.
\newblock \href {https://openreview.net/forum?id=oDbiL9CLoS} {Physics of
  language models: Part 3.2, knowledge manipulation}.
\newblock In \emph{The Thirteenth International Conference on Learning
  Representations, {ICLR} 2025, Singapore, April 24-28, 2025}. OpenReview.net.

\bibitem[{{Anthropic Team}(2024)}]{claude3}
{Anthropic Team}. 2024.
\newblock \href {https://api.semanticscholar.org/CorpusID:268232499} {The
  {Claude} 3 model family: {Opus}, {Sonnet}, {Haiku}}.
\newblock Technical report, Anthropic.

\bibitem[{Auer et~al.(2007)Auer, Bizer, Kobilarov, Lehmann, Cyganiak, and
  Ives}]{dbpedia}
S{\"{o}}ren Auer, Christian Bizer, Georgi Kobilarov, Jens Lehmann, Richard
  Cyganiak, and Zachary~G. Ives. 2007.
\newblock \href {https://doi.org/10.1007/978-3-540-76298-0\_52} {{DB}pedia: {A}
  nucleus for a web of open data}.
\newblock In \emph{The Semantic Web, 6th International Semantic Web Conference,
  2nd Asian Semantic Web Conference, {ISWC} 2007 + {ASWC} 2007, Busan, Korea,
  November 11-15, 2007}, volume 4825 of \emph{Lecture Notes in Computer
  Science}, pages 722--735. Springer.

\bibitem[{Baek et~al.(2023)Baek, Aji, and Saffari}]{kaping}
Jinheon Baek, Alham~Fikri Aji, and Amir Saffari. 2023.
\newblock \href {https://doi.org/10.48550/ARXIV.2306.04136}
  {Knowledge-augmented language model prompting for zero-shot knowledge graph
  question answering}.
\newblock \emph{CoRR}, abs/2306.04136.

\bibitem[{Belinkov and Bisk(2018)}]{cano1}
Yonatan Belinkov and Yonatan Bisk. 2018.
\newblock \href {https://openreview.net/forum?id=BJ8vJebC-} {Synthetic and
  natural noise both break neural machine translation}.
\newblock In \emph{6th International Conference on Learning Representations,
  {ICLR} 2018, Vancouver, BC, Canada, April 30 - May 3, 2018, Conference Track
  Proceedings}. OpenReview.net.

\bibitem[{Bhagavatula et~al.(2020)Bhagavatula, Bras, Malaviya, Sakaguchi,
  Holtzman, Rashkin, Downey, Yih, and Choi}]{bhagavatulaabductive}
Chandra Bhagavatula, Ronan~Le Bras, Chaitanya Malaviya, Keisuke Sakaguchi, Ari
  Holtzman, Hannah Rashkin, Doug Downey, Wen{-}tau Yih, and Yejin Choi. 2020.
\newblock \href {https://openreview.net/forum?id=Byg1v1HKDB} {Abductive
  commonsense reasoning}.
\newblock In \emph{8th International Conference on Learning Representations,
  {ICLR} 2020, Addis Ababa, Ethiopia, April 26-30, 2020}. OpenReview.net.

\bibitem[{Bhakthavatsalam et~al.(2021)Bhakthavatsalam, Khashabi, Khot, Mishra,
  Richardson, Sabharwal, Schoenick, Tafjord, and Clark}]{arc}
Sumithra Bhakthavatsalam, Daniel Khashabi, Tushar Khot, Bhavana~Dalvi Mishra,
  Kyle Richardson, Ashish Sabharwal, Carissa Schoenick, Oyvind Tafjord, and
  Peter Clark. 2021.
\newblock \href {https://arxiv.org/abs/2102.03315} {Think you have solved
  direct-answer question answering? try arc-da, the direct-answer {AI2}
  reasoning challenge}.
\newblock \emph{CoRR}, abs/2102.03315.

\bibitem[{Bisk et~al.(2020)Bisk, Zellers, Bras, Gao, and Choi}]{bisk2020piqa}
Yonatan Bisk, Rowan Zellers, Ronan~Le Bras, Jianfeng Gao, and Yejin Choi. 2020.
\newblock \href {https://doi.org/10.1609/AAAI.V34I05.6239} {{PIQA:} reasoning
  about physical commonsense in natural language}.
\newblock In \emph{The Thirty-Fourth {AAAI} Conference on Artificial
  Intelligence, {AAAI} 2020, The Thirty-Second Innovative Applications of
  Artificial Intelligence Conference, {IAAI} 2020, The Tenth {AAAI} Symposium
  on Educational Advances in Artificial Intelligence, {EAAI} 2020, New York,
  NY, USA, February 7-12, 2020}, pages 7432--7439. {AAAI} Press.

\bibitem[{Brown et~al.(2020)Brown, Mann, Ryder, Subbiah, Kaplan, Dhariwal,
  Neelakantan, Shyam, Sastry, Askell, Agarwal, Herbert{-}Voss, Krueger,
  Henighan, Child, Ramesh, Ziegler, Wu, Winter, Hesse, Chen, Sigler, Litwin,
  Gray, Chess, Clark, Berner, McCandlish, Radford, Sutskever, and
  Amodei}]{gpt3}
Tom~B. Brown, Benjamin Mann, Nick Ryder, Melanie Subbiah, Jared Kaplan,
  Prafulla Dhariwal, Arvind Neelakantan, Pranav Shyam, Girish Sastry, Amanda
  Askell, Sandhini Agarwal, Ariel Herbert{-}Voss, Gretchen Krueger, Tom
  Henighan, Rewon Child, Aditya Ramesh, Daniel~M. Ziegler, Jeffrey Wu, Clemens
  Winter, and 12 others. 2020.
\newblock \href
  {https://proceedings.neurips.cc/paper/2020/hash/1457c0d6bfcb4967418bfb8ac142f64a-Abstract.html}
  {Language models are few-shot learners}.
\newblock In \emph{Advances in Neural Information Processing Systems 33: Annual
  Conference on Neural Information Processing Systems 2020, NeurIPS 2020,
  December 6-12, 2020, virtual}.

\bibitem[{Buckner and Wheeler(2001)}]{na2}
Randy~L. Buckner and Mark~E. Wheeler. 2001.
\newblock \href {https://api.semanticscholar.org/CorpusID:205020040} {The
  cognitive neuroscience og remembering}.
\newblock \emph{Nature Reviews Neuroscience}, 2:624--634.

\bibitem[{Cao et~al.(2021)Cao, Izacard, Riedel, and Petroni}]{GENRE}
Nicola~De Cao, Gautier Izacard, Sebastian Riedel, and Fabio Petroni. 2021.
\newblock \href {https://openreview.net/forum?id=5k8F6UU39V} {Autoregressive
  entity retrieval}.
\newblock In \emph{9th International Conference on Learning Representations,
  {ICLR} 2021, Virtual Event, Austria, May 3-7, 2021}. OpenReview.net.

\bibitem[{Chan et~al.(2023)Chan, Zeng, Lake, Joshi, Chen, and Ren}]{knife}
Aaron Chan, Zhiyuan Zeng, Wyatt Lake, Brihi Joshi, Hanjie Chen, and Xiang Ren.
  2023.
\newblock \href {https://arxiv.org/abs/2212.09721} {{KNIFE}: Distilling
  reasoning knowledge from free-text rationales}.
\newblock \emph{Preprint}, arXiv:2212.09721.

\bibitem[{Chen et~al.(2024)Chen, Shen, Lv, Wang, Ni, and Ye}]{sac-kg}
Hanzhu Chen, Xu~Shen, Qitan Lv, Jie Wang, Xiaoqi Ni, and Jieping Ye. 2024.
\newblock \href {https://doi.org/10.18653/v1/2024.acl-long.238} {{SAC}-{KG}:
  Exploiting large language models as skilled automatic constructors for domain
  knowledge graph}.
\newblock In \emph{Proceedings of the 62nd Annual Meeting of the Association
  for Computational Linguistics (Volume 1: Long Papers)}, pages 4345--4360,
  Bangkok, Thailand. Association for Computational Linguistics.

\bibitem[{Chen et~al.(2025{\natexlab{a}})Chen, Shen, Wang, Wang, Lv, He, Wu,
  Wu, and Ye}]{kg-sft}
Hanzhu Chen, Xu~Shen, Jie Wang, Zehao Wang, Qitan Lv, Junjie He, Rong Wu, Feng
  Wu, and Jieping Ye. 2025{\natexlab{a}}.
\newblock \href {https://openreview.net/forum?id=oMFOKjwaRS} {Knowledge graph
  finetuning enhances knowledge manipulation in large language models}.
\newblock In \emph{The Thirteenth International Conference on Learning
  Representations, {ICLR} 2025, Singapore, April 24-28, 2025}. OpenReview.net.

\bibitem[{Chen et~al.(2025{\natexlab{b}})Chen, Sun, Wenjin, Zhang, Chen, Sun,
  Su, Pan, Klakow, Li, and Shen}]{cotdis}
Xinghao Chen, Zhijing Sun, Guo Wenjin, Miaoran Zhang, Yanjun Chen, Yirong Sun,
  Hui Su, Yijie Pan, Dietrich Klakow, Wenjie Li, and Xiaoyu Shen.
  2025{\natexlab{b}}.
\newblock \href {https://doi.org/10.18653/v1/2025.findings-acl.782} {Unveiling
  the key factors for distilling chain-of-thought reasoning}.
\newblock In \emph{Findings of the Association for Computational Linguistics:
  ACL 2025}, pages 15094--15119, Vienna, Austria. Association for Computational
  Linguistics.

\bibitem[{Chowdhery et~al.(2023)Chowdhery, Narang, Devlin, Bosma, Mishra,
  Roberts, Barham, Chung, Sutton, Gehrmann, Schuh, Shi, Tsvyashchenko, Maynez,
  Rao, Barnes, Tay, Shazeer, Prabhakaran, Reif, Du, Hutchinson, Pope, Bradbury,
  Austin, Isard, Gur{-}Ari, Yin, Duke, Levskaya, Ghemawat, Dev, Michalewski,
  Garcia, Misra, Robinson, Fedus, Zhou, Ippolito, Luan, Lim, Zoph, Spiridonov,
  Sepassi, Dohan, Agrawal, Omernick, Dai, Pillai, Pellat, Lewkowycz, Moreira,
  Child, Polozov, Lee, Zhou, Wang, Saeta, Diaz, Firat, Catasta, Wei,
  Meier{-}Hellstern, Eck, Dean, Petrov, and Fiedel}]{palm}
Aakanksha Chowdhery, Sharan Narang, Jacob Devlin, Maarten Bosma, Gaurav Mishra,
  Adam Roberts, Paul Barham, Hyung~Won Chung, Charles Sutton, Sebastian
  Gehrmann, Parker Schuh, Kensen Shi, Sasha Tsvyashchenko, Joshua Maynez,
  Abhishek Rao, Parker Barnes, Yi~Tay, Noam Shazeer, Vinodkumar Prabhakaran,
  and 48 others. 2023.
\newblock \href {https://jmlr.org/papers/v24/22-1144.html} {{P}a{LM}: Scaling
  language modeling with pathways}.
\newblock \emph{J. Mach. Learn. Res.}, 24:240:1--240:113.

\bibitem[{Coulombe(2018)}]{cano2}
Claude Coulombe. 2018.
\newblock \href {https://arxiv.org/abs/1812.04718} {Text data augmentation made
  simple by leveraging {NLP} cloud {API}s}.
\newblock \emph{CoRR}, abs/1812.04718.

\bibitem[{Dai et~al.(2023)Dai, Liu, Liao, Huang, Wu, Zhao, Liu, Liu, Li, Zhu,
  Cai, Li, Shen, Liu, and Li}]{auggpt}
Haixing Dai, Zhengliang Liu, Wenxiong Liao, Xiaoke Huang, Zihao Wu, Lin Zhao,
  Wei Liu, Ninghao Liu, Sheng Li, Dajiang Zhu, Hongmin Cai, Quanzheng Li,
  Dinggang Shen, Tianming Liu, and Xiang Li. 2023.
\newblock \href {https://doi.org/10.48550/ARXIV.2302.13007} {Chat{A}ug:
  Leveraging chatgpt for text data augmentation}.
\newblock \emph{CoRR}, abs/2302.13007.

\bibitem[{DeepSeek{-}AI(2024)}]{deepseek}
DeepSeek{-}AI. 2024.
\newblock \href {https://doi.org/10.48550/ARXIV.2412.19437} {{D}eep{S}eek-{V}3
  technical report}.
\newblock \emph{CoRR}, abs/2412.19437.

\bibitem[{Deng et~al.(2023)Deng, Zhang, Chen, and Gu}]{dataau1}
Yihe Deng, Weitong Zhang, Zixiang Chen, and Quanquan Gu. 2023.
\newblock \href {https://doi.org/10.48550/ARXIV.2311.04205} {Rephrase and
  respond: Let large language models ask better questions for themselves}.
\newblock \emph{CoRR}, abs/2311.04205.

\bibitem[{Devlin et~al.(2019)Devlin, Chang, Lee, and Toutanova}]{bert}
Jacob Devlin, Ming{-}Wei Chang, Kenton Lee, and Kristina Toutanova. 2019.
\newblock \href {https://doi.org/10.18653/V1/N19-1423} {{BERT:} pre-training of
  deep bidirectional transformers for language understanding}.
\newblock In \emph{Proceedings of the 2019 Conference of the North American
  Chapter of the Association for Computational Linguistics: Human Language
  Technologies, {NAACL-HLT} 2019, Minneapolis, MN, USA, June 2-7, 2019, Volume
  1 (Long and Short Papers)}, pages 4171--4186. Association for Computational
  Linguistics.

\bibitem[{Dhuliawala et~al.(2024)Dhuliawala, Komeili, Xu, Raileanu, Li,
  Celikyilmaz, and Weston}]{cov}
Shehzaad Dhuliawala, Mojtaba Komeili, Jing Xu, Roberta Raileanu, Xian Li, Asli
  Celikyilmaz, and Jason Weston. 2024.
\newblock \href {https://doi.org/10.18653/V1/2024.FINDINGS-ACL.212}
  {{C}hain-of-{V}erification reduces hallucination in large language models}.
\newblock In \emph{Findings of the Association for Computational Linguistics,
  {ACL} 2024, Bangkok, Thailand and virtual meeting, August 11-16, 2024}, pages
  3563--3578. Association for Computational Linguistics.

\bibitem[{Ding et~al.(2024)Ding, Qin, Zhao, Luo, Li, Chen, Xia, Hu, Luu, and
  Joty}]{survey_da}
Bosheng Ding, Chengwei Qin, Ruochen Zhao, Tianze Luo, Xinze Li, Guizhen Chen,
  Wenhan Xia, Junjie Hu, Anh~Tuan Luu, and Shafiq Joty. 2024.
\newblock \href {https://doi.org/10.18653/V1/2024.FINDINGS-ACL.97} {Data
  augmentation using llms: Data perspectives, learning paradigms and
  challenges}.
\newblock In \emph{Findings of the Association for Computational Linguistics,
  {ACL} 2024, Bangkok, Thailand and virtual meeting, August 11-16, 2024}, pages
  1679--1705. Association for Computational Linguistics.

\bibitem[{Dong et~al.(2024)Dong, Yuan, Lu, Li, Xue, Liu, Wang, Yuan, Zhou, and
  Zhou}]{mlt}
Guanting Dong, Hongyi Yuan, Keming Lu, Chengpeng Li, Mingfeng Xue, Dayiheng
  Liu, Wei Wang, Zheng Yuan, Chang Zhou, and Jingren Zhou. 2024.
\newblock \href {https://doi.org/10.18653/V1/2024.ACL-LONG.12} {How abilities
  in large language models are affected by supervised fine-tuning data
  composition}.
\newblock In \emph{Proceedings of the 62nd Annual Meeting of the Association
  for Computational Linguistics (Volume 1: Long Papers), {ACL} 2024, Bangkok,
  Thailand, August 11-16, 2024}, pages 177--198. Association for Computational
  Linguistics.

\bibitem[{Du et~al.(2022)Du, Ding, Xiong, Liu, and Qin}]{du2022care}
Li~Du, Xiao Ding, Kai Xiong, Ting Liu, and Bing Qin. 2022.
\newblock \href {https://doi.org/10.18653/v1/2022.acl-long.33} {e-{CARE}: a new
  dataset for exploring explainable causal reasoning}.
\newblock In \emph{Proceedings of the 60th Annual Meeting of the Association
  for Computational Linguistics (Volume 1: Long Papers)}, pages 432--446,
  Dublin, Ireland. Association for Computational Linguistics.

\bibitem[{Edge et~al.(2024)Edge, Trinh, Cheng, Bradley, Chao, Mody, Truitt, and
  Larson}]{graphrag}
Darren Edge, Ha~Trinh, Newman Cheng, Joshua Bradley, Alex Chao, Apurva Mody,
  Steven Truitt, and Jonathan Larson. 2024.
\newblock \href {https://doi.org/10.48550/ARXIV.2404.16130} {From local to
  global: {A} graph {RAG} approach to query-focused summarization}.
\newblock \emph{CoRR}, abs/2404.16130.

\bibitem[{Fang et~al.(2023)Fang, Lee, and Zhai}]{dataau2}
Luyang Fang, Gyeong{-}Geon Lee, and Xiaoming Zhai. 2023.
\newblock \href {https://doi.org/10.48550/ARXIV.2310.18365} {Using {GPT-4} to
  augment unbalanced data for automatic scoring}.
\newblock \emph{CoRR}, abs/2310.18365.

\bibitem[{Ferragina and Scaiella(2010)}]{TAGME}
Paolo Ferragina and Ugo Scaiella. 2010.
\newblock \href {https://doi.org/10.1145/1871437.1871689} {{TAGME:} on-the-fly
  annotation of short text fragments (by wikipedia entities)}.
\newblock In \emph{Proceedings of the 19th {ACM} Conference on Information and
  Knowledge Management, {CIKM} 2010, Toronto, Ontario, Canada, October 26-30,
  2010}, pages 1625--1628. {ACM}.

\bibitem[{Geva et~al.(2021)Geva, Khashabi, Segal, Khot, Roth, and
  Berant}]{geva2021did}
Mor Geva, Daniel Khashabi, Elad Segal, Tushar Khot, Dan Roth, and Jonathan
  Berant. 2021.
\newblock \href {https://doi.org/10.1162/TACL\_A\_00370} {Did aristotle use a
  laptop? {A} question answering benchmark with implicit reasoning strategies}.
\newblock \emph{Trans. Assoc. Comput. Linguistics}, 9:346--361.

\bibitem[{Goldberg and Harrelson(2005)}]{M_astar}
Andrew~V. Goldberg and Chris Harrelson. 2005.
\newblock \href {http://dl.acm.org/citation.cfm?id=1070432.1070455} {Computing
  the shortest path: \emph{A} search meets graph theory}.
\newblock In \emph{Proceedings of the Sixteenth Annual {ACM-SIAM} Symposium on
  Discrete Algorithms, {SODA} 2005, Vancouver, British Columbia, Canada,
  January 23-25, 2005}, pages 156--165. {SIAM}.

\bibitem[{{Google DeepMind Team}(2023)}]{gemini}
{Google DeepMind Team}. 2023.
\newblock \href {https://doi.org/10.48550/ARXIV.2312.11805} {Gemini: {A} family
  of highly capable multimodal models}.
\newblock \emph{CoRR}, abs/2312.11805.

\bibitem[{Gordon et~al.(2012)Gordon, Kozareva, and
  Roemmele}]{roemmele2011choice}
Andrew Gordon, Zornitsa Kozareva, and Melissa Roemmele. 2012.
\newblock {S}em{E}val-2012 task 7: Choice of plausible alternatives: An
  evaluation of commonsense causal reasoning.
\newblock In \emph{*{SEM} 2012: The First Joint Conference on Lexical and
  Computational Semantics {--} Volume 1: Proceedings of the main conference and
  the shared task, and Volume 2: Proceedings of the Sixth International
  Workshop on Semantic Evaluation ({S}em{E}val 2012)}, pages 394--398,
  Montr{\'e}al, Canada. Association for Computational Linguistics.

\bibitem[{Guan et~al.(2024)Guan, Liu, Lin, Lu, He, Han, and Sun}]{kgr}
Xinyan Guan, Yanjiang Liu, Hongyu Lin, Yaojie Lu, Ben He, Xianpei Han, and
  Le~Sun. 2024.
\newblock \href {https://doi.org/10.1609/AAAI.V38I16.29770} {Mitigating large
  language model hallucinations via autonomous knowledge graph-based
  retrofitting}.
\newblock In \emph{Thirty-Eighth {AAAI} Conference on Artificial Intelligence,
  {AAAI} 2024, Thirty-Sixth Conference on Innovative Applications of Artificial
  Intelligence, {IAAI} 2024, Fourteenth Symposium on Educational Advances in
  Artificial Intelligence, {EAAI} 2014, February 20-27, 2024, Vancouver,
  Canada}, pages 18126--18134. {AAAI} Press.

\bibitem[{Hart et~al.(1968)Hart, Nilsson, and Raphael}]{astar}
Peter~E. Hart, Nils~J. Nilsson, and Bertram Raphael. 1968.
\newblock \href {https://doi.org/10.1109/TSSC.1968.300136} {A formal basis for
  the heuristic determination of minimum cost paths}.
\newblock \emph{{IEEE} Trans. Syst. Sci. Cybern.}, 4(2):100--107.

\bibitem[{Hendrycks et~al.(2021)Hendrycks, Burns, Basart, Zou, Mazeika, Song,
  and Steinhardt}]{mmlu}
Dan Hendrycks, Collin Burns, Steven Basart, Andy Zou, Mantas Mazeika, Dawn
  Song, and Jacob Steinhardt. 2021.
\newblock \href {https://openreview.net/forum?id=d7KBjmI3GmQ} {Measuring
  massive multitask language understanding}.
\newblock In \emph{9th International Conference on Learning Representations,
  {ICLR} 2021, Virtual Event, Austria, May 3-7, 2021}. OpenReview.net.

\bibitem[{Hsieh et~al.(2023)Hsieh, Li, Yeh, Nakhost, Fujii, Ratner, Krishna,
  Lee, and Pfister}]{dissbs}
Cheng-Yu Hsieh, Chun-Liang Li, Chih-kuan Yeh, Hootan Nakhost, Yasuhisa Fujii,
  Alex Ratner, Ranjay Krishna, Chen-Yu Lee, and Tomas Pfister. 2023.
\newblock \href {https://doi.org/10.18653/v1/2023.findings-acl.507} {Distilling
  step-by-step! outperforming larger language models with less training data
  and smaller model sizes}.
\newblock In \emph{Findings of the Association for Computational Linguistics:
  ACL 2023}, pages 8003--8017, Toronto, Canada. Association for Computational
  Linguistics.

\bibitem[{Huang et~al.(2023)Huang, Kwak, and An}]{coe}
Fan Huang, Haewoon Kwak, and Jisun An. 2023.
\newblock \href {https://doi.org/10.1145/3543873.3587320} {Chain of
  explanation: New prompting method to generate quality natural language
  explanation for implicit hate speech}.
\newblock In \emph{Companion Proceedings of the {ACM} Web Conference 2023,
  {WWW} 2023, Austin, TX, USA, 30 April 2023 - 4 May 2023}, pages 90--93.
  {ACM}.

\bibitem[{Jiang et~al.(2023{\natexlab{a}})Jiang, Sablayrolles, Mensch, Bamford,
  Chaplot, de~Las~Casas, Bressand, Lengyel, Lample, Saulnier, Lavaud, Lachaux,
  Stock, Scao, Lavril, Wang, Lacroix, and Sayed}]{mistral}
Albert~Qiaochu Jiang, Alexandre Sablayrolles, Arthur Mensch, Chris Bamford,
  Devendra~Singh Chaplot, Diego de~Las~Casas, Florian Bressand, Gianna Lengyel,
  Guillaume Lample, Lucile Saulnier, L{\'e}lio~Renard Lavaud, Marie-Anne
  Lachaux, Pierre Stock, Teven~Le Scao, Thibaut Lavril, Thomas Wang,
  Timoth{\'e}e Lacroix, and William~El Sayed. 2023{\natexlab{a}}.
\newblock \href {https://api.semanticscholar.org/CorpusID:263830494} {Mistral
  7{B}}.
\newblock \emph{ArXiv}, abs/2310.06825.

\bibitem[{Jiang et~al.(2023{\natexlab{b}})Jiang, Zhou, Dong, Ye, Zhao, and
  Wen}]{structgpt}
Jinhao Jiang, Kun Zhou, Zican Dong, Keming Ye, Xin Zhao, and Ji{-}Rong Wen.
  2023{\natexlab{b}}.
\newblock \href {https://doi.org/10.18653/V1/2023.EMNLP-MAIN.574}
  {{S}truct{GPT}: {A} general framework for large language model to reason over
  structured data}.
\newblock In \emph{Proceedings of the 2023 Conference on Empirical Methods in
  Natural Language Processing, {EMNLP} 2023, Singapore, December 6-10, 2023},
  pages 9237--9251. Association for Computational Linguistics.

\bibitem[{Jin et~al.(2020)Jin, Pan, Oufattole, Weng, Fang, and
  Szolovits}]{medqa}
Di~Jin, Eileen Pan, Nassim Oufattole, Wei{-}Hung Weng, Hanyi Fang, and Peter
  Szolovits. 2020.
\newblock \href {https://arxiv.org/abs/2009.13081} {What disease does this
  patient have? {A} large-scale open domain question answering dataset from
  medical exams}.
\newblock \emph{CoRR}, abs/2009.13081.

\bibitem[{Kang et~al.(2023)Kang, Lee, Baek, Kawaguchi, and Hwang}]{kar}
Minki Kang, Seanie Lee, Jinheon Baek, Kenji Kawaguchi, and Sung~Ju Hwang. 2023.
\newblock \href
  {http://papers.nips.cc/paper\_files/paper/2023/hash/97faedc90260eae5c400f92d5831c3d7-Abstract-Conference.html}
  {Knowledge-augmented reasoning distillation for small language models in
  knowledge-intensive tasks}.
\newblock In \emph{Advances in Neural Information Processing Systems 36: Annual
  Conference on Neural Information Processing Systems 2023, NeurIPS 2023, New
  Orleans, LA, USA, December 10 - 16, 2023}.

\bibitem[{Kasai et~al.(2023)Kasai, Kasai, Sakaguchi, Yamada, and Radev}]{igaku}
Jungo Kasai, Yuhei Kasai, Keisuke Sakaguchi, Yutaro Yamada, and Dragomir Radev.
  2023.
\newblock \href {https://doi.org/10.48550/ARXIV.2303.18027} {Evaluating {GPT-4}
  and chat{GPT} on {J}apanese medical licensing examinations}.
\newblock \emph{CoRR}, abs/2303.18027.

\bibitem[{Kirkpatrick et~al.(2016)Kirkpatrick, Pascanu, Rabinowitz, Veness,
  Desjardins, Rusu, Milan, Quan, Ramalho, Grabska-Barwinska, Hassabis, Clopath,
  Kumaran, and Hadsell}]{forget}
James Kirkpatrick, Razvan Pascanu, Neil~C. Rabinowitz, Joel Veness, Guillaume
  Desjardins, Andrei~A. Rusu, Kieran Milan, John Quan, Tiago Ramalho, Agnieszka
  Grabska-Barwinska, Demis Hassabis, Claudia Clopath, Dharshan Kumaran, and
  Raia Hadsell. 2016.
\newblock \href {https://api.semanticscholar.org/CorpusID:4704285} {Overcoming
  catastrophic forgetting in neural networks}.
\newblock \emph{Proceedings of the National Academy of Sciences}, 114:3521 --
  3526.

\bibitem[{Kujanp{\"{a}}{\"{a}} et~al.(2025)Kujanp{\"{a}}{\"{a}}, Marttinen,
  Valpola, and Ilin}]{prompt_d}
Kalle Kujanp{\"{a}}{\"{a}}, Pekka Marttinen, Harri Valpola, and Alexander Ilin.
  2025.
\newblock \href {https://openreview.net/forum?id=drYpdSnRJk} {Efficient
  knowledge injection in llms via self-distillation}.
\newblock \emph{Trans. Mach. Learn. Res.}, 2025.

\bibitem[{Kullback and Leibler(1951)}]{kl}
Solomon Kullback and Richard~A Leibler. 1951.
\newblock On information and sufficiency.
\newblock \emph{The annals of mathematical statistics}, 22(1):79--86.

\bibitem[{Labrak et~al.(2022)Labrak, Bazoge, Dufour, Daille, Gourraud, Morin,
  and Rouvier}]{frenchmed}
Yanis Labrak, Adrien Bazoge, Richard Dufour, Beatrice Daille, Pierre-Antoine
  Gourraud, Emmanuel Morin, and Mickael Rouvier. 2022.
\newblock \href {https://doi.org/10.18653/v1/2022.louhi-1.5}
  {{F}rench{M}ed{MCQA}: A {F}rench multiple-choice question answering dataset
  for medical domain}.
\newblock In \emph{Proceedings of the 13th International Workshop on Health
  Text Mining and Information Analysis (LOUHI)}, pages 41--46, Abu Dhabi,
  United Arab Emirates (Hybrid). Association for Computational Linguistics.

\bibitem[{Lai et~al.(2017)Lai, Xie, Liu, Yang, and Hovy}]{race}
Guokun Lai, Qizhe Xie, Hanxiao Liu, Yiming Yang, and Eduard Hovy. 2017.
\newblock \href {https://doi.org/10.18653/v1/D17-1082} {{RACE}: Large-scale
  {R}e{A}ding comprehension dataset from examinations}.
\newblock In \emph{Proceedings of the 2017 Conference on Empirical Methods in
  Natural Language Processing}, pages 785--794, Copenhagen, Denmark.
  Association for Computational Linguistics.

\bibitem[{Lee and Hockenmaier(2025)}]{trace_quality}
Jinu Lee and Julia Hockenmaier. 2025.
\newblock \href {https://doi.org/10.48550/ARXIV.2502.12289} {Evaluating
  step-by-step reasoning traces: {A} survey}.
\newblock \emph{CoRR}, abs/2502.12289.

\bibitem[{Li et~al.(2025)Li, Dong, Tang, Wang, Zhang, Huang, Huang, Huang,
  Huang, Zhang, Gu, Cheng, Wang, Chen, Dong, Lu, Sui, Wang, Lam, and
  Wei}]{synthetic_data}
Haoran Li, Qingxiu Dong, Zhengyang Tang, Chaojun Wang, Xingxing Zhang, Haoyang
  Huang, Shaohan Huang, Xiaolong Huang, Zeqiang Huang, Dongdong Zhang, Yuxian
  Gu, Xin Cheng, Xun Wang, Si{-}Qing Chen, Li~Dong, Wei Lu, Zhifang Sui, Benyou
  Wang, Wai Lam, and Furu Wei. 2025.
\newblock \href {https://openreview.net/forum?id=PahnCreCxK} {Synthetic data
  (almost) from scratch: Generalized instruction tuning for language models}.
\newblock \emph{Trans. Mach. Learn. Res.}, 2025.

\bibitem[{Liu et~al.(2025)Liu, Li, Lv, Liu, Zhu, Hu, and Sun}]{pearl}
Tianyu Liu, Yun Li, Qitan Lv, Kai Liu, Jianchen Zhu, Winston Hu, and Xiao Sun.
  2025.
\newblock \href {https://openreview.net/forum?id=QOXrVMiHGK} {{PEARL:} parallel
  speculative decoding with adaptive draft length}.
\newblock In \emph{The Thirteenth International Conference on Learning
  Representations, {ICLR} 2025, Singapore, April 24-28, 2025}. OpenReview.net.

\bibitem[{Liu et~al.(2019)Liu, Ott, Goyal, Du, Joshi, Chen, Levy, Lewis,
  Zettlemoyer, and Stoyanov}]{roberta}
Yinhan Liu, Myle Ott, Naman Goyal, Jingfei Du, Mandar Joshi, Danqi Chen, Omer
  Levy, Mike Lewis, Luke Zettlemoyer, and Veselin Stoyanov. 2019.
\newblock \href {https://arxiv.org/abs/1907.11692} {{R}o{BERT}a: {A} robustly
  optimized {BERT} pretraining approach}.
\newblock \emph{CoRR}, abs/1907.11692.

\bibitem[{Luo et~al.(2024)Luo, Luo, Ding, Yuan, Xiao, and Zhang}]{rsft}
Junyu Luo, Xiao Luo, Kaize Ding, Jingyang Yuan, Zhiping Xiao, and Ming Zhang.
  2024.
\newblock \href {https://doi.org/10.48550/ARXIV.2412.14922} {{R}obust{FT}:
  Robust supervised fine-tuning for large language models under noisy
  response}.
\newblock \emph{CoRR}, abs/2412.14922.

\bibitem[{Lv et~al.(2024)Lv, Wang, Chen, Li, Zhang, and Wu}]{coft}
Qitan Lv, Jie Wang, Hanzhu Chen, Bin Li, Yongdong Zhang, and Feng Wu. 2024.
\newblock \href {https://openreview.net/forum?id=JCG0KTPVYy}
  {{C}oarse-to-{F}ine highlighting: Reducing knowledge hallucination in large
  language models}.
\newblock In \emph{Forty-first International Conference on Machine Learning,
  {ICML} 2024, Vienna, Austria, July 21-27, 2024}. OpenReview.net.

\bibitem[{Lyu et~al.(2024)Lyu, Yan, Wang, Shi, Yin, Ren, Chen, de~Rijke, and
  Ren}]{knowtuning}
Yougang Lyu, Lingyong Yan, Shuaiqiang Wang, Haibo Shi, Dawei Yin, Pengjie Ren,
  Zhumin Chen, Maarten de~Rijke, and Zhaochun Ren. 2024.
\newblock \href {https://doi.org/10.18653/v1/2024.emnlp-main.805}
  {{K}now{T}uning: Knowledge-aware fine-tuning for large language models}.
\newblock In \emph{Proceedings of the 2024 Conference on Empirical Methods in
  Natural Language Processing}, pages 14535--14556, Miami, Florida, USA.
  Association for Computational Linguistics.

\bibitem[{Madaan et~al.(2022)Madaan, Zhou, Alon, Yang, and Neubig}]{few-shot}
Aman Madaan, Shuyan Zhou, Uri Alon, Yiming Yang, and Graham Neubig. 2022.
\newblock \href {https://doi.org/10.48550/ARXIV.2210.07128} {Language models of
  code are few-shot commonsense learners}.
\newblock \emph{CoRR}, abs/2210.07128.

\bibitem[{Mendes et~al.(2011)Mendes, Jakob, Garc{\'{\i}}a{-}Silva, and
  Bizer}]{DB_spotlight}
Pablo~N. Mendes, Max Jakob, Andr{\'{e}}s Garc{\'{\i}}a{-}Silva, and Christian
  Bizer. 2011.
\newblock \href {https://doi.org/10.1145/2063518.2063519} {Dbpedia spotlight:
  shedding light on the web of documents}.
\newblock In \emph{Proceedings the 7th International Conference on Semantic
  Systems, {I-SEMANTICS} 2011, Graz, Austria, September 7-9, 2011}, {ACM}
  International Conference Proceeding Series, pages 1--8. {ACM}.

\bibitem[{Mitra et~al.(2023)Mitra, Corro, Mahajan, Codas, Simoes, Agrawal,
  Chen, Razdaibiedina, Jones, Aggarwal, Palangi, Zheng, Rosset, Khanpour, and
  Awadallah}]{orca2}
Arindam Mitra, Luciano~Del Corro, Shweti Mahajan, Andres Codas, Clarisse
  Simoes, Sahaj Agrawal, Xuxi Chen, Anastasia Razdaibiedina, Erik Jones, Kriti
  Aggarwal, Hamid Palangi, Guoqing Zheng, Corby Rosset, Hamed Khanpour, and
  Ahmed Awadallah. 2023.
\newblock \href {https://api.semanticscholar.org/CorpusID:265295592} {Orca 2:
  Teaching small language models how to reason}.
\newblock \emph{ArXiv}, abs/2311.11045.

\bibitem[{Muennighoff et~al.(2024)Muennighoff, Soldaini, Groeneveld, Lo,
  Morrison, Min, Shi, Walsh, Tafjord, Lambert, Gu, Arora, Bhagia, Schwenk,
  Wadden, Wettig, Hui, Dettmers, Kiela, Farhadi, Smith, Koh, Singh, and
  Hajishirzi}]{olmoe}
Niklas Muennighoff, Luca Soldaini, Dirk Groeneveld, Kyle Lo, Jacob~Daniel
  Morrison, Sewon Min, Weijia Shi, Pete Walsh, Oyvind Tafjord, Nathan Lambert,
  Yuling Gu, Shane Arora, Akshita Bhagia, Dustin Schwenk, David Wadden,
  Alexander Wettig, Binyuan Hui, Tim Dettmers, Douwe Kiela, and 5 others. 2024.
\newblock \href {https://api.semanticscholar.org/CorpusID:272366674} {Olmoe:
  Open mixture-of-experts language models}.
\newblock \emph{ArXiv}, abs/2409.02060.

\bibitem[{OpenAI(2023)}]{gpt4}
OpenAI. 2023.
\newblock \href {https://doi.org/10.48550/ARXIV.2303.08774} {{GPT-4} technical
  report}.
\newblock \emph{CoRR}, abs/2303.08774.

\bibitem[{Ouyang et~al.(2022)Ouyang, Wu, Jiang, Almeida, Wainwright, Mishkin,
  Zhang, Agarwal, Slama, Ray, Schulman, Hilton, Kelton, Miller, Simens, Askell,
  Welinder, Christiano, Leike, and Lowe}]{train_sft}
Long Ouyang, Jeffrey Wu, Xu~Jiang, Diogo Almeida, Carroll~L. Wainwright, Pamela
  Mishkin, Chong Zhang, Sandhini Agarwal, Katarina Slama, Alex Ray, John
  Schulman, Jacob Hilton, Fraser Kelton, Luke Miller, Maddie Simens, Amanda
  Askell, Peter Welinder, Paul~F. Christiano, Jan Leike, and Ryan Lowe. 2022.
\newblock \href
  {http://papers.nips.cc/paper\_files/paper/2022/hash/b1efde53be364a73914f58805a001731-Abstract-Conference.html}
  {Training language models to follow instructions with human feedback}.
\newblock In \emph{Advances in Neural Information Processing Systems 35: Annual
  Conference on Neural Information Processing Systems 2022, NeurIPS 2022, New
  Orleans, LA, USA, November 28 - December 9, 2022}.

\bibitem[{Pan et~al.(2024)Pan, Luo, Wang, Chen, Wang, and Wu}]{roadmap}
Shirui Pan, Linhao Luo, Yufei Wang, Chen Chen, Jiapu Wang, and Xindong Wu.
  2024.
\newblock \href {https://doi.org/10.1109/TKDE.2024.3352100} {Unifying large
  language models and knowledge graphs: {A} roadmap}.
\newblock \emph{{IEEE} Trans. Knowl. Data Eng.}, 36(7):3580--3599.

\bibitem[{Qiu et~al.(2024)Qiu, Wu, Zhang, Lin, Wang, Zhang, Wang, and
  Xie}]{rumed}
Pengcheng Qiu, Chaoyi Wu, Xiaoman Zhang, Weixiong Lin, Haicheng Wang, Ya~Zhang,
  Yanfeng Wang, and Weidi Xie. 2024.
\newblock \href {https://doi.org/10.48550/ARXIV.2402.13963} {Towards building
  multilingual language model for medicine}.
\newblock \emph{CoRR}, abs/2402.13963.

\bibitem[{Radford et~al.(2018)Radford, Narasimhan, Salimans, Sutskever
  et~al.}]{gpt1}
Alec Radford, Karthik Narasimhan, Tim Salimans, Ilya Sutskever, and 1 others.
  2018.
\newblock Improving language understanding by generative pre-training.
\newblock \emph{arXiv preprint}.

\bibitem[{Riviere et~al.(2024)Riviere, Pathak, Sessa, Hardin, Bhupatiraju,
  Hussenot, Mesnard, Shahriari, Ram'e, Ferret, Liu, Tafti, Friesen, Casbon,
  Ramos, Kumar, Lan, Jerome, Tsitsulin, Vieillard, Stańczyk, Girgin, Momchev,
  Hoffman, Thakoor, Grill, Neyshabur, Walton, Severyn, Parrish, Ahmad,
  Hutchison, Abdagic, Carl, Shen, Brock, Coenen, Laforge, Paterson, Bastian,
  Piot, Wu, Royal, Chen, Kumar, Perry, Welty, Choquette-Choo, Sinopalnikov,
  Weinberger, Vijaykumar, Rogozi'nska, Herbison, Bandy, Wang, Noland, Moreira,
  Senter, Eltyshev, Visin, Rasskin, Wei, Cameron, Martins, Hashemi,
  Klimczak-Pluci'nska, Batra, Dhand, Nardini, Mein, Zhou, Svensson, Stanway,
  Chan, Zhou, Carrasqueira, Iljazi, Becker, Fernandez, van Amersfoort, Gordon,
  Lipschultz, Newlan, Ji, Mohamed, Badola, Black, Millican, McDonell, Nguyen,
  Sodhia, Greene, Sjoesund, Usui, Sifre, Heuermann, cia Lago, McNealus, Soares,
  Kilpatrick, Dixon, Martins, Reid, Singh, Iverson, Gorner, Velloso, Wirth,
  Davidow, Miller, Rahtz, Watson, Risdal, Kazemi, Moynihan, Zhang, Kahng, Park,
  Rahman, Khatwani, Dao, shad Bardoliwalla, Devanathan, Dumai, Chauhan,
  Wahltinez, Botarda, Barnes, Barham, Michel, chong Jin, Georgiev, Culliton,
  Kuppala, Comanescu, Merhej, Jana, Rokni, Agarwal, Mullins, Saadat, Carthy,
  Perrin, Arnold, bastian Krause, Dai, Garg, Sheth, Ronstrom, Chan, Jordan, Yu,
  Eccles, Hennigan, Kocisk{\'y}, Doshi, Jain, Yadav, Meshram, Dharmadhikari,
  Barkley, Wei, Ye, Han, Kwon, Xu, Shen, Gong, Wei, Cotruta, Kirk, Rao, Giang,
  Peran, Warkentin, Collins, Barral, Ghahramani, Hadsell, Sculley, Banks,
  Dragan, Petrov, Vinyals, Dean, Hassabis, Kavukcuoglu, Farabet, Buchatskaya,
  Borgeaud, Fiedel, Joulin, Kenealy, Dadashi, and Andreev}]{gemma2}
Gemma Team~Morgane Riviere, Shreya Pathak, Pier~Giuseppe Sessa, Cassidy Hardin,
  Surya Bhupatiraju, L'eonard Hussenot, Thomas Mesnard, Bobak Shahriari,
  Alexandre Ram'e, Johan Ferret, Peter Liu, Pouya~Dehghani Tafti, Abe Friesen,
  Michelle Casbon, Sabela Ramos, Ravin Kumar, Charline~Le Lan, Sammy Jerome,
  Anton Tsitsulin, and 176 others. 2024.
\newblock \href {https://api.semanticscholar.org/CorpusID:270843326} {Gemma 2:
  Improving open language models at a practical size}.
\newblock \emph{ArXiv}, abs/2408.00118.

\bibitem[{Rodr{\'i}guez-Mazahua et~al.(2015)Rodr{\'i}guez-Mazahua,
  Rodr{\'i}guez-Enr{\'i}quez, S{\'a}nchez-Cervantes, Cervantes,
  Garc{\'i}a-Alcar{\'a}z, and Alor-Hern{\'a}ndez}]{daijia}
Lisbeth Rodr{\'i}guez-Mazahua, Cristian~Aar{\'o}n Rodr{\'i}guez-Enr{\'i}quez,
  Jos{\'e}~Luis S{\'a}nchez-Cervantes, Jair Cervantes, Jorge~Luis
  Garc{\'i}a-Alcar{\'a}z, and Giner Alor-Hern{\'a}ndez. 2015.
\newblock \href {https://api.semanticscholar.org/CorpusID:15706968} {A general
  perspective of big data: applications, tools, challenges and trends}.
\newblock \emph{The Journal of Supercomputing}, 72:3073 -- 3113.

\bibitem[{Rossi et~al.(2021)Rossi, Barbosa, Firmani, Matinata, and
  Merialdo}]{kge_survey}
Andrea Rossi, Denilson Barbosa, Donatella Firmani, Antonio Matinata, and Paolo
  Merialdo. 2021.
\newblock \href {https://doi.org/10.1145/3424672} {Knowledge graph embedding
  for link prediction: A comparative analysis}.
\newblock \emph{ACM Trans. Knowl. Discov. Data}, 15(2).

\bibitem[{Sap et~al.(2019)Sap, Rashkin, Chen, Le~Bras, and
  Choi}]{sap2019social}
Maarten Sap, Hannah Rashkin, Derek Chen, Ronan Le~Bras, and Yejin Choi. 2019.
\newblock \href {https://doi.org/10.18653/v1/D19-1454} {Social {IQ}a:
  Commonsense reasoning about social interactions}.
\newblock In \emph{Proceedings of the 2019 Conference on Empirical Methods in
  Natural Language Processing and the 9th International Joint Conference on
  Natural Language Processing (EMNLP-IJCNLP)}, pages 4463--4473, Hong Kong,
  China. Association for Computational Linguistics.

\bibitem[{Shao et~al.(2024)Shao, Wang, Zhu, Xu, Song, Zhang, Li, Wu, and
  Guo}]{grpo}
Zhihong Shao, Peiyi Wang, Qihao Zhu, Runxin Xu, Junxiao Song, Mingchuan Zhang,
  Y.~K. Li, Y.~Wu, and Daya Guo. 2024.
\newblock \href {https://doi.org/10.48550/ARXIV.2402.03300}
  {{D}eep{S}eek{M}ath: Pushing the limits of mathematical reasoning in open
  language models}.
\newblock \emph{CoRR}, abs/2402.03300.

\bibitem[{Shoeybi et~al.(2019)Shoeybi, Patwary, Puri, LeGresley, Casper, and
  Catanzaro}]{megantron}
Mohammad Shoeybi, Mostofa Patwary, Raul Puri, Patrick LeGresley, Jared Casper,
  and Bryan Catanzaro. 2019.
\newblock \href {https://arxiv.org/abs/1909.08053} {{M}egatron-{LM}: Training
  multi-billion parameter language models using model parallelism}.
\newblock \emph{CoRR}, abs/1909.08053.

\bibitem[{Speer et~al.(2017)Speer, Chin, and Havasi}]{conceptnet}
Robyn Speer, Joshua Chin, and Catherine Havasi. 2017.
\newblock \href {https://doi.org/10.1609/AAAI.V31I1.11164} {Conceptnet 5.5: An
  open multilingual graph of general knowledge}.
\newblock In \emph{Proceedings of the Thirty-First {AAAI} Conference on
  Artificial Intelligence, February 4-9, 2017, San Francisco, California,
  {USA}}, pages 4444--4451. {AAAI} Press.

\bibitem[{Srivastava et~al.(2023)Srivastava, Rastogi, Rao, Shoeb, Abid, Fisch,
  Brown, Santoro, Gupta, Garriga{-}Alonso, Kluska, Lewkowycz, Agarwal, Power,
  Ray, Warstadt, Kocurek, Safaya, Tazarv, Xiang, Parrish, Nie, Hussain, Askell,
  Dsouza, Slone, Rahane, Iyer, Andreassen, Madotto, Santilli,
  Stuhlm{\"{u}}ller, Dai, La, Lampinen, Zou, Jiang, Chen, Vuong, Gupta,
  Gottardi, Norelli, Venkatesh, Gholamidavoodi, Tabassum, Menezes, Kirubarajan,
  Mullokandov, Sabharwal, Herrick, Efrat, Erdem, Karakas, Roberts, Loe, Zoph,
  Bojanowski, {\"{O}}zyurt, Hedayatnia, Neyshabur, Inden, Stein, Ekmekci, Lin,
  Howald, Orinion, Diao, Dour, Stinson, Argueta, Ram{\'{\i}}rez, Singh,
  Rathkopf, Meng, Baral, Wu, Callison{-}Burch, Waites, Voigt, Manning, Potts,
  Ramirez, Rivera, Siro, Raffel, Ashcraft, Garbacea, Sileo, Garrette,
  Hendrycks, Kilman, Roth, Freeman, Khashabi, Levy, Gonz{\'{a}}lez, Perszyk,
  Hernandez, Chen, Ippolito, Gilboa, Dohan, Drakard, Jurgens, Datta, Ganguli,
  Emelin, Kleyko, Yuret, Chen, Tam, Hupkes, Misra, Buzan, Mollo, Yang, Lee,
  Schrader, Shutova, Cubuk, Segal, Hagerman, Barnes, Donoway, Pavlick,
  Rodol{\`{a}}, Lam, Chu, Tang, Erdem, Chang, Chi, Dyer, Jerzak, Kim, Manyasi,
  Zheltonozhskii, Xia, Siar, Mart{\'{\i}}nez{-}Plumed, Happ{\'{e}}, Chollet,
  Rong, Mishra, Winata, de~Melo, Kruszewski, Parascandolo, Mariani, Wang,
  Jaimovitch{-}L{\'{o}}pez, Betz, Gur{-}Ari, Galijasevic, Kim, Rashkin,
  Hajishirzi, Mehta, Bogar, Shevlin, Sch{\"{u}}tze, Yakura, Zhang, Wong, Ng,
  Noble, Jumelet, Geissinger, Kernion, Hilton, Lee, Fisac, Simon, Koppel,
  Zheng, Zou, Kocon, Thompson, Wingfield, Kaplan, Radom, Sohl{-}Dickstein,
  Phang, Wei, Yosinski, Novikova, Bosscher, Marsh, Kim, Taal, Engel, Alabi, Xu,
  Song, Tang, Waweru, Burden, Miller, Balis, Batchelder, Berant, Frohberg,
  Rozen, Hern{\'{a}}ndez{-}Orallo, Boudeman, Guerr, Jones, Tenenbaum, Rule,
  Chua, Kanclerz, Livescu, Krauth, Gopalakrishnan, Ignatyeva, Markert, Dhole,
  Gimpel, Omondi, Mathewson, Chiafullo, Shkaruta, Shridhar, McDonell,
  Richardson, Reynolds, Gao, Zhang, Dugan, Qin, Ochando, Morency, Moschella,
  Lam, Noble, Schmidt, He, Col{\'{o}}n, Metz, Senel, Bosma, Sap, ter Hoeve,
  Farooqi, Faruqui, Mazeika, Baturan, Marelli, Maru, Ram{\'{\i}}rez{-}Quintana,
  Tolkiehn, Giulianelli, Lewis, Potthast, Leavitt, Hagen, Schubert,
  Baitemirova, Arnaud, McElrath, Yee, Cohen, Gu, Ivanitskiy, Starritt, Strube,
  Swedrowski, Bevilacqua, Yasunaga, Kale, Cain, Xu, Suzgun, Walker, Tiwari,
  Bansal, Aminnaseri, Geva, Gheini, T., Peng, Chi, Lee, Krakover, Cameron,
  Roberts, Doiron, Martinez, Nangia, Deckers, Muennighoff, Keskar, Iyer,
  Constant, Fiedel, Wen, Zhang, Agha, Elbaghdadi, Levy, Evans, Casares, Doshi,
  Fung, Liang, Vicol, Alipoormolabashi, Liao, Liang, Chang, Eckersley, Htut,
  Hwang, Milkowski, Patil, Pezeshkpour, Oli, Mei, Lyu, Chen, Banjade, Rudolph,
  Gabriel, Habacker, Risco, Milli{\`{e}}re, Garg, Barnes, Saurous, Arakawa,
  Raymaekers, Frank, Sikand, Novak, Sitelew, LeBras, Liu, Jacobs, Zhang,
  Salakhutdinov, Chi, Lee, Stovall, Teehan, Yang, Singh, Mohammad, Anand,
  Dillavou, Shleifer, Wiseman, Gruetter, Bowman, Schoenholz, Han, Kwatra, Rous,
  Ghazarian, Ghosh, Casey, Bischoff, Gehrmann, Schuster, Sadeghi, Hamdan, Zhou,
  Srivastava, Shi, Singh, Asaadi, Gu, Pachchigar, Toshniwal, Upadhyay, Debnath,
  Shakeri, Thormeyer, Melzi, Reddy, Makini, Lee, Torene, Hatwar, Dehaene,
  Divic, Ermon, Biderman, Lin, Prasad, Piantadosi, Shieber, Misherghi,
  Kiritchenko, Mishra, Linzen, Schuster, Li, Yu, Ali, Hashimoto, Wu, Desbordes,
  Rothschild, Phan, Wang, Nkinyili, Schick, Kornev, Tunduny, Gerstenberg,
  Chang, Neeraj, Khot, Shultz, Shaham, Misra, Demberg, Nyamai, Raunak,
  Ramasesh, Prabhu, Padmakumar, Srikumar, Fedus, Saunders, Zhang, Vossen, Ren,
  Tong, Zhao, Wu, Shen, Yaghoobzadeh, Lakretz, Song, Bahri, Choi, Yang, Hao,
  Chen, Belinkov, Hou, Hou, Bai, Seid, Zhao, Wang, Wang, Wang, and Wu}]{bb}
Aarohi Srivastava, Abhinav Rastogi, Abhishek Rao, Abu Awal~Md Shoeb, Abubakar
  Abid, Adam Fisch, Adam~R. Brown, Adam Santoro, Aditya Gupta, Adri{\`{a}}
  Garriga{-}Alonso, Agnieszka Kluska, Aitor Lewkowycz, Akshat Agarwal, Alethea
  Power, Alex Ray, Alex Warstadt, Alexander~W. Kocurek, Ali Safaya, Ali Tazarv,
  and 431 others. 2023.
\newblock \href {https://openreview.net/forum?id=uyTL5Bvosj} {Beyond the
  imitation game: Quantifying and extrapolating the capabilities of language
  models}.
\newblock \emph{Trans. Mach. Learn. Res.}, 2023.

\bibitem[{Sun et~al.(2023)Sun, Xu, Tang, Wang, Lin, Gong, Ni, yeung Shum, and
  Guo}]{tog}
Jiashuo Sun, Chengjin Xu, Lumingyuan Tang, Sai Wang, Chen Lin, Yeyun Gong,
  Lionel~M. Ni, Heung yeung Shum, and Jian Guo. 2023.
\newblock \href {https://api.semanticscholar.org/CorpusID:263333907}
  {{T}hink-on-{G}raph: Deep and responsible reasoning of large language model
  on knowledge graph}.
\newblock In \emph{International Conference on Learning Representations}.

\bibitem[{Suzgun et~al.(2023)Suzgun, Scales, Sch{\"a}rli, Gehrmann, Tay, Chung,
  Chowdhery, Le, Chi, Zhou, and Wei}]{bbh}
Mirac Suzgun, Nathan Scales, Nathanael Sch{\"a}rli, Sebastian Gehrmann, Yi~Tay,
  Hyung~Won Chung, Aakanksha Chowdhery, Quoc Le, Ed~Chi, Denny Zhou, and Jason
  Wei. 2023.
\newblock \href {https://doi.org/10.18653/v1/2023.findings-acl.824}
  {Challenging {BIG}-bench tasks and whether chain-of-thought can solve them}.
\newblock In \emph{Findings of the Association for Computational Linguistics:
  ACL 2023}, pages 13003--13051, Toronto, Canada. Association for Computational
  Linguistics.

\bibitem[{Talmor et~al.(2019)Talmor, Herzig, Lourie, and
  Berant}]{talmor2019commonsenseqa}
Alon Talmor, Jonathan Herzig, Nicholas Lourie, and Jonathan Berant. 2019.
\newblock \href {https://doi.org/10.18653/v1/N19-1421} {{C}ommonsense{QA}: A
  question answering challenge targeting commonsense knowledge}.
\newblock In \emph{Proceedings of the 2019 Conference of the North {A}merican
  Chapter of the Association for Computational Linguistics: Human Language
  Technologies, Volume 1 (Long and Short Papers)}, pages 4149--4158,
  Minneapolis, Minnesota. Association for Computational Linguistics.

\bibitem[{Team(2024)}]{llama3}
Llama Team. 2024.
\newblock \href {https://doi.org/10.48550/ARXIV.2407.21783} {The {L}lama 3 herd
  of models}.
\newblock \emph{CoRR}, abs/2407.21783.

\bibitem[{Touvron et~al.(2023)Touvron, Martin, Stone, Albert, Almahairi,
  Babaei, Bashlykov, Batra, Bhargava, Bhosale, Bikel, Blecher, Canton{-}Ferrer,
  Chen, Cucurull, Esiobu, Fernandes, Fu, Fu, Fuller, Gao, Goswami, Goyal,
  Hartshorn, Hosseini, Hou, Inan, Kardas, Kerkez, Khabsa, Kloumann, Korenev,
  Koura, Lachaux, Lavril, Lee, Liskovich, Lu, Mao, Martinet, Mihaylov, Mishra,
  Molybog, Nie, Poulton, Reizenstein, Rungta, Saladi, Schelten, Silva, Smith,
  Subramanian, Tan, Tang, Taylor, Williams, Kuan, Xu, Yan, Zarov, Zhang, Fan,
  Kambadur, Narang, Rodriguez, Stojnic, Edunov, and Scialom}]{llama2}
Hugo Touvron, Louis Martin, Kevin Stone, Peter Albert, Amjad Almahairi, Yasmine
  Babaei, Nikolay Bashlykov, Soumya Batra, Prajjwal Bhargava, Shruti Bhosale,
  Dan Bikel, Lukas Blecher, Cristian Canton{-}Ferrer, Moya Chen, Guillem
  Cucurull, David Esiobu, Jude Fernandes, Jeremy Fu, Wenyin Fu, and 49 others.
  2023.
\newblock \href {https://doi.org/10.48550/ARXIV.2307.09288} {Llama 2: Open
  foundation and fine-tuned chat models}.
\newblock \emph{CoRR}, abs/2307.09288.

\bibitem[{Vilares and G{\'o}mez-Rodr{\'i}guez(2019)}]{headqa}
David Vilares and Carlos G{\'o}mez-Rodr{\'i}guez. 2019.
\newblock \href {https://doi.org/10.18653/v1/P19-1092} {{HEAD}-{QA}: A
  healthcare dataset for complex reasoning}.
\newblock In \emph{Proceedings of the 57th Annual Meeting of the Association
  for Computational Linguistics}, pages 960--966, Florence, Italy. Association
  for Computational Linguistics.

\bibitem[{Wang et~al.(2024)Wang, Sun, Li, and Gao}]{cok}
Jianing Wang, Qiushi Sun, Xiang Li, and Ming Gao. 2024.
\newblock \href {https://doi.org/10.18653/V1/2024.ACL-LONG.271} {Boosting
  language models reasoning with chain-of-{K}nowledge prompting}.
\newblock In \emph{Proceedings of the 62nd Annual Meeting of the Association
  for Computational Linguistics (Volume 1: Long Papers), {ACL} 2024, Bangkok,
  Thailand, August 11-16, 2024}, pages 4958--4981. Association for
  Computational Linguistics.

\bibitem[{Wang et~al.(2023{\natexlab{a}})Wang, Wang, Li, Gao, Yin, and
  Ren}]{scott}
Peifeng Wang, Zhengyang Wang, Zheng Li, Yifan Gao, Bing Yin, and Xiang Ren.
  2023{\natexlab{a}}.
\newblock \href {https://doi.org/10.18653/v1/2023.acl-long.304} {{SCOTT}:
  Self-consistent chain-of-thought distillation}.
\newblock In \emph{Proceedings of the 61st Annual Meeting of the Association
  for Computational Linguistics (Volume 1: Long Papers)}, pages 5546--5558,
  Toronto, Canada. Association for Computational Linguistics.

\bibitem[{Wang et~al.(2023{\natexlab{b}})Wang, Wei, Schuurmans, Le, Chi,
  Narang, Chowdhery, and Zhou}]{selfconsist}
Xuezhi Wang, Jason Wei, Dale Schuurmans, Quoc~V. Le, Ed~H. Chi, Sharan Narang,
  Aakanksha Chowdhery, and Denny Zhou. 2023{\natexlab{b}}.
\newblock \href {https://openreview.net/forum?id=1PL1NIMMrw} {Self-consistency
  improves chain of thought reasoning in language models}.
\newblock In \emph{The Eleventh International Conference on Learning
  Representations, {ICLR} 2023, Kigali, Rwanda, May 1-5, 2023}. OpenReview.net.

\bibitem[{Wang et~al.(2023{\natexlab{c}})Wang, Kordi, Mishra, Liu, Smith,
  Khashabi, and Hajishirzi}]{cano3}
Yizhong Wang, Yeganeh Kordi, Swaroop Mishra, Alisa Liu, Noah~A. Smith, Daniel
  Khashabi, and Hannaneh Hajishirzi. 2023{\natexlab{c}}.
\newblock \href {https://doi.org/10.18653/V1/2023.ACL-LONG.754} {Self-instruct:
  Aligning language models with self-generated instructions}.
\newblock In \emph{Proceedings of the 61st Annual Meeting of the Association
  for Computational Linguistics (Volume 1: Long Papers), {ACL} 2023, Toronto,
  Canada, July 9-14, 2023}, pages 13484--13508. Association for Computational
  Linguistics.

\bibitem[{Wei et~al.(2022{\natexlab{a}})Wei, Bosma, Zhao, Guu, Yu, Lester, Du,
  Dai, and Le}]{sft}
Jason Wei, Maarten Bosma, Vincent~Y. Zhao, Kelvin Guu, Adams~Wei Yu, Brian
  Lester, Nan Du, Andrew~M. Dai, and Quoc~V. Le. 2022{\natexlab{a}}.
\newblock \href {https://openreview.net/forum?id=gEZrGCozdqR} {Finetuned
  language models are zero-shot learners}.
\newblock In \emph{The Tenth International Conference on Learning
  Representations, {ICLR} 2022, Virtual Event, April 25-29, 2022}.
  OpenReview.net.

\bibitem[{Wei et~al.(2022{\natexlab{b}})Wei, Wang, Schuurmans, Bosma, Ichter,
  Xia, Chi, Le, and Zhou}]{cot1}
Jason Wei, Xuezhi Wang, Dale Schuurmans, Maarten Bosma, Brian Ichter, Fei Xia,
  Ed~H. Chi, Quoc~V. Le, and Denny Zhou. 2022{\natexlab{b}}.
\newblock \href
  {http://papers.nips.cc/paper\_files/paper/2022/hash/9d5609613524ecf4f15af0f7b31abca4-Abstract-Conference.html}
  {Chain-of-thought prompting elicits reasoning in large language models}.
\newblock In \emph{Advances in Neural Information Processing Systems 35: Annual
  Conference on Neural Information Processing Systems 2022, NeurIPS 2022, New
  Orleans, LA, USA, November 28 - December 9, 2022}.

\bibitem[{Wei and Zou(2019)}]{eda}
Jason~W. Wei and Kai Zou. 2019.
\newblock \href {https://doi.org/10.18653/V1/D19-1670} {{EDA:} easy data
  augmentation techniques for boosting performance on text classification
  tasks}.
\newblock In \emph{Proceedings of the 2019 Conference on Empirical Methods in
  Natural Language Processing and the 9th International Joint Conference on
  Natural Language Processing, {EMNLP-IJCNLP} 2019, Hong Kong, China, November
  3-7, 2019}, pages 6381--6387. Association for Computational Linguistics.

\bibitem[{Wu et~al.(2020)Wu, Petroni, Josifoski, Riedel, and
  Zettlemoyer}]{BLINK}
Ledell Wu, Fabio Petroni, Martin Josifoski, Sebastian Riedel, and Luke
  Zettlemoyer. 2020.
\newblock \href {https://doi.org/10.18653/v1/2020.emnlp-main.519} {Scalable
  zero-shot entity linking with dense entity retrieval}.
\newblock In \emph{Proceedings of the 2020 Conference on Empirical Methods in
  Natural Language Processing (EMNLP)}, pages 6397--6407, Online. Association
  for Computational Linguistics.

\bibitem[{Xie et~al.(2024)Xie, Chen, Yu, Sun, and Wu}]{minor_sft}
Shiming Xie, Hong Chen, Fred Yu, Zeye Sun, and Xiuyu Wu. 2024.
\newblock \href {https://doi.org/10.48550/ARXIV.2408.10642} {Minor {SFT} loss
  for {LLM} fine-tune to increase performance and reduce model deviation}.
\newblock \emph{CoRR}, abs/2408.10642.

\bibitem[{Xiong et~al.(2023)Xiong, Ding, Cao, Liu, and Qin}]{common}
Kai Xiong, Xiao Ding, Yixin Cao, Ting Liu, and Bing Qin. 2023.
\newblock \href {https://doi.org/10.18653/v1/2023.findings-emnlp.508}
  {Examining inter-consistency of large language models collaboration: An
  in-depth analysis via debate}.
\newblock In \emph{Findings of the Association for Computational Linguistics:
  EMNLP 2023}, pages 7572--7590, Singapore. Association for Computational
  Linguistics.

\bibitem[{Xiong et~al.(2024)Xiong, Ding, Liu, Qin, Xu, Yang, Liu, and
  Cao}]{meaningful}
Kai Xiong, Xiao Ding, Ting Liu, Bing Qin, Dongliang Xu, Qing Yang, Hongtao Liu,
  and Yixin Cao. 2024.
\newblock \href
  {http://papers.nips.cc/paper\_files/paper/2024/hash/da5498f88193ff61f0daea1940b819da-Abstract-Conference.html}
  {Meaningful learning: Enhancing abstract reasoning in large language models
  via generic fact guidance}.
\newblock In \emph{Advances in Neural Information Processing Systems 38: Annual
  Conference on Neural Information Processing Systems 2024, NeurIPS 2024,
  Vancouver, BC, Canada, December 10 - 15, 2024}.

\bibitem[{Yadav et~al.(2022)Yadav, Noble, Niemeyer, Terceros, Victor, Liston,
  and Rajasethupathy}]{na1}
Nakul Yadav, Chelsea Noble, James~E. Niemeyer, Andrea Terceros, Jonathan
  Victor, Conor Liston, and Priya Rajasethupathy. 2022.
\newblock \href {https://api.semanticscholar.org/CorpusID:250529641}
  {Prefrontal feature representations drive memory recall}.
\newblock \emph{Nature}, 608:153 -- 160.

\bibitem[{Yang et~al.(2025)Yang, Li, Yang, Zhang, Hui, Zheng, Yu, Gao, Huang,
  Lv, Zheng, Liu, Zhou, Huang, Hu, Ge, Wei, Lin, Tang, Yang, Tu, Zhang, Yang,
  Yang, Zhou, Lin, Dang, Bao, Yang, Yu, Deng, Li, Xue, Li, Zhang, Wang, Zhu,
  Men, Gao, Liu, Luo, Li, Tang, Yin, Ren, Wang, Zhang, Ren, Fan, Su, Zhang,
  Zhang, Wan, Liu, Wang, Cui, Zhang, Zhou, and Qiu}]{Qwen3}
An~Yang, Anfeng Li, Baosong Yang, Beichen Zhang, Binyuan Hui, Bo~Zheng, Bowen
  Yu, Chang Gao, Chengen Huang, Chenxu Lv, Chujie Zheng, Dayiheng Liu, Fan
  Zhou, Fei Huang, Feng Hu, Hao Ge, Haoran Wei, Huan Lin, Jialong Tang, and 40
  others. 2025.
\newblock \href {https://doi.org/10.48550/ARXIV.2505.09388} {Qwen3 technical
  report}.
\newblock \emph{CoRR}, abs/2505.09388.

\bibitem[{Yang et~al.(2024{\natexlab{a}})Yang, Yang, Zhang, Hui, Zheng, Yu, Li,
  Liu, Huang, Dong, Wei, Lin, Yang, Tu, Zhang, Yang, Yang, Zhou, Lin, Dang, Lu,
  Bao, Yang, Yu, Li, Xue, Zhang, Zhu, Men, Lin, Li, Xia, Ren, Ren, Fan, Su,
  Zhang, Wan, Liu, Cui, Zhang, Qiu, Quan, and Wang}]{qwen2.5}
Qwen~An Yang, Baosong Yang, Beichen Zhang, Binyuan Hui, Bo~Zheng, Bowen Yu,
  Chengyuan Li, Dayiheng Liu, Fei Huang, Guanting Dong, Haoran Wei, Huan Lin,
  Jian Yang, Jianhong Tu, Jianwei Zhang, Jianxin Yang, Jiaxin Yang, Jingren
  Zhou, Junyang Lin, and 25 others. 2024{\natexlab{a}}.
\newblock \href {https://api.semanticscholar.org/CorpusID:274859421} {Qwen2.5
  technical report}.
\newblock \emph{ArXiv}, abs/2412.15115.

\bibitem[{Yang et~al.(2024{\natexlab{b}})Yang, Pang, Feng, Wang, Chen, Zhu, and
  Liu}]{sdt}
Zhaorui Yang, Tianyu Pang, Haozhe Feng, Han Wang, Wei Chen, Minfeng Zhu, and
  Qian Liu. 2024{\natexlab{b}}.
\newblock \href {https://doi.org/10.18653/V1/2024.ACL-LONG.58}
  {Self-{D}istillation bridges distribution gap in language model fine-tuning}.
\newblock In \emph{Proceedings of the 62nd Annual Meeting of the Association
  for Computational Linguistics (Volume 1: Long Papers), {ACL} 2024, Bangkok,
  Thailand, August 11-16, 2024}, pages 1028--1043. Association for
  Computational Linguistics.

\bibitem[{Yoo et~al.(2021)Yoo, Park, Kang, Lee, and Park}]{gpt3mix}
Kang~Min Yoo, Dongju Park, Jaewook Kang, Sang{-}Woo Lee, and Woo{-}Myoung Park.
  2021.
\newblock \href {https://doi.org/10.18653/V1/2021.FINDINGS-EMNLP.192}
  {{GPT}3{M}ix: Leveraging large-scale language models for text augmentation}.
\newblock In \emph{Findings of the Association for Computational Linguistics:
  {EMNLP} 2021, Virtual Event / Punta Cana, Dominican Republic, 16-20 November,
  2021}, pages 2225--2239. Association for Computational Linguistics.

\bibitem[{Zan et~al.(2023)Zan, Chen, Zhang, Lu, Wu, Guan, Wang, and
  Lou}]{code_llm}
Daoguang Zan, Bei Chen, Fengji Zhang, Dianjie Lu, Bingchao Wu, Bei Guan, Yongji
  Wang, and Jian{-}Guang Lou. 2023.
\newblock \href {https://doi.org/10.18653/V1/2023.ACL-LONG.411} {Large language
  models meet {NL}2{C}ode: {A} survey}.
\newblock In \emph{Proceedings of the 61st Annual Meeting of the Association
  for Computational Linguistics (Volume 1: Long Papers), {ACL} 2023, Toronto,
  Canada, July 9-14, 2023}, pages 7443--7464. Association for Computational
  Linguistics.

\bibitem[{Zelikman et~al.(2024)Zelikman, Wu, Mu, and Goodman}]{star}
Eric Zelikman, YH~Wu, Jesse Mu, and Noah~D Goodman. 2024.
\newblock {ST}a{R}: Self-taught reasoner bootstrapping reasoning with
  reasoning.
\newblock In \emph{Proc. the 36th International Conference on Neural
  Information Processing Systems}, volume 1126.

\bibitem[{Zhang et~al.(2023)Zhang, Dong, Li, Zhang, Sun, Wang, Li, Hu, Zhang,
  Wu, and Wang}]{instruction_survey}
Shengyu Zhang, Linfeng Dong, Xiaoya Li, Sen Zhang, Xiaofei Sun, Shuhe Wang,
  Jiwei Li, Runyi Hu, Tianwei Zhang, Fei Wu, and Guoyin Wang. 2023.
\newblock \href {https://doi.org/10.48550/ARXIV.2308.10792} {Instruction tuning
  for large language models: {A} survey}.
\newblock \emph{CoRR}, abs/2308.10792.

\bibitem[{Zhang et~al.(2026)Zhang, Jin, Meng, Wang, and Tan}]{domain3}
Yang Zhang, Hanlei Jin, Dan Meng, Jun Wang, and Jinghua Tan. 2026.
\newblock \href {https://doi.org/10.1016/J.NEUCOM.2025.131928} {A comprehensive
  survey on automatic text summarization with exploration of llm-based
  methods}.
\newblock \emph{Neurocomputing}, 663:131928.

\bibitem[{Zhang et~al.(2024)Zhang, Wang, Liang, Xia, Chen, and Xiao}]{cokcok}
Yifei Zhang, Xintao Wang, Jiaqing Liang, Sirui Xia, Lida Chen, and Yanghua
  Xiao. 2024.
\newblock \href {https://doi.org/10.48550/ARXIV.2407.00653}
  {{C}hain-of-{K}nowledge: Integrating knowledge reasoning into large language
  models by learning from knowledge graphs}.
\newblock \emph{CoRR}, abs/2407.00653.

\bibitem[{Zhao et~al.(2024)Zhao, Zhao, Wang, Wang, and Xu}]{kgcot}
Ruilin Zhao, Feng Zhao, Long Wang, Xianzhi Wang, and Guandong Xu. 2024.
\newblock \href {https://www.ijcai.org/proceedings/2024/734} {{KG-CoT}:
  Chain-of-thought prompting of large language models over knowledge graphs for
  knowledge-aware question answering}.
\newblock In \emph{Proceedings of the Thirty-Third International Joint
  Conference on Artificial Intelligence, {IJCAI} 2024, Jeju, South Korea,
  August 3-9, 2024}, pages 6642--6650. ijcai.org.

\bibitem[{Zhao et~al.(2021)Zhao, Wallace, Feng, Klein, and Singh}]{domain1}
Zihao Zhao, Eric Wallace, Shi Feng, Dan Klein, and Sameer Singh. 2021.
\newblock \href {http://proceedings.mlr.press/v139/zhao21c.html} {Calibrate
  before use: Improving few-shot performance of language models}.
\newblock In \emph{Proceedings of the 38th International Conference on Machine
  Learning, {ICML} 2021, 18-24 July 2021, Virtual Event}, volume 139 of
  \emph{Proceedings of Machine Learning Research}, pages 12697--12706. {PMLR}.

\bibitem[{Zheng et~al.(2023)Zheng, Chiang, Sheng, Zhuang, Wu, Zhuang, Lin, Li,
  Li, Xing, Zhang, Gonzalez, and Stoica}]{vicuna}
Lianmin Zheng, Wei{-}Lin Chiang, Ying Sheng, Siyuan Zhuang, Zhanghao Wu,
  Yonghao Zhuang, Zi~Lin, Zhuohan Li, Dacheng Li, Eric~P. Xing, Hao Zhang,
  Joseph~E. Gonzalez, and Ion Stoica. 2023.
\newblock \href
  {http://papers.nips.cc/paper\_files/paper/2023/hash/91f18a1287b398d378ef22505bf41832-Abstract-Datasets\_and\_Benchmarks.html}
  {Judging {LLM}-as-a-{J}udge with {MT}-bench and chatbot arena}.
\newblock In \emph{Advances in Neural Information Processing Systems 36: Annual
  Conference on Neural Information Processing Systems 2023, NeurIPS 2023, New
  Orleans, LA, USA, December 10 - 16, 2023}.

\bibitem[{Zheng et~al.(2024)Zheng, Zhang, Zhang, Ye, and Luo}]{llamafactory}
Yaowei Zheng, Richong Zhang, Junhao Zhang, Yanhan Ye, and Zheyan Luo. 2024.
\newblock \href {https://doi.org/10.18653/v1/2024.acl-demos.38}
  {{L}lama{F}actory: Unified efficient fine-tuning of 100+ language models}.
\newblock In \emph{Proceedings of the 62nd Annual Meeting of the Association
  for Computational Linguistics (Volume 3: System Demonstrations)}, pages
  400--410, Bangkok, Thailand. Association for Computational Linguistics.

\bibitem[{Zhu et~al.(2023)Zhu, Yuan, Galkin, Xhonneux, Zhang, Gazeau, and
  Tang}]{a*net}
Zhaocheng Zhu, Xinyu Yuan, Michael Galkin, Louis{-}Pascal A.~C. Xhonneux, Ming
  Zhang, Maxime Gazeau, and Jian Tang. 2023.
\newblock \href
  {http://papers.nips.cc/paper\_files/paper/2023/hash/b9e98316cb72fee82cc1160da5810abc-Abstract-Conference.html}
  {A*net: {A} scalable path-based reasoning approach for knowledge graphs}.
\newblock In \emph{Advances in Neural Information Processing Systems 36: Annual
  Conference on Neural Information Processing Systems 2023, NeurIPS 2023, New
  Orleans, LA, USA, December 10 - 16, 2023}.

\end{thebibliography}
\end{document}